\documentclass{naturemi}
\usepackage{graphicx}
\usepackage{subcaption}
\usepackage{caption}
\usepackage{amsmath, amssymb}
\usepackage{natbib}
\usepackage{url}
\usepackage{xr-hyper}
\usepackage{hyperref}
\usepackage{times}
\usepackage{geometry}
\usepackage{soul}
\usepackage{algorithmicx}
\usepackage{algorithm}
\usepackage{xcolor}
\usepackage{algpseudocode}
\usepackage{tabularx}   
\usepackage{lmodern} 
\usepackage{booktabs}
\usepackage{multirow}
\usepackage{longtable}
\newtheorem{proposition}{Proposition}

\title{Variational decomposition autoencoding improves disentanglement of latent representations}
\author{
  Ioannis N. Ziogas$^{1,2}$, Aamna Al Shehhi$^{1}$, Ahsan H. Khandoker$^{1}$, Leontios J. Hadjileontiadis$^{2,1}$
  \\
  \small{$^{1}$Dept. of Biomedical Engineering, Khalifa University of Science and Technology, Abu Dhabi, UAE} \\
  \small{$^{2}$School of Electrical and Computer Engineering, Aristotle University of Thessaloniki, Thessaloniki, Greece} \\
  \centering
  \small{Corresponding author: \texttt{leontios\{@auth.gr,.hadjileontiadis@ku.ac.ae\}}} \\
}

\begin{document}

\date{}

\maketitle

\begin{abstract}
Understanding the structure of complex, nonstationary, high-dimensional time-evolving signals is a central challenge in scientific data analysis. In many domains, such as speech and biomedical signal processing, the ability to learn disentangled and interpretable representations is critical for uncovering latent generative mechanisms. Traditional approaches to unsupervised representation learning, including variational autoencoders (VAEs), often struggle to capture the temporal and spectral diversity inherent in such data. Here we introduce variational decomposition autoencoding (VDA), a framework that incorporates a strong structural bias toward signal decomposition into variational representation learning. VDA is instantiated through variational decomposition autoencoders (DecVAEs), i.e., encoder-only neural networks that perform latent autoencoding at the level of distributions to learn multiple latent subspaces aligned with time-frequency characteristics. We evaluate DecVAEs on simulated data and three publicly available scientific datasets, spanning speech recognition, dysarthric speech evaluation, and emotional speech classification. DecVAEs improve on VAE-based methods in disentanglement quality and robustness, and their latent subspaces become factor-specific when the generative factors have distinct spectral signatures. These findings position decomposition-aware architectures as potential tools for extracting structured representations from speech, and potentially from other dynamic signals with factors exhibiting distinct spectral signatures, with applications in clinical diagnostics, human-computer interaction, and adaptive neurotechnologies.

\end{abstract}

 \section*{Introduction}
\begin{figure*}[h!]
    \centering
    \includegraphics[width=0.95\textwidth,height = 15cm]{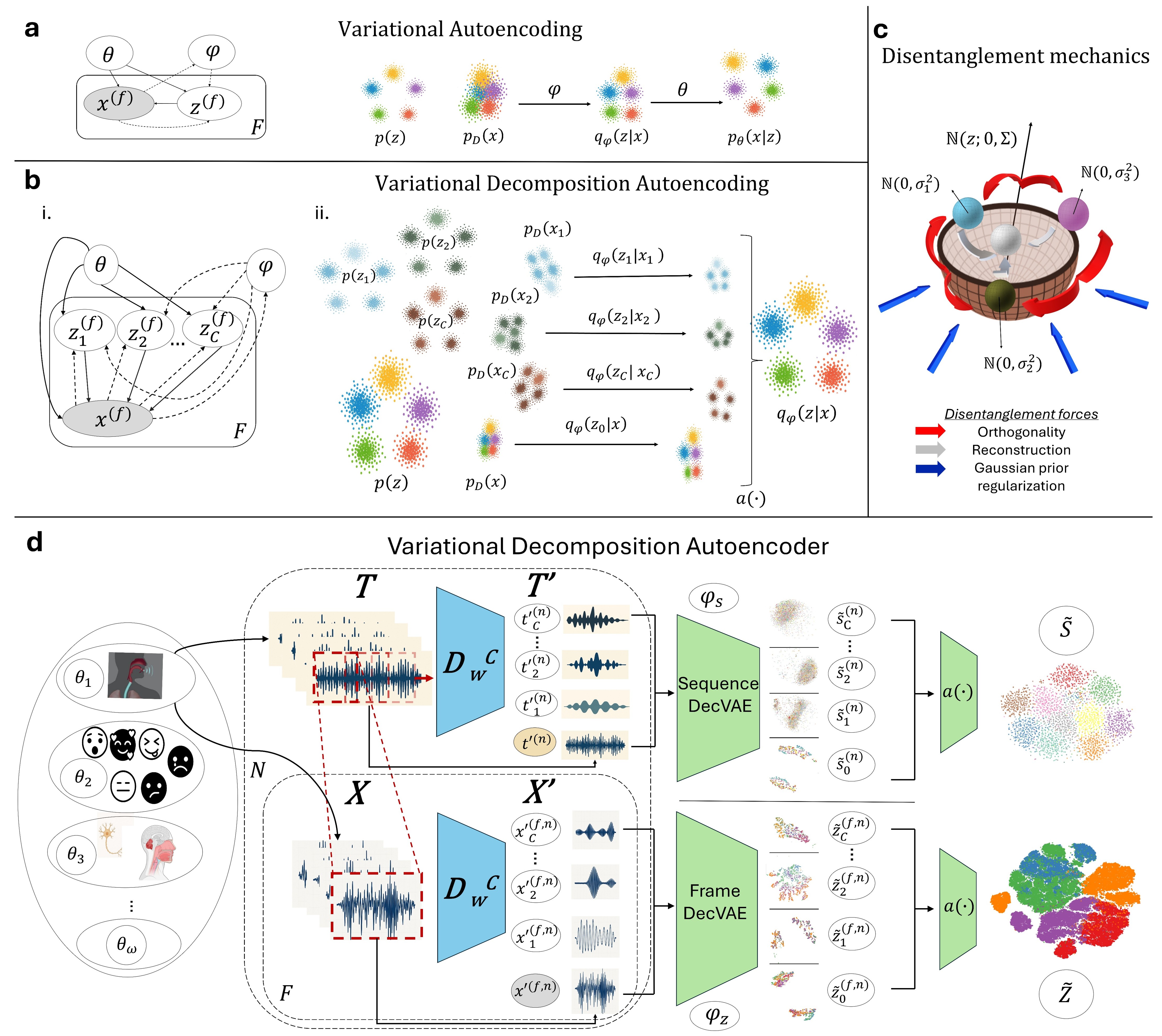}
    \caption[]{\footnotesize{\textsf{\textbf{$|$ Variational decomposition autoencoding (VDA) uses a neural network encoder to approximate latent distributions assuming a decomposition generative process and novel disentanglement mechanics.} \textbf{a)}, variational autoencoding: directed factor graph over the $F$ frames $x^{(f)}$ of a sequence, and illustration of inference ($\varphi$) and generation ($\theta$). \textbf{b)}, (i) directed factor graph for the proposed VDA; (ii) VDA assumes that each observation is a superposition of component signals in mutually orthogonal subspaces of the signal space, each generated by its own latent subspace $z_i$ with a Gaussian prior $p(z_i)$, so that $p(z)=\prod_i p(z_i)$. Under 
    this assumption the posterior factorizes over components 
    (Proposition~1, Methods): a single recognition model $q_\varphi$, 
    shared across components, maps each decomposed component $x_i$ to its 
    own latent subspace, $q_\varphi(z_i\,|\,x_i)$, and the original signal 
    $x$ to $z_0$, without conditioning between subspaces; an aggregation 
    function $a(\cdot)$ combines the subspaces. \textbf{c)}, besides the 
    Gaussian prior regularization, VDA imposes two structural biases that 
    come from the decomposition, reconstruction and orthogonality, and 
    together they shape a latent space that disentangles. DecVAE is 
    trained on the VDA objective, which combines a self-supervised loss 
    embodying these biases with Gaussian prior regularization of the 
    component subspaces (detailed in Methods). \textbf{d)}, the variational 
    decomposition autoencoder (DecVAE) can be applied across several domains $\theta_1,\dots,\theta_\omega$ (speech, 
    emotion, dysarthria severity). It can optionally learn latent 
    variables at two time scales: the decomposition model $\mathcal{D}^{C}_{w}$ is applied to the $N$ input sequences ($T$) and to 
    the $F$ windowed frames of each sequence ($X$); e.g., in speech 
    production ($\theta_1$) the long-term speaker identity variable $S$ 
    influences the short-term phonetic content $Z$. In this panel, 
    superscripts index the frame $f$ and the sequence $n$, subscripts 
    index the components, and primes denote decomposed signals. The 
    decomposed sequences $\{t'^{(n)}, t'^{(n)}_1, \dots, t'^{(n)}_C\}$ 
    (domain $T'$) and frames $\{x'^{(f,n)}, x'^{(f,n)}_1, \dots, 
    x'^{(f,n)}_C\}$ (domain $X'$) are analysed by the sequence and frame 
    branches of DecVAE, with parameters $\varphi_S$ and $\varphi_Z$, which 
    learn the latent subspaces $\{\tilde s^{(n)}_0, \dots, \tilde 
    s^{(n)}_C\}$ and $\{\tilde z^{(f,n)}_0, \dots, \tilde z^{(f,n)}_C\}$. 
    Each branch aggregates its subspaces through $a(\cdot)$ (concatenation 
    in our experiments) into sequence-level and frame-level 
    representations $\tilde S$ and $\tilde Z$ (detailed in Methods).}}}
    \label{fig1}
\end{figure*}

Observing and disentangling the surrounding environment into basic properties -such as the shape of an object or the timbre of a sound- is  a fundamental cognitive trait deeply embedded in intelligence hardwired into human brains since infancy \cite{fiser2010}. 
Disentangled representations are crucial for the interpretability, generalization, and control in neural networks, as they enable robust encoding of underlying generative factors while suppressing noise and domain-irrelevant variation. For instance, self-supervised speech representation learning \cite{vandenOordDeepMind2018RepresentationCoding}, healthy and dysarthric speech recognition \cite{Dubbioso2024VoiceControls}, emotional speech recognition \cite{chen2022disentangled}, time series prediction \cite{Isomura2021DimensionalityCapability,Woo2022CoST:Forecasting}, separation of venous effects from neural activity \cite{Kay2020AFMRI}, separation of cellular properties at the single-cell level \cite{CohenKalafut2023JointEmbedding}, are all problems where high-dimensional, nonstationary inputs often conflate multiple latent factors. To uncover the underlying generative factors of these processes without prior knowledge about the task, unsupervised machine learning models typically assume statistical independence among latent variables \cite{Ridgeway2016ARepresentation-Learning}; this is a foundational bias in the unsupervised model of variational autoencoders (VAEs) \cite{Kingma2014Auto-EncodingBayes}. VAEs encode independent generative factors into distinct dimensions of a latent space \cite{Rolinek2018VariationalAccident}, but this assumption often fails in complex time-evolving processes, where generative factors are numerous, interdependent, and not well-defined (Fig.~\ref{fig1}a). 

Many methods have explored structural biases as a mechanism to promote disentanglement in such settings and achieve separation or decomposition of physically meaningful time-localized representations that have distinct spectral signatures \cite{apostolidis2017swd}. This idea of decomposition is not new in signal processing, stirring an increasing exploration of inductive biases as structural interventions in unsupervised and self-supervised learning (SSL).
For instance, time-contrastive learning (TCL) \cite{Hyvarinen2016UnsupervisedICA} hypothesizes nonlinear mixtures and time-evolving distributions, as a bias to disentangle time series events. Neural basis expansion analysis for interpretable time series (N-BEATS) \cite{Oreshkin2019N-BEATS:Forecasting} introduces inductive bias layers with task-specific weights to disentangle time series by performing a hierarchical decomposition of the input layer by layer. Contrastive learning of seasonal-trend (CoST) representations \cite{Woo2022CoST:Forecasting} views disentanglement as a gradual decomposition of the generative factors into seasonality and trend components utilizing layers with built-in Fourier transforms. 
Hence, imposing structural biases on the architecture or the learning objective is a mechanism that promotes disentanglement \cite{Mathieu2018DisentanglingAutoencoders}. Nevertheless, while effective for task-specific performance, these methods are not explicitly designed to maximize disentanglement.  

In the context of variational inference, disentanglement emerges indirectly from the pressure to approximate an orthogonal latent space governed by a multivariate Gaussian prior \cite{Rolinek2018VariationalAccident}. Extensions such as factorized hierarchical VAE (FHVAE) \cite{Hsu2017UnsupervisedData} and multiple filtered latent VAE (MFL-VAE) \cite{Boulianne2020ALearning} introduce hierarchical time-scale modeling, while $\beta$-VAE \cite{Higgins2017Beta-VAE:Framework}, total correlation VAE (TCVAE) \cite{Chen2018IsolatingAutoencoders}, and FactorVAE \cite{kim2018factorVAE} explicitly regularize latent independence. 
However, these approaches lack structural priors on the generative process itself, limiting their applicability to complex nonstationary time-evolving processes with multiple factors.
Here, we introduce variational decomposition autoencoding (VDA), a representation learning framework that advances disentanglement by assuming that observations are generated as a superposition of orthogonal components, each driven by its own latent subspace (Fig.~\ref{fig1}b), and by building the inference model on the factorization that this assumption implies. Its neural implementation, i.e., the variational decomposition autoencoder (DecVAE), is designed to uncover interpretable factors of variation in complex, time-evolving signals through a structured decomposition of latent subspaces (Fig.~\ref{fig1}c,d, Supplementary Table~1
). Notably, DecVAEs are encoder-only networks, without a decoder, and they perform autoencoding at the level of distributions in a latent embedding space where the distribution of the original signal's embedding is reconstructed from those of its components (see Methods).

Unlike conventional VAEs, which assume independence among latent dimensions, VDA reformulates the generative model to incorporate frequency-resonant subspaces, i.e., latent subspaces tied to one frequency component of the input, that precede and construct the target prior distribution (Fig.~\ref{fig1}b). This architecture enables DecVAE to tie disentanglement explicitly to time-localized spectral dynamics, leveraging a signal decomposition model and deep encoder to approximate latent structure. To further enforce disentanglement, DecVAE integrates a self-supervised contrastive loss that guides latent components toward orthogonality and reconstructive fidelity, while maintaining Gaussian priors within each subspace (Fig.~\ref{fig1}c). This mechanism separates the embeddings of the components by construction (Methods, Proposition~2); whether the resulting subspaces align with generative factors depends on how well the decomposition matches the structures that carry them, which we test directly.

We evaluated DecVAEs on two synthetic benchmarks and three real-world speech datasets spanning speech recognition, dysarthria severity progression, and emotion classification. Across tasks, DecVAEs learned more disentangled and robust representations than VAE-based models, separated phoneme and speaker in zero-shot transfer to dysarthric speech, and yielded latent subspaces that are factor specific when the generative factors reside in different frequency bands. Compared with nine time-series and self-supervised baselines, DecVAEs learn the most robust representations on real speech, whereas self-supervised large audio models (LAMs) gave more informative ones. Our experiments on a synthetic benchmark with factors that are not spectrally localized, sets the limits of the utility of the decomposition prior of VDA. These findings position VDA as a robust and versatile framework for disentangling the structure of complex generative processes whose factors have distinct spectral signatures. To enable researchers to leverage and further develop VDA, we have made all R/Python code publicly available in an easy-to-use format.

\section*{Results}
\subsection*{Method Overview}
VDA leverages the DecVAE architecture to learn unsupervised deep representations with enhanced disentanglement and interpretability (Fig.~~\ref{fig1}d). The input to DecVAE consists of time series denoted by $ X $, which are decomposed into $C$ distinct components using a signal decomposition model $\mathcal{D}_w^C$. Each component captures localized time–frequency features, enabling the model to isolate spectral signatures that correspond to latent generative factors (Supplementary Fig. 2
). This decomposition introduces a strong inductive bias, which assists DecVAE in approximating the factors of variation when these are tied to distinct spectral structures (see `Where the decomposition prior helps, and where it does not'). DecVAE learns a structured latent space $\tilde{Z} $, where each subspace $\tilde{z}_i \in \tilde{Z} $ corresponds to a distinct component of the input (Fig.~~\ref{fig1}b). The generative assumption underlying VDA posits that the observed data $ x \sim p(x) $ arises from multiple latent subspaces $ z_i $, each tied to a decomposed input $x'^(i)$. DecVAE operationalizes this by encoding each $x'^(i)$ into its corresponding $ \tilde{z}_i$, a latent approximation of the true $ z_i $, and enforces distribution-level reconstruction between the embeddings of the original signal and of its components through the SSL contrastive task (Fig.~~\ref{fig1}c,d). This structure enables the model to separate generative factors that occupy different frequency regions, by aligning latent subspaces with time-localized frequency dynamics (Fig.~~\ref{fig1}b).

To promote disentanglement during training, DecVAE incorporates a self-supervised contrastive loss which separates the distributions of the components' embeddings from each other while reconstructing the original's distribution as their mixture (see Methods, Proposition 2), with a Gaussian prior regularizer on each latent subspace $\tilde{z}_i $ (Fig.~~\ref{fig1}d). These mechanics—orthogonalization, reconstruction dynamics, and subspace-specific prior regularization—collectively define the disentanglement-promoting structure of VDA. Furthermore, DecVAE can accommodate multi-scale temporal dynamics, enabling the simultaneous learning of multiple latent representations when factors of variation operate across distinct time scales (e.g. $X$, $T$ in Fig.~~\ref{fig1}d) (see details in \textit{Methods} section).

Following pre-training, DecVAE functions as a mapping from input \( X \) to a disentangled representation \( \tilde{Z} \), which can be evaluated across downstream tasks. We assess the utility of \( \tilde{Z} \) in domains such as speech recognition, dysarthria severity estimation, and emotional speech classification. DecVAE supports both zero-shot and fine-tuned evaluation, allowing it to be applied to new datasets with additional factors of variation, such as emotional state, with or without fine-tuning. Additionally, latent response analysis provides interpretability by revealing how individual subspaces encode specific generative factors (see details in the \textit{Applications} section).

\begin{figure*}[h!]
    \centering
    \includegraphics[width=0.99\textwidth,height=11cm]{figures_main/fig2_vowels.jpg}
    \caption[]{\footnotesize{\textsf{\textbf{$|$ Simulated speech data (SimVowels)}. \textbf{a)}, TSNE \cite{maaten2008visualizing} of development set frame-level vowel-colored inputs, (left): Mel filterbank features of original signal, (right): decomposed Mel filterbank features of the original and components signals, after aggregation. \textbf{b)}, TSNE of training set sequence-level speaker-colored inputs, (left): Mel filterbank features of original signal, (right): decomposed Mel filterbank features of the third component, after aggregation. \textbf{c)}, TSNE of latent spaces using \textbf{a} as inputs (left) DecAE with empirical wavelet transform (EWT) \cite{gilles2013ewt} frequency-resonant frame embedding, (middle) DecVAE with EWT \cite{gilles2013ewt} vowel-colored embedding, (right) VAE \cite{Kingma2014Auto-EncodingBayes} vowel-colored embedding. \textbf{d)},TSNE of latent spaces using \textbf{a,b} as inputs, (left) DecVAE with EWT frequency-resonant sequence embedding, (mid-left) DecVAE with EWT \cite{gilles2013ewt} speaker-colored sequence embedding, (mid-right) DecVAE with EWT \cite{gilles2013ewt} speaker-colored frame embedding, (right) VAE \cite{Kingma2014Auto-EncodingBayes} speaker-colored frame embedding. \textbf{e)}, TSNE projection of DecVAE with EWT \cite{gilles2013ewt} vowel-colored latent space (two mini-batches) along with a multivariate Gaussian distribution (sphere). \textbf{f)}, TSNE projection of VAE \cite{Kingma2014Auto-EncodingBayes} vowel-colored latent space (two mini-batches) along with a multivariate Gaussian distribution (sphere). \textbf{g)}, (left) disentanglement (x-axis) versus informativeness (y-axis) from DCI \cite{Eastwood2018ARepresentations} and robustness (circle size), (right) modularity (x-axis) versus explicitness (y-axis) \cite{Ridgeway2018LearningLoss} and robustness \cite{suter2019irs}(circle size). \textbf{h)}, speaker identification (x-axis) versus phoneme (vowel) recognition (y-axis) and disentanglement \cite{Eastwood2018ARepresentations} (circle size) performance of $\beta$-DecVAE variants ($\beta=0.1$) and VAE \cite{Kingma2014Auto-EncodingBayes}, $\beta$-VAE ($\beta=0.1$) \cite{Higgins2017Beta-VAE:Framework}, ICA \cite{hyvarinen2001independent}, PCA \cite{Greenacre2022PrincipalAnalysis}. In the frequency-resonant embeddings (\textbf{c}, \textbf{d}, left), each point is the latent embedding of a frame (\textbf{c}) or an utterance (\textbf{d}) of the original signal or of one of its components, colored by its source (original signal, component 1,2, or 3); the separated regions for the different sources show that each component occupies its own region of the latent space. In the rest of the TSNE panels, points are colored by generative factor named in the panel title, and compact well-separated clusters indicate that the factor is captured by the learned representation. The sphere in \textbf{e}, \textbf{f} is a standard multivariate Gaussian, shown as a reference for the prior. In \textbf{g}, \textbf{h}, models towards the upper right corner perform better on both axes, and the circle size (the bigger the better) encodes a third metric.}  }}
    \label{fig2}
\end{figure*}

\subsection*{Simulated speech data}
We first test DecVAE on the SimVowels simulated speech dataset (see 'Datasets'). The SimVowels data represents a generative speech process with two exclusively frequency-related factors, the phonetic content (vowel) and the speaker identity encoded into the three formants that generate each vowel (Supplementary Fig. 4
). We observe that DecVAE models are able to disentangle frequency components in the frame $Z$ (Fig.~\ref{fig2}c) and sequence $S$ (Fig.~\ref{fig2}d) latent spaces.
Using the decomposition model $D_w^C,C=3$ on the input $X$ acts as a good starting point for disentanglement (Fig.~\ref{fig2}a,b). DecVAE's optimization through the VDA objective (see 'Variational decomposition autoencoding objective') further promotes this structure resulting in smooth, well-separated frequency embeddings (Supplementary Fig. 5
).
Consequently, DecVAE can successfully approximate  the prior $p(z)$ in both frame and sequence branches (Fig.~\ref{fig2}a,b), leading to enhanced disentanglement performance for vowels (Fig.~\ref{fig2}c, middle) and speakers (Fig.~\ref{fig2}d, middle), compared to a VAE model without frequency sensitivity (Fig.~\ref{fig2}c,d, right). DecVAE constructs an elaborated latent space shaped by decomposition orthogonality, reconstruction and the prior approximation (Fig.~\ref{fig2}e), compared to the weaker structural bias of only prior approximation in VAE-based models  (Fig.~\ref{fig2}f) and other supporting methods (Supplementary Fig. 6
).  

To quantify disentanglement quality, we utilize several disentanglement metrics (see `Disentanglement Evaluation') to measure the representation properties of disentanglement, informativeness and total correlation. DecVAE representations are more disentangled and more informative according to two measurement frameworks, DCI \cite{Eastwood2018ARepresentations} (Fig.~\ref{fig2}g, left) and modularity-explicitness \cite{Ridgeway2018LearningLoss} (Fig.~\ref{fig2}g, right) outperforming VAE-based models and dimensionality reduction methods ICA and PCA (details and ablations in Supplementary Table~7
). DecVAE representations are robust to perturbations \cite{suter2019irs}, with higher IRS robustness (Fig.~\ref{fig2}g). The enhanced disentanglement performance of DecVAE models is translated into better task-related generalization in both vowel recognition and speaker identification accuracies, compared to VAE-based models, PCA and ICA (Fig.~\ref{fig2}h, Supplementary Tables~8,9
). We also examine the learned representations under a Gaussian total correlation and mutual information between dimensions perspective, with unsupervised metrics GCN \cite{Locatello2019ChallengingRepresentations} and MI. DecVAE models achieve very low MI and GCN scores slightly behind other methods despite the residuals of shared information that might exist in the aggregated latent in the components and original signal (Supplementary Fig. 7a, 
Supplementary Table~7
).  

Moreover, we examine the importance of the choice of decomposition model $D_w^C$ by comparing our custom-made filter decomposition (FD) with three different state-of-the-art methods, empirical mode decomposition (EMD) \cite{huang1998empirical}, variational mode decomposition (VMD) \cite{dragomiretskiy2014vmd}, EWT \cite{gilles2013ewt}. We find out that the choice of decomposition is very important for a frequency-resonant embedding that translates into good task-related performance (Supplementary Fig. 5,8
). The decomposition should successfully recover frequency-related events that are orthogonal and assign them to the correct component without mode mixing, i.e., information shared between different components. FD and EWT provide the best performance due to the nature of filters and wavelet transform that promotes time-frequency orthogonality, resulting in less correlated components in the input domain (Supplementary Fig. 9
). In VMD and EMD, input correlations propagate to the latent space and contaminate the generative factors separation (Supplementary Fig. 8
). This is naturally translated into worse performance in disentanglement and task-related metrics (Supplementary Tables~7,8,9
). The prior knowledge that SimVowels is generated using three formants informs the selection of $C=3$ for our main experiments; we test sensitivity to $C$ in subsequent sections (see `Where the decomposition prior helps, and where it does not.'). Similarly to $\beta$-VAEs~\cite{Higgins2017Beta-VAE:Framework}, the latent space of $\beta$-DecVAE is compressed as $\beta$ increases (Supplementary Fig.~11), but $\beta$-DecVAE is more robust to $\beta$ than $\beta$-VAE (see Supplementary Fig.~12); the effect of $\beta$ on the separation of latent component subspaces is analyzed in `Information allocation and shaping of latent subspaces' (Fig.~\ref{fig7}a).

\begin{figure*}[h!]
    \centering
    \includegraphics[width=0.99\textwidth,height = 11cm]{figures_main/fig3_timit.jpg}
    \caption[]{\footnotesize{\textsf{\textbf{$|$ Real speech data TIMIT \cite{garofolo1993timit}}. \textbf{a)}, TSNE \cite{maaten2008visualizing} of a subset (13 phonemes from 6 mini-batches) of the development set, frame-level phoneme-colored inputs, (left): Mel filterbank features of original signal, (right): decomposed Mel filterbank features of the original and components signals, after aggregation. \textbf{b)}, TSNE of a subset (10 speakers from 6 mini-batches) of the development set, (left): Mel filterbank features of original signal, (right): decomposed Mel filterbank features of aggregated components. \textbf{c)}, TSNE of latent spaces using \textbf{a} as inputs (6 mini-batches) (left) DecVAE with EWT \cite{gilles2013ewt} frequency-resonant frame embedding ($C=4$), (middle) DecVAE with EWT \cite{gilles2013ewt} phoneme-colored embedding, (right) VAE \cite{Kingma2014Auto-EncodingBayes} phoneme-colored embedding. \textbf{d)}, TSNE of latent spaces using \textbf{a,b} as inputs (6 mini-batches), (left) DecVAE with EWT \cite{gilles2013ewt} speaker-colored frame embedding, (right) VAE \cite{Kingma2014Auto-EncodingBayes} speaker-colored frame embedding. \textbf{e)}, TSNE projection of DecVAE with EWT \cite{gilles2013ewt} phoneme-colored (left) and speaker-colored (right) latent spaces (6 mini-batches) along with a multivariate Gaussian distribution (sphere). \textbf{f)}, TSNE projection of VAE \cite{Kingma2014Auto-EncodingBayes} phoneme-colored (left) and speaker-colored (right) latent spaces (6 mini-batches) along with a multivariate Gaussian distribution (sphere). \textbf{g)}, (left) disentanglement (x-axis) versus informativeness (y-axis) from DCI  \cite{Eastwood2018ARepresentations} and robustness \cite{suter2019irs} (circle size), (right) modularity (x-axis) versus explicitness (y-axis) \cite{Ridgeway2018LearningLoss} and robustness \cite{suter2019irs} (circle size). \textbf{h)}, disentanglement metrics scores of $\beta$-DecVAE variants, and VAE-based and eigenanalysis methods, VAE \cite{Kingma2014Auto-EncodingBayes}, $\beta$-VAE \cite{Higgins2017Beta-VAE:Framework}, ICA \cite{hyvarinen2001independent}, PCA \cite{Greenacre2022PrincipalAnalysis}. Results for a subsampled set of IEMOCAP (see `Speech emotion recognition fine-tuning evaluation'), for DecVAEs, VAE-based, time-series and self-supervised baselines are presented in Supplementary Tables~16-19. Panels in this figure are organized and read as in Fig.~\ref{fig2}.}}}
    \label{fig3}
\end{figure*}

\subsection*{Disentanglement of real speech data}
We apply DecVAE on real speech to disentangle phoneme from speaker identity in the TIMIT \cite{garofolo1993timit} dataset (Fig.~\ref{fig3}). We note that DecVAE extends the capability of learning a frequency-resonant embedding to real speech that contains other phonetic content beside vowels (Fig.~\ref{fig3}b, Supplementary Fig. 5
). The decomposition model $D_w^C,C=4$ provides a more separated input space for phonemes (Fig.~\ref{fig3}a) and speakers (Fig.~\ref{fig3}b). DecVAE learns smoother representations where phonemes (Fig.~\ref{fig3}c) and speakers (Fig.~\ref{fig3}d) are more disentangled, and with a more Gaussian latent distribution (Fig.~\ref{fig3}e) compared to VAE-based models \cite{Kingma2014Auto-EncodingBayes,Higgins2017Beta-VAE:Framework} (Fig.~\ref{fig3}f). Although all methods manage to recover an embedding where phonemes are well separated, DecVAEs demonstrate more aptitude in creating representations where both generative factors can be separated (Supplementary Fig. 13
).

In terms of disentanglement quality, DecVAE models with EWT \cite{gilles2013ewt} and FD display higher disentanglement, informativeness, robustness and explicitness, but reduced modularity compared to VAE-based models, PCA and ICA (Fig.~\ref{fig3}g; see Table~\ref{tab:baselines} for time-series and SSL baselines. PCA, ICA and VAE-based models have slightly less correlated latent dimensions than DecVAEs due to the residual correlations between input components in DecVAEs (Supplementary Fig. 7b
). Overall, DecVAE models learn more informative representations for the task of speech recognition (higher informativeness and explicitness), are more disentangled, and can encode relations between generative factors and latents more robustly, compared to VAE-based models, PCA and ICA (Fig.~\ref{fig3}h, Supplementary Table~10
). $\beta$-VAE \cite{Higgins2017Beta-VAE:Framework} needs fewer dimensions to encode speech data (higher completeness) (Fig.~\ref{fig3}h, Supplementary Table~10
). 
PCA \cite{Greenacre2022PrincipalAnalysis} learns the most uncorrelated latent space that is though not disentangled in terms of generative factors, whereas the raw Mel filterbank inputs are already quite informative for the speech recognition task \cite{Hsu2017UnsupervisedData,Boulianne2020ALearning} (Supplementary Table~10
). 

Increasing the number of components $C$ has a beneficial effect on all disentanglement metrics (Supplementary Fig. 14c
) as real speech has more complex, albeit similar, spectral structures to our simulations since consonants do not abide by a clear formant-based generation process like vowels \cite{Ladefoged2006APhonetics} (Supplementary Fig. 4
). This agrees with our simulations, where overestimating $C$ is not detrimental, as $C=4$ lies above the median of three spectral peaks selected in TIMIT per frame (Fig.~\ref{fig8}a). Similarly, we note that $\beta$-VAE models rely only on the structural bias of Gaussian prior approximation, whereas DecAEs ($\beta=0$) can also disentangle (Supplementary Fig. 14a
). Over $\beta \in \{0, 0.1, 1\}$ we find that disentanglement metrics of $\beta$-DecVAE change little on TIMIT (Supplementary Fig. 14a).

\begin{figure*}[h!]
    \centering
    \includegraphics[width=0.99\textwidth,height = 11cm]{figures_main/fig4_voc_als.jpg}
    \caption[]{\footnotesize{\textsf{\textbf{$|$ Zero-shot transfer to dysarthric speech data VOC-ALS \cite{Dubbioso2024VoiceControls}}. \textbf{a)}, TSNE \cite{maaten2008visualizing} of a subset (4 mini-batches) of the dataset, frame-level King's clinical stage \cite{Roche2012ASclerosis}-colored inputs, (left): Mel filterbank features of original signal, (right): decomposed Mel filterbank features of the original and components signals, after aggregation. \textbf{b)}, TSNE of a subset (4 mini-batches) of the dataset, frame-level phoneme-colored inputs (left): Mel filterbank features of original signal, (right): decomposed Mel filterbank features of the aggregated components. \textbf{c)}, TSNE of latent spaces using \textbf{a,b} as inputs and a pre-trained $\beta$-DecVAE on SimVowels ($\beta=0.1$) for zero-shot evaluation (4 mini-batches) (left) $\beta$-DecVAE with FD, frequency-resonant frame embedding ($C=4$), (middle-left) $\beta$-DecVAE with FD, King's clinical stage \cite{Roche2012ASclerosis}-colored embedding, (middle-right) $\beta$-DecVAE phoneme-colored embedding, (right) $\beta$-DecVAE speaker-colored embedding (for 10 speakers). \textbf{d)}, TSNE of latent spaces using \textbf{a,b} as inputs and a pre-trained on SimVowels $\beta$-VAE ($\beta=0.1$) for zero-shot evaluation (4 mini-batches), (left) $\beta$-VAE \cite{Higgins2017Beta-VAE:Framework} King's clinical stage \cite{Roche2012ASclerosis}-colored embedding, (middle) $\beta$-VAE \cite{Higgins2017Beta-VAE:Framework}phoneme-colored embedding, (right) $\beta$-VAE \cite{Higgins2017Beta-VAE:Framework} speaker-colored embedding (for 10 speakers). \textbf{e}, TSNE projection of $\beta$-DecVAE with FD, King's clinical stage \cite{Roche2012ASclerosis}-colored latent space (4 mini-batches) along with a multivariate Gaussian distribution (sphere). \textbf{f)}, TSNE projection of $\beta$-VAE \cite{Higgins2017Beta-VAE:Framework} King's clinical stage \cite{Roche2012ASclerosis}-colored latent space (4 mini-batches) along with a multivariate Gaussian distribution (sphere). \textbf{g)}, (left) King's clinical stage \cite{Roche2012ASclerosis} detection (x-axis) versus speaker identification (y-axis) and phoneme recognition (circle size), (right) disentanglement (x-axis) versus informativeness (y-axis) from DCI \cite{Eastwood2018ARepresentations} and robustness \cite{suter2019irs} (circle size). \textbf{h)}, total classification performance (accuracies and F1-macro with $95\%$ confidence intervals error bars) of $\beta$-DecVAE variants, and VAE-based and eigenanalysis methods VAE \cite{Kingma2014Auto-EncodingBayes}, $\beta$-VAE \cite{Higgins2017Beta-VAE:Framework}, ICA \cite{hyvarinen2001independent}, PCA \cite{Greenacre2022PrincipalAnalysis}, in predicting King's clinical stage, disease duration, phoneme recognition, speaker identification. King's stage and disease duration classification are evaluated with participants shared between folds, as is done for speaker and phoneme recognition. If participants are held out, all classifiers are near chance (see Supplementary Fig.~27, Supplementary Table~21). PCA and ICA are fitted on VOC-ALS without labels and serve as non-zero-shot references. Panels in this figure are organized and read as in Fig.~\ref{fig2}.}}}
    \label{fig4}
\end{figure*}

\subsection*{Zero-shot transfer to dysarthric speech}
To further test the capacity of DecVAE to disentangle more challenging biomedical data, we apply DecVAE as a zero-shot transfer learned module 
without any adaptation on dysarthric speech data \cite{Dubbioso2024VoiceControls} where we assume three generative factors, phoneme, speaker identity and amyotrophic lateral sclerosis (ALS) progression. Dysarthric speech mainly manifests through an altered fundamental frequency $f0$, which alters the entirety of the speech spectrum \cite{Dubbioso2024VoiceControls}. 
$\beta$-DecVAEs retain the frequency-resonant structure of the latent space even on unseen data distributions, and separate phoneme and speaker identity in the latent space (Fig.~\ref{fig4}c), which VAE-based models, PCA, and ICA fail to do, even though PCA and ICA are fitted on VOC-ALS and are not zero-shot (Fig.~\ref{fig4}d, Supplementary Fig.~15). King's clinical stage appears separated in the latent embeddings, yet this effect can be attributed to a confounding effect of the speaker variable on King's stage, as each participant is characterized by a single stage, and thus ALS King's staging cannot be distinguished from speaker identity in this dataset (see Supplementary Fig.~27). The strong structural biases of $\beta$-DecVAE allow disentanglement of more complex processes compared to healthy speech, providing an interpretable latent structure that cannot be learned with the prior approximation only (Fig.~\ref{fig4}c,d,e,f). 

DecVAEs demonstrate strong performance in phoneme recognition and speaker identification (Fig.~\ref{fig4}h), with more disentangled, informative and robust representations compared to VAE-based models, ICA and PCA (Fig.~\ref{fig4}g), with the latter two fitted on VOC-ALS, and therefore not zero-shot. However, their higher accuracy for King's stage and disease duration prediction, comes from the fact that disease staging and speaker identity are confounded in a small clinical dataset such as VOC-ALS, since each participant has a single stage. Our results on a speaker-independent evaluation (i.e. participants held out of the training folds) verifies the confounding effect of speaker identity on disease staging predictions (see Supplementary Table~21, Supplementary Fig.~27), where all evaluated models fall to chance or just above it. PCA~\cite{Greenacre2022PrincipalAnalysis} and ICA~\cite{hyvarinen2001independent}, fitted on VOC-ALS, learn the most uncorrelated latent spaces in terms of MI and GCN (Supplementary Table~11
), however DecVAEs display uncorrelated latents comparable to VAEs (Supplementary Fig.~7c
). DecVAEs outperform VAE-based models in the remaining disentanglement and task-specific metrics except completeness, while PCA, that is fitted on VOC-ALS, has slightly higher DCI disentanglement (Supplementary Tables~11,12
, Table~\ref{tab:baselines}b).

\begin{figure*}[t!]
    \centering
    \includegraphics[width=0.99\textwidth,height = 11cm]{figures_main/fig5_iemocap.jpg}
    \caption[]{\footnotesize{\textsf{\textbf{$|$ Fine-tuning disentanglement on emotional speech data IEMOCAP \cite{busso2008iemocap}}. \textbf{a)}, TSNE \cite{maaten2008visualizing} of a subset (6 mini-batches) of the dataset, frame-level categorical emotion-colored inputs, (left): Mel filterbank features of original signal, (right): decomposed Mel filterbank features of the original and components signals, after aggregation. \textbf{b)}, TSNE of a subset (6 mini-batches) of the dataset, frame-level speaker-colored inputs (left): Mel filterbank features of original signal, (right): decomposed Mel filterbank features of the aggregated components. \textbf{c)}, TSNE of latent spaces using \textbf{a} as inputs and a pre-trained DecVAE on SimVowels for fine-tuning evaluation (6 mini-batches) (left) DecVAE with FD, frequency-resonant frame embedding ($C=4$), (middle-left) DecVAE with FD, emotion-colored embedding, (middle-right) DecVAE speaker-colored embedding, (right) DecVAE phoneme-colored embedding (for 11 phonemes). \textbf{d)}, TSNE of latent spaces using \textbf{b} as inputs and a pre-trained on SimVowels $\beta$-VAE ($\beta=0.1$) for fine-tuning evaluation (6 mini-batches), (left) $\beta$-VAE \cite{Higgins2017Beta-VAE:Framework} emotion-colored embedding, (middle) $\beta$-VAE \cite{Higgins2017Beta-VAE:Framework}speaker-colored embedding, (right) $\beta$-VAE \cite{Higgins2017Beta-VAE:Framework} phoneme-colored embedding (for 11 phonemes). \textbf{e)}, TSNE projection of DecVAE with FD, emotion-colored latent space (6 mini-batches) along with a multivariate Gaussian distribution (sphere). \textbf{f)}, TSNE projection of $\beta$-VAE \cite{Higgins2017Beta-VAE:Framework} emotion-colored latent space (6 mini-batches) along with a multivariate Gaussian distribution (sphere). \textbf{g)}, (left) phoneme recognition (x-axis) versus speaker identification (y-axis) and emotion recognition (circle size) performances, (right) disentanglement (x-axis) versus informativeness (y-axis) from DCI \cite{Eastwood2018ARepresentations} and robustness \cite{suter2019irs} (circle size), for the full IEMOCAP protocol (Supplementary Tables~14,15). \textbf{h)}, total classification performance (weighted and unweighted accuracy, weighted F1 with $95\%$ confidence intervals error bars) of $\beta$-DecVAE variants, and VAE-based and eigenanalysis methods VAE \cite{Kingma2014Auto-EncodingBayes}, $\beta$-VAE \cite{Higgins2017Beta-VAE:Framework}, ICA \cite{hyvarinen2001independent}, PCA \cite{Greenacre2022PrincipalAnalysis}, in predicting phoneme, speaker and categorical emotion classes, for the full IEMOCAP protocol (Supplementary Tables~14,15). Panels in this figure are organized and read as in Fig.~\ref{fig2}.}}}
    \label{fig5}
\end{figure*}

\subsection*{Fine-tuning disentanglement on emotional speech data}
To test VDA in an even more challenging task, we apply DecVAEs on emotional speech data from IEMOCAP \cite{busso2008iemocap} in a transfer learning and fine-tuning scheme. As there is no direct correspondence of emotion in the time-frequency domain, this is an even more challenging task with three generative factors of emotion, phoneme and speaker. The input data and their decomposition does not offer any separation for emotion or phonetic content (Fig.~\ref{fig5}a,b). DecVAE can adapt to the new data distribution and learns a frequency-resonant embedding (Fig.~\ref{fig5}c), that is organized differently from the $\beta$-VAE embedding (Fig.~\ref{fig5}d, Supplementary Fig.~16). On frame-level representations, DecVAEs yield the most disentangled, informative and robust representations when compared with VAE-based, PCA, and ICA methods (Supplementary Table~16; the full-protocol evaluation of the core models in Fig.~\ref{fig5}g,h and Supplementary Tables 14,15 shows a similar pattern): DecAE with FD has the highest DCI disentanglement (0.237 compared to 0.182 for raw Mel features), whereas all DecVAE variants are more informative (0.414 to 0.423 against at most 0.403), and more robust (IRS 0.613 to 0.827, against at most 0.589); $\beta$-DecVAE with EWT is the most robust of all models, including the baselines of Table~\ref{tab:baselines}b. On the other hand, PCA and ICA learn the most uncorrelated latent spaces (lower MI, GCN) and have higher completeness and explicitness. However, the increased disentanglement performance compared to VAE-based methods, does not translate into better emotion recognition; DecVAE identifies speakers from single latent frames more accurately than all models except wav2vec2 embeddings, but recognises emotion from utterance-level representation less accurately than other models, including $\beta$-VAE (0.43 to 0.46 against 0.46 to 0.58, Supplementary Table~17). Holding out whole sessions instead of single speakers at IEMOCAP cross-validation evaluation lowers accuracy by at most 0.021 (SFA) and at most 0.014 for DecVAE (Supplementary Table~18). Class-wise performance and confusion matrices for IEMOCAP when training on a subsample (see`Speech emotion recognition fine-tuning evaluation') are given in Supplementary Tables~18,19 and Supplementary Fig.~28; the sad emotion is recognized best by all models except wav2vec2 (which recognise angry best). The most frequent confusions are sad with neutral, and in DecVAE, $\beta$-VAE and FHVAE, angry with happy.

\subsection*{Comparison with time-series and self-supervised baselines}
In the above results, DecVAE has been compared with a core set of VAE-based, PCA and ICA baselines that share theoretical and intuitive functionality to DecVAE. To place the performance of DecVAE under a broader state-of-the-art perspective, we compare DecVAE with nine representation learning baselines from four different model families (Table~\ref{tab:baselines}, Supplementary Tables~7-20): frozen self-supervised pre-trained large audio models (LAMs) (wav2vec2.0 \cite{Baevski2020Wav2vecRepresentations}, WavLM \cite{chen2022wavlm}, HuBERT \cite{Hsu2021HuBERT:Units}), time-series self-supervised and contrastive models (CPC \cite{vandenOordDeepMind2018RepresentationCoding}, CoST \cite{Woo2022CoST:Forecasting}, TCL \cite{Hyvarinen2016UnsupervisedICA}, TF-C \cite{Zhang2022Self-SupervisedConsistency}, a dual-timescale sequential VAE (FHVAE \cite{Hsu2017UnsupervisedData}) and slow feature analysis (SFA \cite{wiskott2002slow}). We pre-trained all baseline models, except LAMs, in an unsupervised (SFA, FHVAE) and self-supervised (CPC, CoST, TCL, TF-C) manner, on the training set of each dataset; per-method specifics defined the input type of features, for instance, LAMs and TF-C require a waveform input, whereas the rest of the methods were trained with the same Mel filterbank input features as DecVAE. Moreover, the same metrics and classifiers as DecVAE were used to evaluate the baselines (see `Baselines', Supplementary Section~11.8 for details). Notably, the frozen LAMs were pretrained on large speech corpora (960 hours of English read speech for wav2vec2.0 \cite{Baevski2020Wav2vecRepresentations} and HuBERT \cite{Hsu2021HuBERT:Units}, 94,000 hours for WavLM \cite{chen2022wavlm}). As an increased number of latent dimensions can inflate disentanglement when read through some metrics (see Supplementary Fig.~29 and `Metric Sensitivity'), for LAMs we adopt a dual evaluation plan: evaluating once their full representations from the last transformer layer (768-dimensional embeddings) and once by fitting a PCA on the same embeddings to reduce them to the dimensionality of the DecVAE representation.

Table~\ref{tab:baselines} summarizes the comparison of DecVAE to all evaluated baselines. The clearest advantage of DecVAE over baselines is its representational robustness to confounding factors, as evident by evaluation using IRS \cite{suter2019irs} on TIMIT (IRS of 0.875 against 0.669 for the best performing alternative), on the correlated test set of SimCoupled (0.526 against 0.485), and on SimVowels, where DecVAE comes second after TCL (0.600 against 0.632). In DCI \cite{Eastwood2018ARepresentations} disentanglement DecVAE using EWT \cite{gilles2013ewt} outperforms other baselines trained on TIMIT (DCI disentanglement 0.256) and is second among the models trained on SimVowels after CoST~\cite{Woo2022CoST:Forecasting} (0.582 against 0.817); on SimCoupled, it only trails SFA~\cite{wiskott2002slow} (0.361 against 0.386). These comparisons exclude LAMs, which reach higher disentanglement and informativeness on TIMIT, both as released and after PCA reduction, and on SimVowels with full dimensionality. In DCI informativeness, SFA~\cite{wiskott2002slow}, FHVAE~\cite{Hsu2017UnsupervisedData}, wav2vec2~\cite{Baevski2020Wav2vecRepresentations} and HuBERT~\cite{Hsu2021HuBERT:Units} learn representations that are more informative on SimVowels, SFA~\cite{wiskott2002slow}, FHVAE~\cite{Hsu2017UnsupervisedData}, wav2vec2~\cite{Baevski2020Wav2vecRepresentations} on SimCoupled, and FHVAE~\cite{Hsu2017UnsupervisedData} on TIMIT; the same models identify SimVowels speakers more accurately than DecVAE (see Supplementary Tables~8,9). When evaluating representations in an unsupervised manner, DecVAE has a less correlated representation (GCN 0.15 to 0.17) than the raw LAM embeddings and most time-series baselines, but more correlated than PCA~\cite{Greenacre2022PrincipalAnalysis}, PCA-reduced LAM embeddings, and on SimVowels and TIMIT, $\beta$-VAE~\cite{Higgins2017Beta-VAE:Framework}; the dimensions in these two models are de-correlated by structure of the assumed prior.

To test whether the ranking of models in Table~\ref{tab:baselines} holds under different evaluation conditions, we re-scored seven SimVowels representations under the interventions of factor imbalance, correlated factors, label noise, coarsened speaker labels and added noise dimensions, leaving the latents unchanged (see `Metric Sensitivity' and Supplementary Section~11.5). Using the core set of metrics reported in Table~\ref{tab:baselines} preserves the rankings of models (Kendall's $\tau$ ranges from 0.60 to 0.95 across interventions against the reference, see Supplementary Figs.~29,30): $\beta$-DecVAE with FD ranks first in eight of ten interventions, and both DecVAE variants in Table~\ref{tab:baselines} are among the first three models. These rankings depend on the metrics averaged, and hence averaging over all disentanglement metrics places $\beta$-DecVAE with EWT third, and $\beta$-DecVAE with FD sixth (see Supplementary Fig.~30). Finally, the reported metrics and rankings are influenced by the dimensionality of a representation, with chance levels that vary by metric and latent dimensionality (see Supplementary Fig.~31). For instance, DCI disentanglement grows with latent dimensionality even for a random representation, whereas IRS and modularity have high chance levels that some baselines do not clear (see Supplementary Fig.~31).

On IEMOCAP (see Table~\ref{tab:baselines}b), where baselines models are pre-trained on SimVowels and fine-tuned on IEMOCAP (the LAMs are frozen), $\beta$-DecVAE is characterized by the most robust representations as per IRS~\cite{suter2019irs} (IRS 0.827 compared to 0.617 from second best SFA~\cite{wiskott2002slow}). Reducing the wav2vec2~\cite{Baevski2020Wav2vecRepresentations} embeddings through PCA yields the most disentangled and informative embeddings on IEMOCAP (DCI~\cite{Eastwood2018ARepresentations} D 0.297, DCI I 0.500), whereas DecVAE is close to SFA~\cite{wiskott2002slow}, CoST~\cite{Woo2022CoST:Forecasting} and FHVAE~\cite{Hsu2017UnsupervisedData} in terms of disentanglement and informativeness (DCI~\cite{Eastwood2018ARepresentations} D 0.17 to 0.21, DCI I 0.38 to 0.43). In IEMOCAP frame-level speaker identification, the best DecVAE variant outperforms other baselines except wav2vec2~\cite{Baevski2020Wav2vecRepresentations}, but recognizes emotion less accurately as a standalone classification task, compared to other baselines (see Supplementary Tables~16-19, Supplementary Fig.~28).

On VOC-ALS (Table~\ref{tab:baselines}b), the models pretrained on SimVowels are evaluated without any adaptation, whereas PCA, SFA and the PCA of wav2vec2 embeddings, are fitted on VOC-ALS and are therefore not zero-shot. Among the zero-shot models, $\beta$-DecVAE learns the most disentangled, informative and robust representations (DCI D 0.115 to 0.125, DCI I 0.679 and IRS 0.834 with FD, against 0.106 and 0.548 for FHVAE \cite{Hsu2017UnsupervisedData}, and 0.481 for $\beta$-VAE \cite{Greenacre2022PrincipalAnalysis}, the best zero-shot alternatives). Disentanglement is low for all models (DCI D 0.07 to 0.13), with PCA and SFA that were adapted to target data, being slightly ahead of DecVAE in disentanglement (0.130 (PCA) and 0.146 (SFA) against 0.115 to 0.125), but less informative and robust (DCI I at most 0.611, IRS at most 0.654). The DecVAE latents are less correlated than those of CPC \cite{vandenOordDeepMind2018RepresentationCoding} and FHVAE (GCN 0.14 against 0.71 and 7.7). 
CoST \cite{Woo2022CoST:Forecasting} is not evaluated, because its Fourier layer must be adapted to each dataset, which rules out zero-shot transfer. With speakers shared between training and test, King's stage is predicted well above chance (weighted accuracy up to 0.86 for DecVAE and 0.94 for wav2vec2 \cite{Baevski2020Wav2vecRepresentations} + PCA, which is fitted on VOC-ALS; Supplementary Table~12), but with participants held out every model falls to or below the majority-class level for both King's stage and disease duration (0.27 to 0.34, against 0.34; Supplementary Table~13), confirming that these predictions are confounded by speaker identity in this dataset (Supplementary Table~21, Supplementary Fig.~27). Phoneme recognition stays above chance under this protocol, but DecVAE (0.51 to 0.53) falls behind zero-shot FHVAE (0.73 to 0.76), and behind PCA (0.68) and wav2vec2 + PCA (0.77) that were fitted on VOC-ALS (Supplementary Table~13).

Overall, we observe that the decomposition structural bias of VDA allows the learning of the most robust representations (highest IRS) on real speech and the SimCoupled synthetic dataset (see `SimCoupled'), and disentanglement comparable to or above the time-series baselines; however, the informativeness of VDA representations is not comparable to self-supervised LAMs that are pre-trained on far more speech hours. In addition, several of these representation learning baselines draw on information beyond the current frame, whereas DecVAE’s frame-level embeddings are computed from a single frame. SFA is fitted on the whole utterance, while TCL and CPC are pre-trained on self-supervised tasks that predict the temporal position of a frame or the latent embeddings of other frames; FHVAE, in turn, models the frame-level latent jointly with the utterance-level latent. These mechanisms could inflate the frame-level results of these baselines. We discuss the implications of these results in `Discussion'.

\newsavebox{\TabOneA}\newsavebox{\TabOneB}
\begin{table}[ht!]
\centering
\caption{\textbf{Comparison with time-series and self-supervised baselines.} Frame-level DCI disentanglement (D) and informativeness (I) [23], interventional robustness score (IRS) [25] and Gaussian correlation normalized (GCN) [28]; bold values with superscripts $^{1}$, $^{2}$, $^{3}$ denote the three best values for each metric and dataset (lowest for GCN, highest otherwise); on VOC-ALS, only models evaluated zero-shot are ranked. \textbf{a}, Models trained on each dataset (SimCoupled: correlated test set). \textbf{b}, Models pretrained on SimVowels, fine-tuned on IEMOCAP and evaluated zero-shot on VOC-ALS; PCA, SFA and the PCA of wav2vec2 are fitted on each dataset without labels. DecVAE: $\beta = 0.1$, aggregated representation (X + OCs). $^{a}$PCA to the dimensionality of the DecVAE representation. $^{b}$PCA to 240 dimensions, the dimensionality of the transferred DecVAE representation. $^{c}$Not transferable zero-shot: the Fourier layer of CoST\cite{Woo2022CoST:Forecasting} must be adapted to each dataset. $^{d}$On VOC-ALS, fitted on the target data and therefore not zero-shot; shown for reference and not ranked. Evaluations on IEMOCAP, VOC-ALS, and SimCoupled carry a few best performing baselines from SimVowels and TIMIT. The PCA reduced version of wav2vec2 (as the best performing LAM on SimVowels and TIMIT) is used on IEMOCAP and VOC-ALS, for efficiency. n/a: not evaluated. Protocol in Methods; all metrics and confidence intervals in Supplementary Tables~7,10,11,16,20.}
\label{tab:baselines}
\footnotesize
\setlength{\tabcolsep}{2.5pt}
\sbox{\TabOneA}{%
\begin{tabular}{@{}lcccccccccccc@{}}
\multicolumn{13}{@{}l}{\textbf{a}} \\[2pt]
\toprule
 & \multicolumn{4}{c}{SimVowels} & \multicolumn{4}{c}{TIMIT~\cite{garofolo1993timit}} & \multicolumn{4}{c}{SimCoupled} \\
\cmidrule(lr){2-5}\cmidrule(lr){6-9}\cmidrule(l){10-13}
Model & D $\uparrow$ & I $\uparrow$ & IRS $\uparrow$ & GCN $\downarrow$ & D $\uparrow$ & I $\uparrow$ & IRS $\uparrow$ & GCN $\downarrow$ & D $\uparrow$ & I $\uparrow$ & IRS $\uparrow$ & GCN $\downarrow$ \\
\midrule
\multicolumn{13}{@{}l}{\textit{DecVAE}} \\
$\beta$-DecVAE + FD & 0.526 & 0.809 & \textbf{0.600}$^{2}$ & 0.173 & 0.172 & 0.394 & \textbf{0.872}$^{2}$ & 0.148 & 0.225 & 0.786 & \textbf{0.517}$^{2}$ & \textbf{0.156}$^{3}$ \\
$\beta$-DecVAE + EWT \cite{gilles2013ewt} & 0.582 & 0.742 & 0.523 & 0.147 & 0.256 & 0.376 & \textbf{0.875}$^{1}$ & 0.155 & \textbf{0.361}$^{3}$ & 0.810 & \textbf{0.526}$^{1}$ & 0.171 \\
\multicolumn{13}{@{}l}{\textit{VAE family and linear}} \\
$\beta$-VAE \cite{Higgins2017Beta-VAE:Framework} & 0.062 & 0.805 & \textbf{0.552}$^{3}$ & 0.061 & 0.202 & 0.184 & \textbf{0.669}$^{3}$ & 0.074 & 0.037 & 0.585 & 0.456 & 0.242 \\
PCA \cite{Greenacre2022PrincipalAnalysis} & 0.230 & 0.801 & 0.477 & \textbf{0.035}$^{2}$ & 0.229 & 0.258 & 0.551 & \textbf{$<$0.001}$^{1}$ & 0.252 & \textbf{0.896}$^{2}$ & \textbf{0.485}$^{3}$ & \textbf{$<$0.001}$^{1}$ \\
\multicolumn{13}{@{}l}{\textit{Self-supervised speech models (frozen)}} \\
wav2vec2 (960h LS) \cite{Baevski2020Wav2vecRepresentations} & \textbf{0.892}$^{1}$ & \textbf{0.944}$^{2}$ & 0.469 & 0.552 & \textbf{0.429}$^{3}$ & \textbf{0.626}$^{2}$ & 0.503 & 0.503 & \textbf{0.475}$^{1}$ & 0.824 & 0.409 & 0.476 \\
wav2vec2 (960h LS) \cite{Baevski2020Wav2vecRepresentations} + PCA  $^{a}$ & 0.383 & \textbf{0.891}$^{3}$ & 0.480 & \textbf{0.035}$^{2}$ & \textbf{0.464}$^{1}$ & \textbf{0.660}$^{1}$ & 0.602 & 0.014 & 0.174 & 0.807 & 0.471 & \textbf{0.022}$^{2}$ \\
WavLM (94,000h) \cite{chen2022wavlm} & 0.794 & 0.795 & 0.434 & 0.519 & 0.363 & 0.436 & 0.433 & 0.248 & \multicolumn{4}{c}{n/a} \\
WavLM (94,000h) \cite{chen2022wavlm} + PCA $^{a}$ & 0.341 & 0.781 & 0.480 & \textbf{0.010}$^{1}$ & 0.414 & 0.459 & 0.529 & \textbf{0.004}$^{2}$ & \multicolumn{4}{c}{n/a} \\
HuBERT (960h LS) \cite{Hsu2021HuBERT:Units} & \textbf{0.842}$^{2}$ & 0.871 & 0.454 & 0.765 & 0.351 & 0.441 & 0.448 & 0.395 & \multicolumn{4}{c}{n/a} \\
HuBERT (960h LS) \cite{Hsu2021HuBERT:Units} + PCA $^{a}$ & 0.399 & 0.838 & 0.478 & 0.055 & \textbf{0.433}$^{2}$ & \textbf{0.472}$^{3}$ & 0.511 & \textbf{0.010}$^{3}$ & \multicolumn{4}{c}{n/a} \\
\multicolumn{13}{@{}l}{\textit{Time-series self-supervised and contrastive}} \\
CPC \cite{vandenOordDeepMind2018RepresentationCoding} & 0.333 & 0.584 & 0.133 & 1.794 & 0.177 & 0.301 & 0.005 & 0.594 & \multicolumn{4}{c}{n/a} \\
CoST \cite{Woo2022CoST:Forecasting} & \textbf{0.817}$^{3}$ & 0.763 & 0.475 & 216 & 0.203 & 0.241 & 0.480 & 0.204 & 0.149 & 0.606 & 0.381 & 0.736 \\
TCL \cite{Hyvarinen2016UnsupervisedICA} & 0.137 & 0.538 & \textbf{0.632}$^{1}$ & 0.866 & 0.135 & 0.179 & 0.608 & 0.723 & \multicolumn{4}{c}{n/a} \\
TF-C \cite{Zhang2022Self-SupervisedConsistency} & 0.219 & 0.608 & 0.475 & 0.155 & 0.064 & 0.186 & 0.479 & 0.181 & \multicolumn{4}{c}{n/a} \\
\multicolumn{13}{@{}l}{\textit{Sequential VAE and slow features}} \\
FHVAE \cite{Hsu2017UnsupervisedData} & 0.379 & 0.885 & 0.468 & 0.513 & 0.205 & 0.433 & 0.479 & 0.248 & 0.225 & \textbf{0.864}$^{3}$ & 0.387 & 2.143 \\
SFA \cite{wiskott2002slow} & 0.425 & \textbf{0.945}$^{1}$ & 0.485 & 1.110 & 0.206 & 0.366 & 0.549 & 0.205 & \textbf{0.386}$^{2}$ & \textbf{0.907}$^{1}$ & 0.427 & 0.346 \\
\bottomrule
\end{tabular}}
\sbox{\TabOneB}{%
\begin{tabular}{@{}lcccccccc@{}}
\multicolumn{9}{@{}l}{\textbf{b}} \\[2pt]
\toprule
 & \multicolumn{4}{c}{IEMOCAP~\cite{busso2008iemocap} (fine-tuned)} & \multicolumn{4}{c}{VOC-ALS~\cite{Dubbioso2024VoiceControls} (zero-shot)} \\
\cmidrule(lr){2-5}\cmidrule(l){6-9}
Model & D $\uparrow$ & I $\uparrow$ & IRS $\uparrow$ & GCN $\downarrow$ & D $\uparrow$ & I $\uparrow$ & IRS $\uparrow$ & GCN $\downarrow$ \\
\midrule
\multicolumn{9}{@{}l}{\textit{DecVAE}} \\
$\beta$-DecVAE + FD & 0.168 & \textbf{0.423}$^{3}$ & \textbf{0.711}$^{2}$ & 0.115 & \textbf{0.115}$^{2}$ & \textbf{0.679}$^{1}$ & \textbf{0.834}$^{1}$ & \textbf{0.144}$^{2}$ \\
$\beta$-DecVAE + EWT \cite{gilles2013ewt} & 0.191 & 0.416 & \textbf{0.827}$^{1}$ & \textbf{0.052}$^{3}$ & \textbf{0.125}$^{1}$ & \textbf{0.675}$^{2}$ & \textbf{0.654}$^{2}$ & \textbf{0.140}$^{1}$ \\
\multicolumn{9}{@{}l}{\textit{VAE family and linear}} \\
$\beta$-VAE \cite{Higgins2017Beta-VAE:Framework} & 0.075 & 0.336 & 0.526 & 0.092 & 0.074 & 0.365 & \textbf{0.481}$^{3}$ & \textbf{0.219}$^{3}$ \\
PCA \cite{Greenacre2022PrincipalAnalysis}$^{d}$ & 0.109 & 0.400 & 0.570 & \textbf{$<$0.001}$^{1}$ & 0.130 & 0.434 & 0.620 & $<$0.001 \\
\multicolumn{9}{@{}l}{\textit{Self-supervised speech models (frozen)}} \\
wav2vec2 (960h LS) \cite{Baevski2020Wav2vecRepresentations} + PCA $^{b,d}$ & \textbf{0.297}$^{1}$ & \textbf{0.500}$^{1}$ & 0.542 & \textbf{0.013}$^{2}$ & 0.208 & 0.797 & 0.523 & 0.001 \\
\multicolumn{9}{@{}l}{\textit{Time-series self-supervised and contrastive}} \\
CPC \cite{vandenOordDeepMind2018RepresentationCoding} & 0.087 & 0.359 & 0.154 & 0.616 & 0.104 & 0.284 & 0.027 & 0.705 \\
CoST \cite{Woo2022CoST:Forecasting} & \textbf{0.199}$^{3}$ & 0.380 & 0.518 & 0.400 & \multicolumn{4}{c}{n/a$^{c}$} \\
\multicolumn{9}{@{}l}{\textit{Sequential VAE and slow features}} \\
FHVAE \cite{Hsu2017UnsupervisedData} & 0.176 & \textbf{0.423}$^{3}$ & 0.513 & 0.760 & \textbf{0.106}$^{3}$ & \textbf{0.548}$^{3}$ & 0.439 & 7.713 \\
SFA \cite{wiskott2002slow}$^{d}$ & \textbf{0.207}$^{2}$ & \textbf{0.425}$^{2}$ & \textbf{0.617}$^{3}$ & 0.114 & 0.146 & 0.611 & 0.654 & $<$0.001 \\
\bottomrule
\end{tabular}}
\edef\TabOneScale{\fpeval{min(1, \textwidth/\wd\TabOneA)}}%
\scalebox{\TabOneScale}{\usebox{\TabOneA}}

\medskip
\scalebox{\TabOneScale}{\usebox{\TabOneB}}
\end{table}

\subsection*{Interpretability of disentangled latent spaces}
\begin{figure*}[t!]
    \centering

    \begin{subfigure}{0.89\textwidth}
        \includegraphics[width=\linewidth,height=10.5cm]{figures_main/fig6_latent_responses.jpg}
        \label{fig:6a}
    \end{subfigure}

    \begin{subfigure}{0.89\textwidth}
        \includegraphics[width=\linewidth,height=7cm]{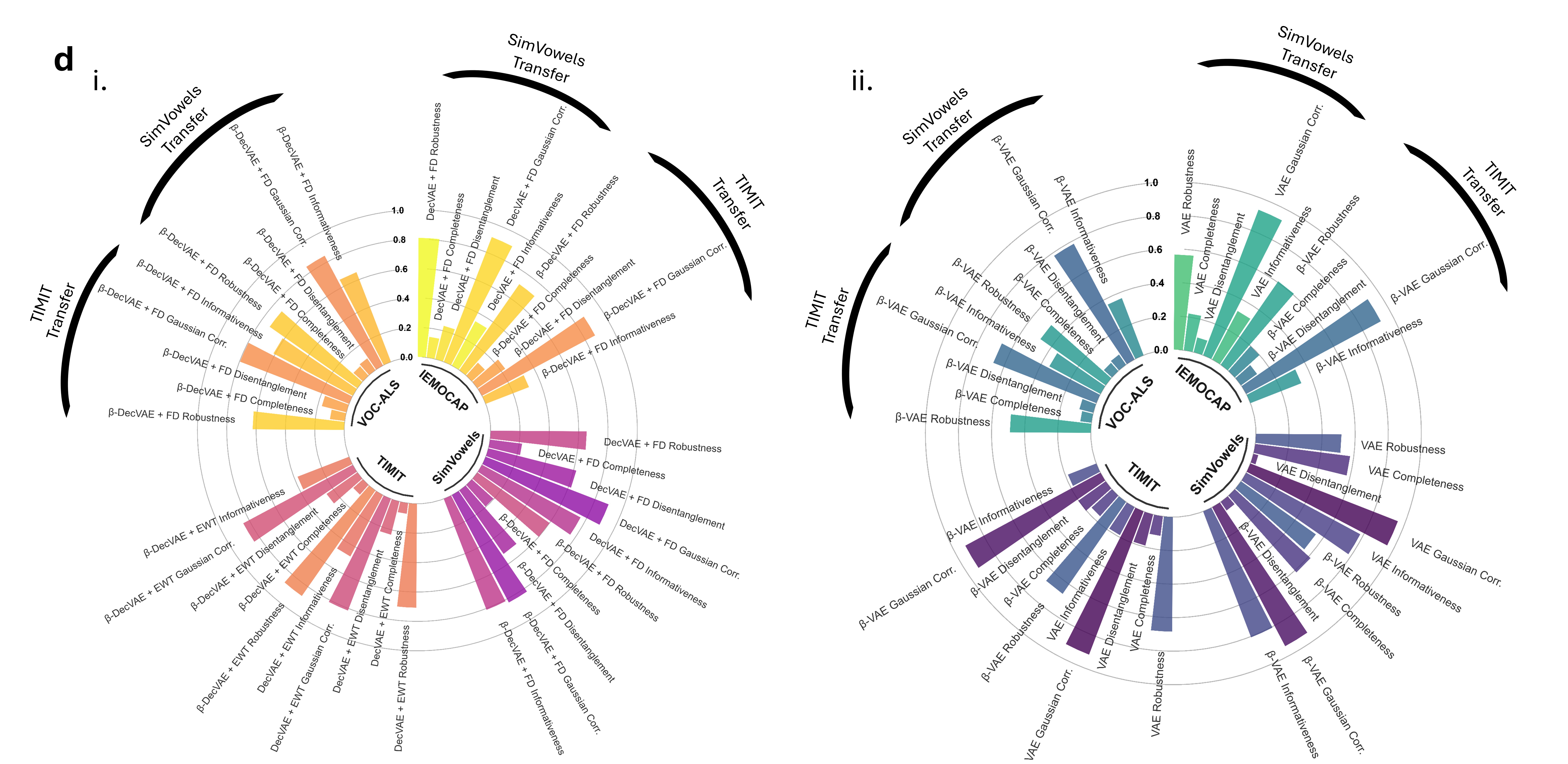}
        \label{fig:6b}
    \end{subfigure}

    \caption{\footnotesize{\textsf{\textbf{$|$ Latent response interpretability analysis and overall performance}. \textbf{a)}, latent responses on TIMIT~\cite{garofolo1993timit} for the DecVAE + EWT model; rows are the latent subspaces of the original signal (X) and of the components (OC1 to OC4), columns are the four dimensions of each subspace with the largest response variance and two with the smallest (`unused'), and each line is one phoneme, drawn as the speaker varies along the horizontal axis. \textbf{b)}, latent responses on VOC-ALS~\cite{Dubbioso2024VoiceControls} for the $\beta$-DecVAE ($\beta=0.1$) + EWT model, laid out as in \textbf{a}; each line is one King's clinical stage, traced as the phoneme varies. King's stage is constant within each participant and therefore confounded with speaker identity (Supplementary Section~11.4. \textbf{c)}, latent responses on IEMOCAP~\cite{busso2008iemocap} for the DecVAE + EWT model (first five columns, one per subspace) and the $\beta$-VAE ($\beta=0.1$) (last column), both transferred from TIMIT~\cite{garofolo1993timit}; for each emotion (rows), the surface shows the response of the most informative latent dimension over the phoneme-speaker plane. DecVAE spreads the responses over its subspaces, each covering a different range of the factors' values, 
    whereas the $\beta$-VAE concentrates them in a single latent space. \textbf{d)}, overall disentanglement performance of (i) $\beta$-DecVAE and (ii) $\beta$-VAE models, across all utilized datasets; DecVAEs disentangle better than $\beta$-VAEs on most metrics in all four datasets.}}}
    \label{fig6}
\end{figure*}

To shed light upon the way DecVAEs encode and disentangle information in their latent spaces, we perform a latent response analysis; since DecVAE has no decoder, instead of decoding traversed latent values we vary single generative factors in the input and visualize the responses of the latent dimensions (see 'Applications'). In our simulations, we compare $\beta$-VAEs with our DecVAEs and various decomposition methods (Supplementary Fig. 18
). The choice of decomposition affects how well DecVAE can encode different vowel and speaker combinations in each latent space. We see that DecVAE splits the information across the different latent subspaces of oscillatory components, allowing for a more efficient allocation of information (Supplementary Fig. 18
), whereas VAE models compress all the information in a few dimensions (higher compactness). We observe similar latent patterns for real speech data (Fig.~\ref{fig6}a, Supplementary Fig.~19
); the separation of speakers and phonemes across DecVAE subspaces is more prominent here than in VAEs and $\beta$-VAEs, whose responses concentrate in few dimensions. We also notice greater separability in some components than others; this qualitative assessment of responses is quantified in the next section.

Similarly, the zero-shot evaluation on dysarthric speech shows that DecVAE can efficiently allocate information in different subspaces to create disentangled and separable embeddings for phoneme and speaker, whereas King's stage, which is constant within each participant, cannot be disentangled from speaker identity in this dataset (Fig.~\ref{fig6}b, Supplementary Fig. 20
, Supplementary Section~11.4).
Since King’s clinical staging reflects a fixed classification of disease severity for each individual \cite{Roche2012ASclerosis}, we treat it as a static reference point and focus our visualization on how phonetic content varies across these clinical stages. This approach allows us to explore how specific acoustic or articulatory features align with progressive motor decline, without implying intra-speaker variability over time.

In the emotional speech case where we have three varying factors, we isolate emotions and visualize the contour of the most informative latent dimension for the phoneme-speaker response (Fig.~\ref{fig6}c, Supplementary Fig. 21
). DecVAEs again allocate information in a complementary fashion, with each oscillatory component latent subspace covering different ranges of the generative factors' values. $\beta$-VAEs on the other hand, encode information by using a single latent space, thus being limited in the encoding optimization that they can achieve (Fig.~\ref{fig6}c last column).

\begin{figure*}[h!]
    \centering
    \includegraphics[width=0.95\textwidth,height = 13cm]{figures_main/fig_7.png}
    \caption[]{\footnotesize{\textsf{\textbf{$|$ How the VDA objective shapes the latent subspaces on SimVowels.} \textbf{a)}, Separation of the components (mean JS divergence between the normalized embeddings of pairs of components, warm colours) and their divergence from the original (cold colours) against $\beta$, FD and EWT, sequence branch, test set; dotted line, maximum separation $\ln 2$; dashed line, divergence from the original at the optimum, $j_3 = 0.318$ (Proposition 2). Below: speaker (circles) and vowel (triangles) accuracy of the same checkpoints. (b-d) Subspace-wise tests on the frame-level latents of DecVAE + EWT ($\beta = 0.1$); X, subspace of the original; OC1 to OC3, component subspaces in increasing frequency. \textbf{b)}, Prediction from a single subspace (chance-normalized balanced accuracy of a linear probe), with the full latent as bottom row. \textbf{c)}, Ablation: drop in the full-latent score when the subspace is removed. \textbf{d)}, Swap: transfer of the factor when the subspace is taken from a frame that differs only in it, minus a matched control ($\Delta T$). Mean ± s.d. over three seeds. \textbf{e)}, Selectivity of the DecVAE subspaces (coloured, one marker per seed) and of 20 random partitions (grey), for prediction and ablation, across decompositions and $\beta$. \textbf{f)}, Linear CKA~\cite{kornblith2019similarity} between the subspaces of DecVAE + EWT models trained with different seeds (mean ± s.d. over seed pairs).}}}
    \label{fig7}
\end{figure*}

\subsection*{Information allocation and shaping of latent subspaces}
The VDA objective acts on latent representations through three distinct disentanglement dynamics (Fig.~\ref{fig1}c) and in the previous results, we observed how the latent representations learned by VDA, at their equilibrium, lead to disentangled and interpretable representations across datasets. One of the mechanisms that acts on separation of the components only indirectly, the prior weight $\beta$ draws component subspaces towards the same prior, a dynamic that contradicts the separation-promoting objective of $L_{ortho}$ (see `Variational decomposition autoencoding objective'). We measured separation of components as the mean Jensen-Shannon divergence (JSD) between embeddings of pairs of components, a quantity bounded by $\ln2$ (Methods Proposition~2a) in the sequence branch of the SimVowels test set for different $\beta$ values (see Fig.~\ref{fig7}a, Supplementary Fig.~32a-c). Large values of $\beta>1$ reduce separation, with separation staying close to the predicted by Proposition 2a value up to $\beta=1$ (EWT 0.61 to 0.69 nats, close to $\ln2$). Notably, downstream task performance is more robust to separation: speaker identification is best at $\beta=0.1$ and vowel recognition varies little (see Fig.~\ref{fig7}a lower panel, Supplementary Fig.~32a,b). These results justify the choice of a small $\beta$ value that costs little to theoretical separation results, and is accompanied by the best downstream task results.

For $\beta\leq1$, DecVAE with EWT~\cite{gilles2013ewt} reaches the equilibrium prediction of Proposition~2 (see Supplementary Fig.~33): components are near-orthogonal within each utterance (mean cosine similarity of at most 0.03) and divergence from the original is close to the predicted value ($j_3=0.318$, see Fig.~\ref{fig7}a). Moreover, each component keeps largely the same embedding dimensions in every utterance, as evidenced by the low cosine similarity between embeddings of different components (0.03 to 0.20, against 0.997 for an untrained model, see Supplementary Fig.~33); this behavior is similar to the fixed loadings of PCA~\cite{Greenacre2022PrincipalAnalysis}. With FD the equilibrium is reached only partly (see Supplementary Section~6.2).

Our results across synthetic, real speech, dysarthric and emotional speech, have shown that VDA learns disentangled, visually interpretable embeddings where generative factors can be separated up to a degree (Figs.~\ref{fig2},\ref{fig3},\ref{fig4},\ref{fig5},\ref{fig6}a,b). Nevertheless, our theoretical support (Proposition 2) does not imply that the orthogonal latent subspaces also correspond to generative factors; to that end, we test this hypothesis on frame-level latents with linear probes~\cite{alain2016understanding} (see `Subspace-wise analysis'). We perform experiments that include (i) the prediction of a factor from a single subspace, (ii) information loss when a subspace is removed i.e. ablation, (iii) the prediction of a factor when a subspace is exchanged (swapped) between two frames that differ only in that factor, measured against a matched control~\cite{geiger2021causal}. In the SimVowels generator the first two formants vary mainly with vowel, and the third with speaker (Cohen's $\eta^2$~\cite{cohen1973eta} of 0.76, 0.81, 0.08 for variance explained by vowel in formants F1,F2,F3, respectively), and in DecVAE with EWT~\cite{gilles2013ewt} subspaces this allocation pattern is retained (Fig.~\ref{fig7}b-d); OC1 and OC2 better predict vowel, whereas OC3 better predicts the speaker. Similarly, removing OC3 in Fig.~\ref{fig7}c costs mainly speaker information (a drop of 0.12 in balanced accuracy) and not vowel information (drops 0.02, balanced accuracy), and swapping OC1 and OC2, or OC3, costs vowel, or speaker accuracy, respectively (Fig~\ref{fig7}d). It should be noted that the subspace of the original signal, $X$, carries most of the vowel information as evident in Fig.~\ref{fig7}b. The same pattern is observed when analyzing DecVAE with FD as was done with EWT (Supplementary Fig.~34b-d).

To further summarize the organizational properties of DecVAE into subspace, we also measure selectivity, the subspace-level counterpart of DCI disentanglement, and compare it with 20 random partitions of the same latent dimensions (as in~\cite{hewitt2019designing}, see Fig.~\ref{fig7}e). For different decomposition methods, we observe that DecVAE subspaces are more selective in prediction of generative factors than random partitions (Fig.~\ref{fig7}e, left), and in ablation (Fig.~\ref{fig7}e, right), when using the four decomposition methods~(FD, EWT~\cite{gilles2013ewt}, VMD~\cite{dragomiretskiy2014vmd}, EMD~\cite{huang1998empirical}), with the exception of EMD in ablation where the mode mixing across components leads to subspaces less selective than the random partitions. Since disentanglement can depend on random initialization~\cite{Locatello2019ChallengingRepresentations}, we also measure the consistency of subspaces between random seeds using centered kernel alignment (CKA)~\cite{kornblith2019similarity}; CKA results shows a high consistency between the embeddings of a subspace and the same subspace in a different seed, than for subspaces of the same seed (linear CKA 0.28 to 0.77, against 0.01 to 0.19, Fig.~\ref{fig7}f), and the same latent dimensions carry the same factors in every cross-validation fold (Supplementary Section~11.6).

On TIMIT, speaker information is localized in the highest frequency components (prediction from OC4 of 0.34 against at most 0.14 for the other subspaces; selectivity 0.45, against at most 0.10 for random partitions; see Supplementary Fig.~34g, Supplementary Table~24), whereas phonemes are not linearly decodable from the subspaces or the full latent (0.07, see Supplementary Fig.~34g, Supplementary Table~24). These results show that the latent subspaces of DecVAE are frequency-band representations that become factor-specific, to the extent that the generative factors reside in the different analyzed bands; following we experiment with a synthetic dataset where factors do not reside in different frequency bands.

\subsection*{Where the decomposition prior helps, and where it does not.}
The decomposition prior of VDA assumes that generative factors are tied to distinct spectral structures. We test this assumption by misspecifying the decomposition on the known ground truth of SimVowels, real speech on TIMIT, and a dataset where generative factors are not tied to spectral structures, SimCoupled (see 'Datasets', Supplementary Section~3).

\begin{figure*}[h!]
    \centering
    \includegraphics[width=0.95\textwidth,height = 14cm]{figures_main/fig_8.png}
    \caption[]{\footnotesize{\textsf{\textbf{$|$ Sensitivity of DecVAE to the decomposition parameters}. \textbf{a)}, Components FD keeps per frame ($K$) on 100 test utterances per dataset, for the default peak detection intervals (bars) and, on SimVowels, six alternative layouts (lines); dashed line, generator $C = 3$. \textbf{b)}, Share of frames whose peaks are grouped ($K > C$), match $C$, are completed with empty components ($0 < K < C$) or have none ($K = 0$), per $C$. \textbf{c-e)}, $\beta$-DecVAE (FD, $\beta = 0.1$) with $C = 2$ to $6$ on SimVowels (blue; dashed line, generator $C$) and SimCoupled (red): \textbf{c)}, DCI disentanglement; \textbf{d)}, effective number of component subspaces carrying each factor (dotted line, $n_{\mathrm{eff}} = C$); \textbf{e)}, selectivity of the component subspaces minus the mean of 20 random partitions, for prediction, ablation and swap (see Methods). On SimCoupled ablation and swap fall below 0 because the undecomposed signal subspace holds the information (Fig. 9d). \textbf{f)}, DCI disentanglement on SimVowels ($C = 3$) for five interval layouts; dots, single runs. Error bars, s.d. over three seeds or three random layouts.}}}
    \label{fig8}
\end{figure*}

\subsubsection*{Sensitivity to the decomposition parameters}
Using FD, our simple band-pass filtering decomposition when compared to more complex decomposition approaches (see `Decomposition model and self-supervised learning masking'), a variable number of spectral peaks per frame $K$ are kept, due to the peak detection algorithm in FD; if less spectral peaks are found, FD zero-pads components to obtain exactly $C$ components, whereas if more than $C$ components are found, FD groups them to $C$. Fig.~\ref{fig8}a,b demonstrates how often this happens, and how many components FD typically detects in SimVowels, TIMIT, and SimCoupled. The spectral peak detection results show that FD detects the correct $C=3$ components (ground truth number of formants) in SimVowels in $73\%$ of frames, even if the peak detection algorithm is confounded (Fig.~\ref{fig8}a,f) to search for peaks on few, too many, or random intervals; this demonstrates the robustness of FD with parameter selection (DCI-D remains stable across different peak detection schemes Fig.~\ref{fig8}f), hinting that the choice of hard band-limited components is not a prerequisite for good performance, as the peak detection algorithm can detect prominent spectral peaks nevertheless. On real speech, $K$ varies widely with a median value of 3, and on SimCoupled a wide range of $K$ is also observed (median 8), due to the harmonics of the oscillators in the stochastic autoregressive model of SimCoupled (see `Datasets', Supplementary Section~3).
Furthermore, when we evaluate $\beta$-DecVAE + FD on SimVowels with $C \in \{2,3,4,5,6\}$, performance asymmetry manifests (Fig.~\ref{fig8}c-e); underestimating $C$ is harmful, as evident by the drop in performance when $C=2$ (DCI-D drops from 0.58 to 0.36, subspace selectivity drops below the random partition level under ablation and swap). On the other hand, overestimating $C$ spreads each factor over more subspaces (Fig.~\ref{fig8}d) without harming disentanglement performance (DCI-D increases from 0.65 to 0.77, see Supplementary Fig.~35 for other metrics), although part of this increase might be from the larger latent space (Supplementary Fig.~31). On SimCoupled, no value of $C$ produces structure, as factors spread evenly across all components and selectivity stays at or below the random-partition level (Fig.~\ref{fig8}d,e).

\subsubsection*{Testing the decomposition prior on non-frequency, correlated generative factors}
SimCoupled is a synthetic dataset built upon a stochastic, coupled autoregressive oscillator model introduced in~\cite{ding200617} (Supplementary Section~3, Supplementary Fig.~37), a standard benchmark for causal interaction between time series that has also been analysed with signal decomposition~\cite{yang2018causal}. We generate SimCoupled to have two latent generative factors, the coupling lag, which reshapes the frequency spectrum as a comb filter, and gain, which offsets the amplitude of the whole frequency spectrum equally; neither factor is localized in a specific frequency band (see Fig.~\ref{fig9}a,b). In the generation process, factors are drawn with a high correlation (0.77) and assigned to samples; we also generate a held-out independent split that draws factors independently. Our results show that DecVAE can predict both factors better than $\beta$-VAE~\cite{Higgins2017Beta-VAE:Framework} and CoST~\cite{Woo2022CoST:Forecasting} (lag accuracy 0.91 against 0.64, 0.81, and gain accuracy 0.77 against 0.55, 0.71) but worse than PCA~\cite{Greenacre2022PrincipalAnalysis}, SFA~\cite{wiskott2002slow}, FHVAE~\cite{Hsu2017UnsupervisedData}, wav2vec2~\cite{Baevski2020Wav2vecRepresentations} (0.95 to 0.99). DecVAE and CoST~\cite{Woo2022CoST:Forecasting} lose the most lag accuracy when evaluating on the independent test split (0.91 to 0.64 from correlated to independent; Fig.~\ref{fig9}c, Supplementary Table~20). In disentanglement metrics on the correlated test set, DecVAE has higher DCI disentanglement than most methods (trails only wav2vec2 and SFA), has lower GCN than most methods (trails only PCA and wav2vec2 + PCA), has the best robustness score (IRS 0.526), and is very informative, as it scores a DCI informativeness of 0.81 against top performing SFA (DCI I 0.907) and beats worst performing $\beta$-VAE (DCI I 0.585) (Table~\ref{tab:baselines}a). In the independent test set (Supplementary Table~20), DecVAE remains the most robust disentangler and performs moderately in GCN, DCI disentanglement and informativeness. Fig.~\ref{fig9}d shows that the subspace of the original, undecomposed signal holds most of the information. Notably, even though spectral content presents a large spread in SimCoupled (Fig.~\ref{fig9}a,b, Supplementary Fig.~37), the decomposition structure is able to capture lag and gain traits that survive in individual components; OC3 holds a significant part of the lag and gain information in the correlated set, and of the gain information in the independent set (Supplementary Fig.~38d). These results indicate that even though the generative factors are not tied to distinct spectral structures, a frequency decomposition can partly recover the underlying information, though without disentangling the non-frequency related generative factors as clearly as in the case of SimVowels, where individual component subspaces specialise in predicting specific generative factors (Fig.~\ref{fig7}b-d, Fig.~\ref{fig9}d).

\begin{figure*}[h!]
    \centering
    \includegraphics[width=0.95\textwidth,height = 13cm]{figures_main/fig_9.png}
    \caption[]{\footnotesize{\textsf{\textbf{$|$ DecVAE on SimCoupled, whose factors are not band-limited.} \textbf{a)}, Mean log-power spectrum per lag class (left; gain removed) and per gain class (right), independent split. \textbf{b)}, Share of each factor’s effect on band log-energy per Mel band, independent split; dotted line, uniform share.\textbf{c)}, Lag and gain accuracy of DecVAE (red) and baselines on the correlated test set (filled) and the independent split (open), classifiers trained on correlated data. \textbf{d)}, Prediction score of each subspace of $\beta$-DecVAE + FD ($\beta = 0.1$, three random seeds; mean ± s.d.), with the full latent as bottom row.}}}
    \label{fig9}
\end{figure*}

\section*{Discussion}
Variational decomposition autoencoding is a disentangled representation learning framework for complex, nonstationary, high dimensional time series data whose generative factors have distinct spectral signatures. DecVAE facilitates disentanglement through the synergy of decomposition and prior approximation mechanics in separating frequency-related factors inside latent subspaces (Fig.~\ref{fig1}c). VDA is applicable to different time scales beyond the short-term phonetic content (Fig.~\ref{fig1}d); a second sequence-level latent proves to be quite useful as a natural disentanglement mechanism to isolate slow variables and brings performance improvements in the cases of SimVowels, TIMIT and VOC-ALS, compared to VAE-based methods, ICA, and PCA (Fig.~\ref{fig2},\ref{fig3},\ref{fig4}, Supplementary Tables~8-12). 
On IEMOCAP all examined models could recognise emotion better from sequence-level representations, rather than frame-level representations (Supplementary Tables~17,18).

We evaluated DecVAE using a wide variety of disentanglement and task-specific metrics (see 'Disentanglement Evaluation'). A domain where DecVAE consistently underperformed compared to other methods is compactness of representations, measured by DCI completeness \cite{Eastwood2018ARepresentations}. 
In VAEs compactness is enforced as an effect of the combined Gaussian prior approximation and the decoder reconstruction that embeds similar samples close to each other in the latent space. We observe the effects of having a compact but less disentangled representation in the latent responses analysis of VAE models, whereas DecVAE representations are not compact but they are disentangled and informative (Fig.~\ref{fig6}a,b,c, Supplementary Fig. 18,19,20,21). 
This lower compactness in representations comes with a practical cost. Each latent subspace of DecVAE has the size of the whole latent space of VAE baselines (see Supplementary Tables~4,6); this leads to aggregated representations having $C+1$ times as many dimensions (e.g. 192 against 48 on SimVowels, and 320 against 64 on TIMIT) and every observation needs to be decomposed and then encoded $C+1$ times. In edge or clinical deployments where memory and compute are limited, this redundancy increases storage, inference time and the cost of downstream models. A reduction of the cost could be achieved by keeping only informative subspaces for the task at hand, by a learned projection of the concatenated subspaces onto a space with reduced dimensionality (although this can reduce performance, see Supplementary Fig.~26), or by simply learning smaller subspaces. In this work, we have not experimented on how each of these affects disentanglement at fixed budgets.

Our experiments show that DecVAEs disentangle better than VAE-based models in all four speech datasets (Fig.~\ref{fig6}d), and that the decomposition prior is a fundamental reason why: the component subspaces of DecVAE specialise when the generative factors reside in different frequency bands (Fig.~\ref{fig7}) and do not when the factors are not band-limited, as revealed by our analysis on a correlated coupled dataset of non-frequency factors (Fig.~\ref{fig9}). DecVAE is a self-supervised model that undergoes a pre-training phase; after pre-training it can be transferred as a representation function, zero-shot or fine-tuning, and maps high-dimensional data to low-dimensional disentangled representations. We have evaluated DecVAE with simple classifiers in the post-training phase to exemplify the quality of the learned representations. Compared with nine time-series, SSL, and VAE-based baselines (Table~\ref{tab:baselines}), DecVAE yielded the most robust representations on real speech and on SimCoupled, and the highest disentanglement among models trained on TIMIT. Frozen SSL LAMs, pretrained on far more speech (960 to 94,000 hours), give more informative and more disentangled representations. On VOC-ALS, we used zero-shot transfer learning in a challenging scenario, and found that disease stage prediction is confounded by speaker identity, as each speaker is characterized by a single disease stage; prediction results fall to chance when speakers are held out (Supplementary Table~21). On IEMOCAP, we used fine-tuning transfer learning: DecVAE gave more disentangled and more robust frame-level representations than VAE-based models, PCA and ICA; predicting emotion through DecVAE was shown to be less efficient than baselines. Since the DecVAE encoder is essentially a wav2vec2 convolutional encoder, training it with the VDA objective from a starting point of a pretrained audio model, rather than from scratch, is a natural way to combine the robustness of the decomposition prior with informativeness of large-scale pretraining.

Our ablations reveal that disentanglement arises even without the Gaussian prior term. DecAE, i.e., DecVAE with $\beta=0$, has the same architecture, latent dimensionality and decomposition objective, and differs only in the absence of the prior term. We report DecAE results alongside DecVAE in Supplementary Tables~7-10,14,15, and Supplementary Fig.~12 and 14a; DecAE's training dynamics are also shown in Supplementary Fig.~22a,b. This disentanglement behavior emerges because DecVAE consistently maximizes and minimizes divergences of components to other components and original signal in real and simulated data (Supplementary Fig. 22,24). 
The choices of important hyperparameters such as $\beta$ (Supplementary Fig. 22
a,b), decomposition method (Supplementary Fig. 22
e,f), components number (Supplementary Fig. 23
c,d), aggregation function $a(\cdot)$ (Supplementary Fig. 26) dictate the exact structural bias of VDA. Balance between noisy, redundant representations and structural biases is key to performance of DecVAEs. 
When the optimization problem becomes more demanding, such as a large $C$ with $\beta=1$, convergence slows down (Supplementary Fig.~23d), although at $\beta=0.1$ a larger $C$ does not reduce performance (Fig.~\ref{fig8}c).
Similarly, using very few frames in the SSL loss calculation (Supplementary Figs.~14b,23a,b,25) also hinders learning. The same applies when the input is more abstract, such as raw waveform instead of Mel filterbanks (Supplementary Fig. 22
d).
These conditions can negatively impact both disentanglement and informativeness of the learned representations. This is consistently observed across our ablation results (Supplementary Tables~7-9
).



\paragraph{Disentanglement through structured latent decomposition.} We have deliberately excluded the decoder network from our analysis to showcase the strength of our structural biases. The decoder is an essential part of disentanglement in VAE-based methods; yet, aligned with recent theoretical advances in the understanding of VAEs~\cite{Rolinek2018VariationalAccident,Mathieu2018DisentanglingAutoencoders}, our work posits that enforcing structural biases is actually what mostly promotes disentanglement. Our results showcase that, when the generative factors have distinct spectral signatures, decomposition can extend disentanglement beyond the orthogonal Gaussian latent structure \cite{Higgins2017Beta-VAE:Framework,Chen2018IsolatingAutoencoders, kim2018factorVAE}. We provide an intuition on why this happens in Supplementary Information section 7
. More specifically, our analysis shows that VDA is a hybrid of variational methods and PCA-based methods. In Supplementary Information section 8 we relate VDA to PCA: at the optimum of the decomposition terms, components are mapped to mutually orthogonal directions of the normalized embedding space (Proposition~2), and on SimVowels (with EWT) each component uses the same embedding dimensions across utterances (see Supplementary Fig.~33), analogously to the fixed orthogonal loadings of PCA~\cite{Greenacre2022PrincipalAnalysis}. In contrast to VAEs, where an uncorrelated latent structure arises from the prior approximation, in DecVAEs the decomposition bias is inserted explicitly and the representation is aligned with time-frequency properties; this motivates an analogy between DecVAE and a singular value decomposition (SVD) of the input (Supplementary Fig. 1).

We also tested the limits of VDA in a challenging setting where generative factors are not band-limited and are correlated (see `Testing the decomposition prior on non-frequency, correlated generative factors'). DecVAE demonstrated competitive disentanglement performance compared to baselines on a correlated test set, where it tops baselines in representational robustness, and is amongst the three top performing models in DCI disentanglement and GCN, while retaining highly informative representations (see Table~\ref{tab:baselines}a); when the generative factors are uncorrelated, DecVAE demonstrates moderate disentanglement capability (see Supplementary Table~20). Analyzing the content of VDA subspaces in SimCoupled reveals that when generative factors are not band-limited and spread across the spectrum, component subspaces learned through a frequency decomposition can only partly hold lag and gain information, and do not specialise in prediction of specific factors, as in SimVowels (Fig.~\ref{fig7}b-d, Fig.~\ref{fig9}d). VDA therefore performed competitively to baselines trained on much more data (wav2vec2), time-series and self-supervised baselines, as well as methods that discover statistical dependencies (PCA, SFA); although VDA is not equipped with mechanisms that can disentangle non-band-limited generative factors (Proposition~2), it demonstrated robustness in a demanding task that is unfavourable to the assumed decomposition prior.

\paragraph{Design insights and extension pathways of variational decomposition autoencoding.} We showcased the versatility of DecVAEs in disentangling various generative processes in speech data using minimal data preprocessing or modality assumptions. We implemented DecVAEs as deep encoders with one-dimensional convolution layers; this makes them applicable in principle to any other sequence modality that contains distinct spectral structures, such as EEG, ECG or fMRI. Our real-world evaluations tackled a challenging and complex time series modality -speech - a modality whose generative factors are closely tied to its spectral structure; whether the gains observed in speech can carry over to other time series where this is less the case, remains to be tested.
We focused our evaluation on disentanglement quality and not generation; disentanglement is an important prerequisite for high quality generation, an important task that we leave as future work. The VDA dynamics of orthogonality and reconstruction implemented through a self-supervised contrastive task, allow the potential extension of our method to disentangle multi-modal datasets by using modalities as components instead of utilizing a decomposition. In such settings, each modality—be it audio, visual, textual, or physiological—can be treated as a distinct input channel with its own time-frequency or spatial signature, enabling DecVAE to learn modality-specific latent subspaces while preserving cross-modal coherence. This formulation naturally accommodates fusion tasks, where disentangling modality-specific generative factors is critical for downstream interpretability, such as in emotion recognition from audiovisual data, clinical decision support from speech and biosignals, or scene understanding from video and text. By aligning latent dimensions with modality-driven structure rather than decomposed frequency bands, DecVAE could carry its disentanglement mechanics over to heterogeneous data sources, offering a unified framework for structured representation learning across domains. 

VDA employs the inductive bias of decomposition, whose benefit depends on how well the components can match the structure carrying the generative factors. Merging the signal of two factors into one component stops the subspaces from specialising, whereas splitting a factor over more components than needed spreads it without reducing performance; $C$ should be at least the number of spectrally distinct structures expected in the signal. When the factors have no distinct spectral structure, as in SimCoupled, the decomposition does not create subspaces that specialise. As demonstrated by our experimental results, VDA's utility is exemplified in signals whose generative factors have distinct spectral signatures, as in speech; learning a decomposition that can project non-band-limited factors to orthogonal subspaces is a direction for future work. It should be noted that the physical utility of the coupled oscillators system that SimCoupled is based on \cite{ding200617}, is the identification of causal relationships between sources. The formulation of SimCoupled is disentanglement-oriented, and therefore based on prediction of the generative factors lag and gain. We did not investigate the utility of latent component subspaces in isolating causal interactions; decomposition has been shown to aid in extracting causal interactions from this system~\cite{yang2018causal}, and therefore using VDA as a causal discovery system remains a direction for future research.

While the proposed framework demonstrates broad applicability, several aspects invite further exploration. Our evaluation is limited by several aspects of the utilized metrics; supervised disentanglement metrics depend on the factor labels, and under factor imbalance, correlated factors, label noise and coarsened labels, the ranking of models on the core metric set that we used is preserved (Kendall's $\tau$ 0.60 to 0.95) but it changes depending on the set of metrics averaged. We also reported the effect of growing latent dimensionality on metrics, especially DCI disentanglement, and chance levels of metrics IRS and modularity, estimated by cyclically shifting the labels and comparing to Gaussian latent noise (Supplementary Section~11.5). We therefore reported several metric families, evaluated LAMs after PCA dimensionality reduction, and compare rankings across different metric sets rather than relying on single scores. Our analysis also revealed a confounding effect of speaker identity on clinical labels that are constant within a participant, such as King's stage in VOC-ALS; under participant-independent evaluations, none of the tested representation learners, DecVAE included, managed to capture more than weak stage information on VOC-ALS. The performance of DecVAE is influenced by the choice of decomposition method, as certain techniques may introduce residual correlations that affect latent separation. To address this, we systematically evaluated multiple decomposition strategies and selected those that promote frequency-resonant orthogonality. Additionally, the current architecture is optimized for sequential data, which may require structural adaptation for spatial modalities such as images. Our encoder-only design also prioritizes representation quality over generative synthesis, leaving decoder integration as a promising direction for future work. Finally, while contrastive regularization and orthogonality constraints help mitigate latent overlap, further refinement may enhance modularity and independence in complex, multi-modal settings. These considerations highlight the flexibility of the VDA framework and suggest clear pathways for extension across data types and application domains.

Building on the novel optimization objective of VDA, future work could further explore its extension to non-sequential data domains, such as images, by adapting the DecVAE architecture to accommodate spatial rather than temporal decomposition. This would involve restructuring the encoder to capture localized spatial features and redefining the decomposition model to reflect image-specific priors. Additionally, the contrastive self-supervision and orthogonality constraints could be repurposed to enforce semantic separation across modalities in multi-modal datasets. Treating each modality as a distinct latent subspace may enhance robustness to missing inputs, domain shifts, and modality-specific noise, paving the way for interpretable, disentangled representations in heterogeneous data environments. Taken together, these insights affirm that VDA is not merely a refinement of existing latent modeling techniques—it is a principled way to build a decomposition prior into representation learning, demonstrated here on speech, whose benefit is greatest when the generative factors are tied to distinct spectral structures, and holds potential for disentangling structure in dynamic data, poised to advance interpretability, generalization, and scientific discovery across modalities.





\section*{Methods}
Fig.~\ref{fig1} displays the main working principles of variational decomposition autoencoding (VDA) and its neural network implementation of variational decomposition autoencoder (DecVAE) (Supplementary Fig. 2
). DecVAE accepts sequences from a dataset $X = \{x_1,x_2,...,x_N\} \in \mathbb{R}^{K \times N}$ and applies a decomposition module $\mathcal{D}_w^C$ to decompose the input sequence $x_n$ into a series of components (Supplementary Fig. 2
). The DecVAE encoder then receives the inputs and their components and embeds them in latent subspaces $\{\tilde{z}^0, \tilde{z}^1, ..., \tilde{z}^C\}$, according to their distinct time and frequency domain characteristics. An aggregation function $a(\cdot)$ then projects the collection of latent subspaces onto the latent space $\tilde{Z}$. In our ablations we find that the best choice of aggregator is a simple concatenation of subspaces contrary to more complex aggregations such as learnable projection (see `Aggregation of latent subspaces'). Fig.~\ref{fig1}d shows how DecVAE can also simultaneously handle latent variables in different time scales (see 'Multiple latent variables'). The notation used in the Methods and the Supplementary Information is summarized at the beginning of Supplementary Information ('Supplementary Note: Notation').

In contrast to variational autoencoding \cite{Kingma2014Auto-EncodingBayes,Higgins2017Beta-VAE:Framework} (Fig.~\ref{fig1}a, Supplementary Table~2
), VDA (Fig.~\ref{fig1}b) assumes generative subfactors that are characterized by prior distributions $p_\theta(z_1), p_\theta(z_2), ..., p_\theta(z_C)$. The generative factors are then generated by superposing information from these subspaces. Whereas VAEs operate to directly recover the true distribution of generative factors $p(z)$, DecVAEs recover the sub-factor distributions that synthesize $p(z)$. As in VAEs, there is a single inference model $q_\varphi$ with parameters shared across components, which maps each component to a Gaussian latent subspace. Starting from the input space $X$ and observed data distributions ${p_D(x), p_D(x^{(1)}), p_D(x^{(2)}),..., p_D(x^{(C)})}$ after the decomposition model $\mathcal{D}_w^C$ has been applied, DecVAE projects each component to its own latent subspace, following the factorization of the posterior implied by the generative process (Proposition 1, see `Problem Formulation'). 

Fig.~\ref{fig1}c illustrates the disentanglement mechanics employed in VDA; VDA is an enhanced variational inference scheme that complements the Gaussian prior approximation of VAEs, with two disentanglement forces from the input decomposition domain, orthogonality and reconstruction. During the training of DecVAEs, the dynamics between the latent subspaces of components and the original signal, as well as the Gaussian prior approximation, lead to an equilibrium in the embedding space $H$ that leads to disentangled representations. This optimization objective which we term variational decomposition autoencoding (VDA) objective, is expressed as follows:

\begin{equation}
    \mathcal{L}_{VDA}(\varphi; X) = L_{recon} + L_{ortho} + \beta L_{prior}
    \label{vda_overview}
\end{equation}
\noindent
The decomposition dynamics of reconstruction and orthogonality are enforced in the hidden space $H$ of embeddings through the terms $L_{recon}$ and $L_{ortho}$. $L_{recon}$ encourages the distribution of each component's embedding $h_i$ to align with that of the original signal's embedding $h$, by encouraging their Jensen-Shannon divergence (JSD)~\cite{Menendez1997TheDivergence} to approach zero. $L_{ortho}$ encourages the distributions of every pair of components to separate, by diverging their JSD towards its maximum, which is reached when the two distributions have disjoint supports, i.e., when the normalized embeddings are orthogonal (Proposition~\ref{prop_equilibrium}). This orthogonality is a property of the normalized embeddings in the intermediate hidden space $H$; the VDA objective itself does not enforce orthogonality or statistical independence of the latent subspaces, whose exact dependence dynamics are assessed empirically.

The importance of $L_{recon}$ and $L_{ortho}$ is critical for the disentanglement of the latent space (Supplementary Fig. 22a
, upper left and right panels). Even in the absence of a Gaussian prior pressure ($\beta = 0$), these terms converge to respect the input decomposition dynamics. $L_{prior}$ is the Kullback-Leibler divergence (KLD) between the Gaussian latent projection head of each component subspace and the standard normal prior, weighted with a user-defined pressure $\beta$ \cite{Higgins2017Beta-VAE:Framework}, and applied to each of the component subspaces $\{\tilde z_1,\dots,\tilde z_C\}$. In our preliminary experiments, regularizing the subspace of the original signal $\tilde z_0$ practically had no effect on the representation and results, and was therefore omitted. $L_{prior}$ also acts on component embeddings through these Gaussian latent heads; empirically a stronger prior regularization delays convergence with increasing $\beta$, acting as a countering force to the separation force of $L_{ortho}$, and making distributions less separable (Supplementary Fig.~22a,b upper panels, and Fig.~\ref{fig7}a).

Increasing the value of $\beta$ (Supplementary Fig. 22a
), results in a slower convergence of the objective as an additional term is added, and also influences the convergence of the decomposition terms $L_{recon}, L_{ortho}$. The effect of $L_{prior}$ is more pronounced in the early stages of training and its convergence is fast (Supplementary Fig. 22a
, lower right panel). Moreover, a high value of $\beta$ can be detrimental to the training as it can significantly slow down convergence of the other two terms (Supplementary Fig. 22
a,b, upper panels). 



\subsection*{Problem formulation}
\label{problem_formulation}
We consider $N$ i.i.d. observations of a dataset $X = {\{x_1,x_2,...,x_N\}}$, $x_n\in\mathbb{R}^{K}$ and we assume that each observation is generated by a three-step process (Fig.~\ref{fig1}.b.i), which involves a set of latent variables $z_1,z_2,...,z_C$ that are parts or components of an unobserved continuous latent variable $z$. Each component $z_i$ takes values in its own subspace $Z_i$ of dimension $k_i$ and the latent space of generative factors is their direct sum, i.e. $Z=Z_1 \oplus Z_2 \oplus\cdots\oplus Z_C$. In order to identify each $z_i$ with its embedding in $Z$ (zero outside $Z_i$),
components are characterized by orthogonality with respect to each other (Eq.~(\ref{eq_ortho_generative_space})) and their superposition satisfies the reconstruction principle and yields $z$ (Eq.~(\ref{eq_recon_generative_space})): 

\begin{equation}
    \langle z_i, z_j\rangle=\sum_{u=1}^{k} z_{i,u}\,z_{j,u}=0,\qquad \
    \forall\, i\neq j,
    \label{eq_ortho_generative_space}
\end{equation}
\begin{equation}
    z=\sum_{i=1}^{C} z_i=[\,z_1;z_2;\dots;z_C\,],
    \label{eq_recon_generative_space}
\end{equation}
\noindent
where $\langle \cdot,\cdot \rangle$ is defined as the inner product in the space $Z$, $k=\sum_i k_i$, and $[\,\cdot\,;\,\cdot\,]$ denotes concatenation. Under this formulation, Eq.~(\ref{eq_ortho_generative_space}) is valid for every instance of components in the dataset $X$, whereas Eq.~(\ref{eq_recon_generative_space}) expresses a superposition that is a concatenation of the component subspaces; this coincides with the practical choice of an aggregation function that we found most successful in our experiments (see `Aggregation of latent subspaces').

For a single observation $x_n$ the  generative process $\theta$ generates components $z_1^{n},z_2^{n},...,z_C^{n}$ from prior distributions $p_\theta(z_1),p_\theta(z_2),\dots,p_\theta(z_C)$ that compose $z^n$ (as in Eq.~(\ref{eq_recon_generative_space})), which follows the prior $p_\theta(z)=\prod_{i}p_\theta(z_i)$. Subsequently, $x_n$ is generated by the conditional distribution $p_\theta(x\,|\,z_1,z_2,...,z_C) \equiv p_\theta(x\,|\,z)$ as a superposition of component signals that stem from the latent components. The parameters of the process $\theta$ and the values of the set of latent variables $\{z^n; z_1^n,z_2^n,...,z_C^n\}$ are unknown and we want to estimate them. The joint probability for this three-step generative process is given by Eq.~(\ref{eq_joint_probability}):

\begin{align}
\label{eq_joint_probability}
p_{\theta}(X,z_{n,1},z_{n,2},\dots,z_{n,C})_{n=1}^N &= \prod_{n=1}^{N} p_\theta(x_n\,|\,z_{n,1},z_{n,2},\dots,z_{n,C})\,p_{\theta}(z_{n,1})\,p_\theta(z_{n,2})\cdots p_{\theta}(z_{n,C}), \nonumber\\
p_\theta(z_{n}) &= p_\theta(z_{n,1},z_{n,2},\dots,z_{n,C})=p_\theta(z_{n,1})\,
p_{\theta}(z_{n,2})\cdots p_{\theta}(z_{n,C}),  \\
p_\theta(x_n\,|\,z_{n}) &= p_\theta(x_n\,|\,z_{n,1},z_{n,2},\dots,z_{n,C}) \nonumber
\end{align}
\noindent
where Eq.~(4b) is the joint density of the components; we write $z_i$ when $n$ is not needed. Because the map from $(z_1,\cdots,z_C)$ to $z$ in Eq.~(\ref{eq_recon_generative_space}) is a concatenation, and thus a bijection, Eq.~(4b) is also the density of $z$. Each term of the joint probability in Eq.~(\ref{eq_joint_probability}) can be expressed as:

\begin{align*}
p_\theta(x\,|\,z_1,z_2,\dots,z_C) &= \mathcal{N}\Big(x \;\Big|\; \textstyle\sum_{i=1}^{C} g_i(z_i),\; \sigma_x^{2} I_K\Big)\\
p_\theta(z_i) &= \mathcal{N}\big(z_i \,|\, 0,\; \sigma_{z_i}^{2} 
I_{k_i}\big),\qquad \forall\, i\in\{1,\dots,C\}
\end{align*}
\noindent
where $\theta$ parameterizes the generative process, $g_i:Z_i\rightarrow V_i$ maps each latent component to a component signal in a subspace $V_i\subset\mathbb{R}^K$ of the observation space, and $V_1,\cdots,V_C$ are mutually orthogonal with $V_1\oplus\cdots\oplus V_C=\mathbb{R}^{K}$. The component signals $g_i(z_i)$ that we assume in the observation space $\mathbb{R}^{K}$, are mutually orthogonal and their sum $\sum_i g_i(z_i)$ satisfies reconstruction in $\mathbb{R}^{K}$. These properties are the counterparts of the orthogonality and reconstruction principles of Eq.~(\ref{eq_ortho_generative_space}) and Eq.~(\ref{eq_recon_generative_space}) expressed in the input observation domain, and are properties that we require of our decomposition model used below (see 'Decomposition model and self-supervised learning masking'). Because DecVAE has no decoder, the maps $g_i$ that belong to the assumed data generating process of VDA are not learned in DecVAE. The priors $p_\theta(z_i)$ that we assume here over the latent variables $z_i$ are isotropic multivariate Gaussian distributions centered around 0, so that $p_\theta(z)$ is a Gaussian with block-diagonal covariance; in practice $\sigma_{z_i}=1$ is used, which gives the standard normal prior.

Moreover, as is also the case in the variational autoencoding framework \cite{Kingma2014Auto-EncodingBayes}, the true posterior distribution $p_\theta(z|x)$ is in general intractable, and an inference model $q_\varphi(z|x)$ is used in its place. For the generative process of Eq.~(\ref{eq_joint_probability}), however, the dependency structure of the posterior is known. Let $P_i$ denote the orthogonal projection of $\mathbb{R}^K$ onto $V_i$ and $x^{(i)} = P_i\,x$ the $i$-th component of an observation $x$.

\begin{proposition}[Posterior factorization under orthogonal 
superposition]\label{eq_prop_posterior}
Under the generative process of Eq.~(\ref{eq_joint_probability}) with mutually orthogonal subspaces $V_i$ and isotropic noise, the posterior 
factorizes over components, and each factor depends on $x$ only 
through its own component:
\begin{equation}
p_\theta(z_1,z_2,\dots,z_C\,|\,x)=\prod_{i=1}^{C} p_\theta\big(z_i\,|\,x^{(i)}\big),\qquad x^{(i)}=P_i\,x .
\label{eq_posterior}
\end{equation}
\noindent
Moreover, the components $x^{(1)},\dots,x^{(C)}$ are mutually 
orthogonal, sum to $x$, and are mutually independent.
\end{proposition}
The proof is given in Supplementary Information 'Posterior factorization under orthogonal superposition'. Proposition \ref{eq_prop_posterior} has two consequences for the inference model of VDA . First, given $x$ the latent components are conditionally independent, so that no conditional dependencies between the components need to be modelled. Then, inference reduces to $C$ separate problems (one per component) instead of a single joint posterior with conditionally dependent factors; the decomposition model performs the separation that would have to be learned by a single encoder otherwise. Alternatively, this would either require multiple inference models, one for every latent variable, or hierarchical relations between the latent variables, as is done in \cite{Hsu2017UnsupervisedData}. Second, each of the $C$ problems needs only its own component $x^{(i)}$. The decomposition of $x$ into orthogonal components that reconstruct it (the input domain counterparts of Eqs.~(\ref{eq_ortho_generative_space}) and~(\ref{eq_recon_generative_space}) supplies these conditioning variables, so that an encoder applied separately to each component does not have to learn to separate the components itself. It should be clarified that Eq.~(\ref{eq_posterior}) is a property of the true posterior $p_\theta$ of the generative model, not of the network. Then, the VDA inference model is given the same structure, one factor per component, conditioned only on that component:

\begin{equation}
q_\varphi(z_1,\dots,z_C\,|\,x)=\prod_{i=1}^{C} q_\varphi\big(z_i\,|\,x^{(i)}\big),\qquad
q_\varphi\big(z_i\,|\,x^{(i)}\big)=\mathcal{N}\Big(z_i \;\Big|\; f_{\mu_{z_i}}\big(x^{(i)}\big),\; \mathrm{diag}\, (f_{\sigma^{2}_{z_i}}\big(x^{(i)}\big))\Big),
\end{equation}
\label{inference_model}
\noindent


where $f_{\mu_{z_i}}, f_{\sigma^2_{z_i}}$ are neural networks that share the set of parameters $\varphi$, except their final linear projections (see `Variational decomposition autoencoder'); there is no conditioning between components in the inference model.
Because parameters $\varphi$ are optimized through the VDA objective (see 'Variational decomposition autoencoding objective') rather than by maximizing a likelihood term, Proposition~\ref{eq_prop_posterior} specifies the structure of the inference model, but not how closely it approximates the posterior. It should be noted that DecVAE also encodes the original signal $x$ through the same encoder, and utilizes its embedding $h$ as an anchor for the self-supervised decomposition objective detailed right below; neither $x$ nor $h$ is part of the generative model of Eq.~(\ref{eq_joint_probability}) that is defined solely by the components. Including its latent projection $\tilde{z}_0$ benefits the performance of the aggregated representation when included (see 'Aggregation of latent subspaces', Supplementary Fig.~26).

\subsection*{Self-supervised latent decomposition}

Proposition~\ref{eq_prop_posterior} requires the components $x^{(i)}=P_i\,x$, however in real data the subspaces $V_i$ are unknown. To that end, a decomposition model $\mathcal{D}^C_w$ parameterized by $w$ is used to approximate the projections $P_i$. The decomposition model can operate in the observed input domain $X$ as a front-end to the inference model $q_\varphi$, or can be a part of the inference model $q_\varphi$ and operate directly on the latent space $\tilde{Z}$ or any intermediate space $H$. Here we present variational decomposition autoencoding with the former approach. 

We define a decomposition model $\mathcal{D}_w^C$ as an iterative optimization algorithm with a stopping criterion over $C$ iterations, that is applied on input time series observations $x_n$ of the dataset $X\in \mathbb{R}^{K\times N}$ and yields a set of components $\{x_n^{(1)},x_n^{(2)},...,x_n^{(C)}\}$. The number of components $C$ is a hyperparameter; since overestimating $C$ is not detrimental to performance and underestimating it is (see Fig.~\ref{fig8}c-e), it should be set at or above the number of spectrally distinct structures expected in the data. The oscillatory components yielded by $\mathcal{D}_w^C$ may have distinct time and frequency domain characteristics and their nature will depend on the inductive biases of the decomposition algorithm, but they must satisfy, at least approximately, the properties of orthogonality and reconstruction of Eq.~(\ref{eq_ortho_generative_space}) and Eq.~(\ref{eq_recon_generative_space}) in the input observation domain, where $\langle x^{(i)},x^{(j)}\rangle=0$ 
for $i\neq j$ and $\sum_i x^{(i)}=x$ denote the input domain counterparts to Eqs.~(\ref{eq_ortho_generative_space}) and Eq.~(\ref{eq_recon_generative_space}), respectively. To the degree that these hold, $\mathcal{D}_w^C$ approximates the projections $P_i$ and the factorization of Proposition~\ref{eq_prop_posterior} applies. Such an example is adopted in DecVAE by using band-pass filters centered on separate spectral regions, thus approximating orthogonal projections onto band-limited subspaces (see 'Decomposition model and self-supervised learning masking').
Yet, when real data processing is involved, these properties are satisfied in the domain where $\mathcal{D}_w^C$ operates, in this case the input time and frequency domain, and are not guaranteed to be satisfied in the embedding space $H$. 

To address this, we design a self-supervised contrastive latent decomposition optimization algorithm that leverages the structural biases of orthogonality and reconstruction enforced by $\mathcal{D}_w^C$ in the input domain, and aims to retain these structural dynamics in subsequent latent spaces $H$. Specifically, the following contrastive objective is formulated and minimized:

\begin{equation}
\min_{\varphi}\; D_{recon}-D_{ortho},\qquad
D_{recon}=\sum_{i=1}^{C}D\big(h_i\,\|\,h\big),\quad
D_{ortho}=\sum_{i<j}D\big(h_i\,\|\,h_j\big),\quad
h_i=f\big(x^{(i)}\big),\; h=f(x),
\label{eq_ssl_adversarial}
\end{equation}
\noindent
where $D$ is a function that measures divergence between the distributions of components $h_i$ and original signal $h$ in a space $H$. Specifically, the $D_{recon}$ term enforces similarity of all components to the original signal in $H$, and the $D_{ortho}$ term enforces the orthogonality or dissimilarity between each pair of components in $H$. The combination of those two terms enforces the reconstruction and orthogonality properties. The formulation in Eq.~(\ref{eq_ssl_adversarial}) denotes an adversarial objective with two competing terms, as an equilibrium between attracting and repelling forces between the components and the original signal is sought in the optimization problem. 

In practice, a function $D$ must be carefully chosen to avoid representation collapse that arises as a behavior of a neural network optimizer in seeking trivial solutions that can minimize a contrastive learning objective function in Eq.~(\ref{eq_ssl_adversarial}) \cite{Jing2021UnderstandingLearning}. To that end, we use the JSD and to stabilize the adversarial objective in Eq.~(\ref{eq_ssl_adversarial}) we calculate the following quantities:

\begin{align}
s^{pos}_{n,i}&=\min\Big\{1,\sum_{f\in F_n}D_{JS}\big(r_{n,i,f}\,\|\,p_{n,f}\big)\Big\}, &
s^{neg}_{n,ij}&=\min\Big\{1,\sum_{f\in F_n}D_{JS}\big(r_{n,i,f}\,\|\,r_{n,j,f}\big)\Big\}, \nonumber\\
L_{recon}&=\frac{1}{B}\sum_{n=1}^{B}\sum_{i=1}^{C}w^{i}_{pos}\Big[-\log\big(1-s^{pos}_{n,i}\big)\Big], &
L_{ortho}&=\frac{1}{B}\sum_{n=1}^{B}\sum_{i<j}w^{(i,j)}_{neg}\Big[-\log s^{neg}_{n,ij}\Big], \nonumber
\\ & \qquad \qquad \qquad \qquad \qquad \qquad \qquad &\mathcal{L}_{SSLDec} = L_{recon} + L_{ortho}
\label{eq_decomp_loss_compute}
\end{align}
\noindent
where $p_{n,f} = \mathrm{softmax}(h_{n,f})$ and $r_{n,i,f} = \mathrm{softmax}(h_{n,i,f})$ are normalized softmax embeddings over the $d$ hidden dimensions of $H$, for the original signal and component 
$i$, respectively, in frame $f$ of utterance $n$; $F_n$ is the set of frames sampled from utterance $n$, and $B$ is the number of utterances in a mini-batch. Moreover, $D_{JS}(a\|b) = \tfrac12 D_{KL} (a\|m) + \tfrac12 D_{KL}(b\|m)$, with $m=\tfrac12(a+b)$, is the Jensen-Shannon divergence \cite{Menendez1997TheDivergence} computed in nats, and $w^i_{pos}, w^{(i,j)}_{neg}$ are hyperparameters that regulate the weighting of these divergences. In the case that an utterance contributes a single embedding, i.e. $|F_n|=1$, $s\le\ln2 < 1$. For $|F_n| >>1$, the frame-wise divergences are summed over the sampled frames and clipped to $[0,1]$. The latter scenario is the standard DecVAE scenario that is encountered (frame-wise), whereas the sequence embeddings of DecVAE (see 'Multiple latent variables') correspond to the former.

In Eq.~(\ref{eq_decomp_loss_compute}), $\mathcal{L}_{SSLDec}$ consists of the two binary cross-entropy terms $L_{recon}$, $L_{ortho}$, each one containing a Bernoulli variable with parameter $s$, and target 0 for the (component, original) pairs and target 1 for the (component, component) pairs. This formulation sets $\mathcal{L}_{SSLDec}$ as the negative log-likelihood of the pair labels induced by the decomposition under a normalized Bernoulli model.

Formulating the two adversarial JSD quantities in Eq.~(\ref{eq_ssl_adversarial}) as two cross-entropy terms stabilizes the adversarial objective of Eq.~(\ref{eq_ssl_adversarial}) in three ways. First, JSD replaces KLD as a quantity that is bounded ($0\le D_{JS}\le\ln 2$), whereas maximization of the quantity $D_{ortho}$ rewards unbounded growth of KLD towards infinity. In addition, the phenomenon of representation collapse (i.e. when $h_i = h$ for all $i$) is penalized, as in that case $s^{neg}$ approaches 0, and hence $L_{ortho}$ approaches infinity, a direction that does not minimize $\mathcal{L}_{SSLDec}$ and is thus penalized strongly. Finally, the reconstruction divergences can vanish only if all components coincide with the original, where the separation term is infinite (Proposition~\ref{prop_equilibrium}(c)), and hence this objective has no trivial minimizer; Proposition~\ref{prop_equilibrium} characterizes the nature of its minimizer.
The complete derivation of Eq.~(\ref{eq_decomp_loss_compute}) and the algorithm to train a neural network with this loss is given in Supplementary Section~5 `Supplementary Material: Self-supervised contrastive decomposition loss derivation and algorithm' and Supplementary Alg.~3.

\subsection*{Variational decomposition autoencoding objective}
\label{vda_objective}

The VDA objective (Eq.~(\ref{vda_overview})) complements the self-supervised decomposition loss of Eq.~(\ref{eq_decomp_loss_compute}) with a prior regularization term on the component subspaces, and expressed as the following minimization task:

\begin{equation}
\mathcal{L}_{VDA}=L_{recon}+L_{ortho}+\beta L_{prior},\qquad
L_{prior}=\frac{1}{|\mathcal{I}|}\sum_{n=1}^{B}\sum_{f\in F_n}\sum_{i\in\mathcal{I}}D_{KL}\Big(q_\varphi\big(z_i\,|\,x^{(i)}_{n,f}\big)\,\Big\|\,p_\theta(z_i)\Big),
\label{eq_vda}
\end{equation}

where $\mathcal{I}$ denotes the per component regularized latent projection heads $H\to \tilde{Z}$, $x^{(i)}_{n,f}$ is the frame $f$ of component $i$ of utterance $n$, $F_n$ is the set of sampled frames in Eq.~(\ref{eq_decomp_loss_compute}) (in the sequence variable $|F_n|=1$). Unlike decomposition terms that are averaged over $B$ utterances, the KL terms are summed over the utterances and frames of the batch. $\beta$ is a hyperparameter that controls the pressure on the distribution to match the prior \cite{Higgins2017Beta-VAE:Framework}. The KLD terms of $L_{prior}$ in Eq.~(\ref{eq_vda}) are computed from the means and variances of the Gaussian projection heads, and the means are the latent representations that DecVAE outputs. In contrast to classical variational autoencoding \cite{Kingma2014Auto-EncodingBayes}, in the absence of a decoder, we do not draw samples from the approximated posterior $q_\varphi$ to use as the representation. The collective optimization objective combining $L_{recon},L_{ortho},L_{prior}$ gives rise to variational decomposition autoencoding, graphically illustrated in Fig.~\ref{fig1}c. We write DecAE for $\beta=0$, DecVAE for $\beta=1$, and $\beta$-DecVAE for other values of $
beta$; when referring to the method in general, we use DecVAE.

\subsection*{Equilibrium of the decomposition terms in VDA}
\begin{proposition}
\label{prop_equilibrium}
Let $p$ and $r_1,\dots,r_C$ be the normalized embeddings of an observation and of its $C$ components over $d\ge C$ embedding dimensions, with equal weights $w^{i}_{pos}$ (see Eq.~(\ref{eq_decomp_loss_compute})).
(a) \emph{Separation.} $L_{ortho}$ is minimal if and only if the components' mass is allocated on disjoint sets of embedding dimensions, i.e., $\langle r_i, r_j \rangle = 0, \forall i\neq j$; then $D_{JS}(r_i\|r_j)=\ln2$.
(b) \emph{Reconstruction.} For components as in (a), $L_{recon}$ is minimal over $p$ if and only if the original is the even mixture of its components, i.e., $p = \frac{1}{C}\sum_i r_i$; then $D_{JS}(r_i\|p)=j_C>0, \forall i$.
(c) \emph{Trade-off.} For any $i\neq j$, $D_{JS}(r_i\|r_j)\le 2\,\big[D_{JS}(r_i\|p)+D_{JS} (r_j\|p)\big]$. Hence $L_{recon}=0$ requires $r_i=p$ for all $i$, in which case $L_{ortho}\to \infty$. Also, two components separated by a divergence $s$ have reconstruction divergences that sum to at least $s/2$.
\end{proposition}

In Proposition~\ref{prop_equilibrium}, $j_C=\tfrac12\big[\ln\tfrac{2C}{C+1}+\tfrac1C\ln\tfrac{2}{C+1}+\big(1-\tfrac1C\big)\ln 2\big]$, e.g. for $C=3$ components, $j_3\approx0.318$. When the divergences of several frames per utterance in a mini-batch are summed (as in Eq.~(\ref{eq_decomp_loss_compute})), Proposition~\ref{prop_equilibrium}c holds for the sums. In summary, Proposition~\ref{prop_equilibrium} analyzes the counterparts, in $H$, of the properties of Eqs.~(\ref{eq_ortho_generative_space}) and (\ref{eq_recon_generative_space}). At the optimum of these equations, components are orthogonal in the normalized embedding space $H$ and the original is a mixture as their latent reconstruction. The proofs for the parts of this proposition are given in Supplementary Information 'Equilibrium of the decomposition terms'.

\paragraph{What makes VDA variational and autoencoding.} VDA can be termed  ``variational" for $\beta > 0$ in a restricted sense, if viewed under the prism of how variational is defined in $\beta$-VAE \cite{Higgins2017Beta-VAE:Framework}. In VDA, just as in $\beta$-VAE, each component subspace has a Gaussian projection latent whose distribution is regularized to approximate the prior by the KLD term of Eq.~(\ref{eq_vda}), which acts as a rate penalty, and the mean of this approximation is used as the representation. Unlike VAE and $\beta$-VAE, VDA performs autoencoding at the level of distributions in latent embedding space, and DecVAE has no decoder and evaluates no likelihood $p_\theta(x|z)$. Instead, VDA enforces reconstruction between the distributions of the components' embeddings and of the original's; as shown in Proposition~\ref{prop_equilibrium}(b), at the optimum the original's distribution is exactly the mixture of the components' distribution. $\mathcal{L}_{VDA}$ should not be confused with an evidence lower bound (ELBO), as VDA contains a self-supervised latent decomposition term but no likelihood term; the $\mathcal{L}_{SSLDec}$ is computed on the embeddings and does not depend on latent samples, as is the case in the likelihood in ELBO.

\paragraph{Plateau of the VDA training curves.} Proposition~\ref{prop_equilibrium}(c) predicts settlement points for the decomposition terms of $\mathcal{L}_{SSLDec}$ that converge to a plateau with a strictly positive reconstruction term after separation is achieved, as evident by the behavior of the training curves in Supplementary Fig.~22. On SimVowels (Supplementary Fig.~22b) the plateau values of the sequence-level (S branch) divergences are close to the predictions of Proposition~\ref{prop_equilibrium}(a) and (b): for a filter-decomposition model and $\beta=0$ divergences reach 0.68 to 0.69 nats, close to the predicted $\ln2\approx0.693$, and those between each component and the original to about 0.32 to 0.33, close to the predicted $j_3\approx0.318$. An evaluation on the held-out test set of SimVowels with EWT confirms this equilibrium, whereas it shows its partial fulfillment with FD and on real speech data on TIMIT (Supplementary Information 'Equilibrium of the decomposition terms', Supplementary Fig.~32). Ultimately, VDA parameters such as $\beta$, the choice of decomposition model $\mathcal{D}_w^C$ and type of input features, can affect this plateau, as seen in the analysis of Supplementary Fig.~22. To relate $\beta$ to separation (Fig.~\ref{fig7}a) we evaluated every trained checkpoint and computed per utterance in the sequence branch, the JSD on which $L_{ortho}$ and $L_{recon}$ act (Eq.~\ref{eq_decomp_loss_compute}).

\subsection*{Multiple latent variables}
DecVAE can simultaneously function as a single or multiple-branched model, depending on the number of time-scales expected in the problem at hand. Each branch acts as a recognition model $\varphi$, parameterized as a deep convolutional neural network that accepts as input sequences.  
Stochastic processes that construct complex non-stationary time series such as speech, often involve generative factors that can operate at the same or different time scales and thus a sequence can contain slow and fast varying attributes. DecVAE is invariant to the duration of the sequence and can accommodate multiple branches, where each branch handles a specific time-scale. Fig.~\ref{fig1}d demonstrates a dual branched DecVAE for speech, with the sequence branch operating in the timescale of seconds (utterances) and the frame branch operating in the timescale of milliseconds (phonetic content). 
As has been previously shown in speech-centric VAE designs (MFL-VAE \cite{Boulianne2020ALearning}, FHVAE \cite{Hsu2017UnsupervisedData}), speaker identity characteristics such as fundamental frequency or pitch $F0$, and volume have little to no variability across a sequence, whereas phonetic content has high variation within a sequence and is also affected by the speaker identity. 
To accommodate different time scales, we update the generative and inference models in the problem formulation by adding a second set of latent variables $\{s, s_1, s_2, ..., s_C\}$ (Supplementary Fig. 3
). Two different recognition models $\varphi_Z,\varphi_S$ are used to infer each timescale latent variable separately. It can be seen that $\varphi_{S}$ operates in a plane of $N$ sequences, whereas $\varphi_{Z}$ operates in a domain of $F$ frames for the current sequence $n$, making the latent variable $S$ constant for the current $n$-th sequence. 

It has been previously shown \cite{Boulianne2020ALearning} that variational autoencoding can be extended to a second latent variable by assuming independency of generative factors in $Z$ and $S$, and by applying a pooling operation over the frames of each input sequence $S$ to end up with a single latent $s_n$ for the $n$-th sequence. Accordingly, the VDA objective for two latent variables is the sum of the objectives of the two branches:

\begin{equation}
\mathcal{L}_{VDA}^{dual}=L^{Z}_{recon}+L^{Z}_{ortho}+\beta_Z\gamma_Z 
L^{Z}_{prior}
+L^{S}_{recon}+L^{S}_{ortho}+\beta_S\gamma_S 
L^{S}_{prior},
\label{eq_vda_dual}
\end{equation}
\noindent
where the superscripts denote the frame ($Z$) and sequence ($S$) branches, with each term defined as in Eqs.~(\ref{eq_decomp_loss_compute}) and (\ref{eq_vda}), and $\gamma_Z$, $\gamma_S$ are fixed constants that balance the magnitudes of the terms (see Supplementary Table~4).

In our experiments, we use the sequence-level variable $S$ as support for variables that can be assumed to be invariant in the frame-level, such as speaker identity and emotion (Figures~\ref{fig2},\ref{fig5}), as a natural disentanglement mechanism due to the design of the network. In terms of optimization, we see that the sequence branch behaves similarly to the frame branch in terms of training and convergence of the VDA objective (Supplementary Fig. 22
b,c). Prior approximation slows down convergence in the sequence branch as well, an effect that is apparent in the divergence and prior approximation loss terms (Supplementary Fig. 22
b). Moreover, training a DecVAE with two branches has an influence on the $Z$-branch, as $Z$ is influenced by $S$ and this is evident through a slower convergence of $Z$ in the case of a dual model, as $Z$ has to also assimilate information from $S$ in the joint training scenario (Supplementary Fig. 22
c).

\subsection*{Decomposition model and self-supervised learning masking}
\label{decomp_module}

As was analyzed in 'Self-supervised latent decomposition', the decomposition model $\mathcal{D}_w^C$ is a key component in DecVAE; in our implementations we position $\mathcal{D}_w^C$ as the first processing layer in the DecVAE architecture (Supplementary Fig. 2
). $\mathcal{D}_w^C$ is a signal processing module that can be incorporated into deep learning models, following a recent trend in the time-series representation learning literature \cite{Woo2022CoST:Forecasting, Oreshkin2019N-BEATS:Forecasting, Zhou2022FEDformer:Forecasting}. $\mathcal{D}_w^C$ receives a mini-batch $X=\{x_1,x_2,...,x_B\}$, where $B$ is the batch size and returns for each $x_b$ the set $\{x_b^{(0)},x_b^{(1)}, x_b^{(2)},...,x_b^{(C)}\}$ with $x_b^{(0)} \equiv x_b$. As hyperparameters $w,C$ dictate the type of inductive bias that we use to achieve a satisfactory separation in the input domain, we experiment with three different state-of-the-art decomposition algorithms: empirical wavelet transform (EWT) \cite{gilles2013ewt}, empirical mode decomposition (EMD) \cite{huang1998empirical}, variational mode decomposition (VMD) \cite{dragomiretskiy2014vmd}. Incorporation of these methods into deep learning models requires a fast and robust approach that can be applied at scale across large datasets. To that end, we also design a simple filter-based decomposition (FD) that is not based on some iterative optimization objective (Supplementary Alg.~4
). FD detects spectral peaks inside fixed peak detection intervals and places an infinite-impulse response (IIR) band-pass filter around each detected peak, so the filter bands follow the spectral content of each frame. When more than $C$ peaks are detected, components are merged with a pre-defined strategy: for real data, we merge components from higher frequency intervals and keep lower frequency components intact; for simulations, we merge components with their nearest neighbours. When fewer than $C$ peaks are detected, the missing components are set to zero. The results are insensitive to the placement of the intervals (Fig. 8f). The FD algorithm is given in Supplementary Alg.~4 and its hyperparameters in Supplementary Table~5.

FD is a fast approach to apply over big volumes of data as it only relies on simple filtering. Moreover, as the disentanglement properties of the latent space are affected by the input decomposition, FD yields the most orthogonal components among the methods we tested, since its band-pass filters cover separate frequency regions and are therefore approximately orthogonal in the frequency domain, in contrast to time-frequency approaches like the EWT and time-domain approaches like the EMD. This property of FD prevents correlation of the components in the input domain that can propagate in the latent space and worsen the disentanglement quality (Supplementary Fig. 9
). $\mathcal{D}_w^C$ with FD is able to successfully recover properties of real and simulated formants, indicating good performance across simulated and real speech alike (Supplementary Fig. 4
e,f,g,h). Hyperparameters and their values for all used decomposition methods are presented in Supplementary Table~5. 
Most importantly, FD is ideal for SSL as it is applying a ``frequency-domain masking", as filters ``mask" the frequency bands around the detected peaks. $\mathcal{D}_w^C$ incorporates the concept of masking as an augmentation for the success of SSL training (Supplementary Alg. 3
). Masking has been successfully applied as a pre-text task in SSL to create alternate views of an input that are then fed into a neural network, accompanied by an objective that encourages similarity of representations of different views of the same input \cite{balestriero2023} and has been found to be critical in learning informative representations. Here, the ``mask" is applied in the frequency domain of the input, and the alternate views of the input constitute the components; this allows us to formulate ``positive" and ``negative" pairs, between the original signal and the components and enforce their similarity and dissimilarity in the latent space, and calculate the training objective losses based on that as was shown in `Self-supervised latent decomposition'. As it has also been shown in the SSL literature \cite{Baevski2020Wav2vecRepresentations, Mohamied2024BootstrapApproach, grill2020}, we validate that it is beneficial to ``mask" and subsequently use about $50\%$ of the available data at each step for the loss calculation (Supplementary Fig. 25
).

\subsection*{Aggregation of latent subspaces}
\label{aggregation}
As was mentioned in the introductory part of Methods, the representation $\tilde{Z}$ requires an aggregation operation over the decomposed subspaces $\{\tilde{z}^0, \tilde{z}^1, ..., \tilde{z}^C\}$. 
We experimented with different aggregation functions, i.e.: i) using individual subspaces and discarding the others, ii) concatenation of all components' subspaces, i.e., $\tilde{Z} = [\tilde{z}^1, \tilde{z}^2,..., \tilde{z}^C]$, iii) concatenation of all components' and the original's subspaces  $\tilde{Z} = [\tilde{z}^0, \tilde{z}^1, \tilde{z}^2,..., \tilde{z}^C]$, and iv) a learned projection of the concatenated subspaces, i.e., $\tilde{Z} = a([\tilde{z}^1, \tilde{z}^2,..., \tilde{z}^C])$, where $a(\cdot)$ is parameterized as a two-layered fully-connected network with non-linearity. Our results on our simulated data suggest that option (iii) yields an embedding of higher quality across all disentanglement and classification metrics, 
for different decomposition methods (FD,EWT) and values of prior approximation pressure $\beta = (0,0.1,1)$ (Supplementary Fig. 26
). 

\subsection*{Optimization and Pre-training}
DecVAE models are trained on a single NVIDIA RTX 6000 Ada Generation hardware with a dual learning schedule and a batch size of about $1$ minute of data. The sequence branch $S$ trains with a linear schedule up to a peak learning rate of $1*10^{-4}$, whereas the frame branch $Z$ trains with a constant schedule to a peak learning rate of $8*10^{-5}$; in both branches the learning rate is warmed up to its maximal value for 20 epochs and in the $S$ branch the learning rate is decayed after that. The Adam optimizer is used \cite{kingma2014Adam} and training is monitored for early stopping by keeping track of improvement $\tau_\delta$ with a patience of $\tau_{patience}=5$ epochs on the total loss in the validation set requiring $\tau_\delta>0.2\%$, after an initial period of $\tau_{warmup}=100$ epochs. A maximum number of $t_{max} = 150$ epochs is set, but most models converge around 100-110 epochs (Supplementary Fig. 22,24
). Each training example is a single utterance: utterances longer than $d_{max}$ are truncated to their first $d_{max}$ seconds (for VOC-ALS, this is done after skipping the first second of each recording), and shorter ones are zero-padded to $d_{max}$ with the padded frames marked as masked and kept out of the decomposition and the loss. No utterances are concatenated, so every example keeps its own task-related labels. The values for $d_{max}$ are given in Supplementary Table~4.

\subsection*{Variational decomposition autoencoder}
\label{DecVAE_methods}
In the previous Methods sections we laid out the theoretical background and the functionality of important components in VDA and practical considerations for constructing a suitable optimization objective for learning disentangled representations. Here, we present how the collective functionality of individual components is organized into the DecVAE neural network architecture (Supplementary Fig. 2
). DecVAE parameterizes the inference model $q_{\varphi}$ as a parameter collection $\varphi$ of a convolutional encoder and two projectors, with the decomposition model $\mathcal{D}_w^C$ acting as a fixed recognition front-end; $\mathcal{D}_w^C$ decomposes input frames from $X$ into $C$ components in $X'$ and applies a random masking operation that marks a user-specified percentage of frames $l_{SSL\%}$ to be used in the VDA objective in Eq.~(\ref{eq_vda}). Mel filterbank features \cite{Hsu2017UnsupervisedData, Boulianne2020ALearning} with a filterbank size $80$ are then extracted for all components and the original signals in the mini-batch $B$, as they were found to perform better on downstream tasks than raw waveforms (Supplementary Tables~7-9
). A Wav2Vec2-based \cite{Baevski2020Wav2vecRepresentations} shared encoder $f_W(\cdot)$ is then utilized for further feature extraction and projection of the signals through a seven-layered convolutional stack with LayerNorm normalizations \cite{Ba2016LayerNormalization} and Gaussian Error Linear Units (GELU) \cite{Hendrycks2016GaussianGELUs} in an intermediate space of dimension $v$. A shared projector $f_H(\cdot)$ of two fully-connected layers with GELU non-linearity projects the encoder output in the space $H$ of dimension $d$ where the $L_{recon}$ and $L_{ortho}$ are calculated by contrasting embeddings according to Eq.~(\ref{eq_decomp_loss_compute}). Then, $C+1$ latent projections $f_L(\cdot)^{(C+1)}$ with fully-connected linear layers $mu, logvar$ projects the embeddings to their latent subspaces $\tilde{z}_0, ..., \tilde{z}_C$ where the prior approximation term of the VDA objective (Eq.~(\ref{eq_vda})) is enforced for every component subspace of dimension $z_{dim}$ separately. Finally, the aggregator $a(\cdot)$ projects the subspaces to the final representation $\tilde{Z}$ of dimension $z_{dim}(C+1)$. It should be noted that all learnable parameters except the latent projectors $f_L(\cdot)^{(C+1)}$ are shared for all components and the original signal, i.e., there is a single branch that handles all inputs. For a standard DecVAE single latent variable $Z$ architecture, we choose values $v=512$, $d=192$, $z_{dim}=48,64$ for simulated and real speech, respectively, and $C+1$ by using as an aggregation function the concatenation of all latent subspaces. In Algorithm \ref{alg: DecVAE} we describe the training process of DecVAE under the most successful experimental set of parameters (Supplementary Table~4).

\begin{algorithm}[ht!]
\caption{DecVAE}
\label{alg: DecVAE}
\begin{algorithmic}[1]

\Statex \textbf{Input:} $X$
\Statex \textbf{Params:} $d_{max}$, $d_{min}$, $\mathcal{D}_w^C$, $C$, $\beta$, $l_{SSL\%}$, $z_{dim}$, $t_{max}$, $\tau_{warmup}$, $\tau_{\delta}$, $\tau_{patience}$
\Statex \textbf{Output:} $Z$
\vspace{0.25em} 
\State Discard sequences in $X$ smaller than $d_{min}$ 
\State Zero-pad sequences in $X$ shorter than $d_{max}$ to $d_{max}$, and mark the padded samples in an attention mask
\State Truncate sequences in $X$ longer than $d_{max}$ to their first $d_{max}$ seconds
\State Normalize sequences to zero mean and unit variance
\State $t_{current} \gets 0, \tau_{count} \gets 0$
\State $l_{best} \gets \infty$
\State Initialize $f_W(\cdot)$, $f_H(\cdot)$, $f_L(\cdot)^{(C+1)}$ weights with He initialization
\While{$t_{current} < t_{max}$ and $\tau_{count} < \tau_{patience}$}
    \State $x_b \gets next(X), x_b \in R^{B,F,L}$
    \State $x'_b \gets \mathcal{D}_w^C(x_b), mask \gets sample(x_b,l_{SSL\%}), x_b'\in R^{B,C+1,F,l}$
    \State $x'_b \gets Mel(x'_b), x_b'\in R^{B,C+1,F,80*5}$
    \State $v_b \gets f_W(x'_b) \in R^{B,C+1,F,v} $
    \State $h_b \gets f_H(v_b) \in R^{B,C+1,F,d} $
    \State $h^{mask}_b \gets h_b(mask)$
    \State Calculate $L_{recon}, L_{ortho}$ on $h^d_b$ via Eq.~(\ref{eq_decomp_loss_compute})
    \State $\tilde{z}_b \gets f_L(h_b)^{(C+1)}  \in R^{B,C+1,F,z_{dim}}$ 
    \State $\tilde{z}^{mask}_b \gets \tilde{z}_b(mask)$
    \State Calculate $L_{prior}$ on $z^{z_{dim}}_b$ via Eq.~(\ref{eq_vda})
    \State $\mathcal{L}_{VDA} \gets L_{recon} + L_{ortho} + \beta*L_{prior}$
    \If{$t_{current} \geq \tau_{warmup}$}
        \If{$\frac{|{L}_{VDA} - l_{best}|}{l_{best}} > \tau_{\delta}$}
            \State $l_{best} \gets \mathcal{L}_{VDA}$
            \State $\tau_{count} \gets 0$
        \Else
            \State $\tau_{count} \gets \tau_{count} + 1$
        \EndIf
    \EndIf
    \State Update $f_W(\cdot)$, $f_H(\cdot)$, $f_L(\cdot)^{(C+1)}$ using Adam on $\mathcal{L}_{VDA}$
    \State $t_{current} \gets t_{current} + 1$
\EndWhile
\State $\tilde{Z} \gets a(f_L(f_H(f_W(Mel(\mathcal{D}_w^C(X))))))$
\end{algorithmic}
\end{algorithm}

\subsection*{Disentanglement evaluation}
To assess the disentanglement quality of the learned latent spaces we employ an array of different disentanglement metrics; disentanglement is an elusive and multi-faceted property and disentangled representation learning literature has given multiple definitions and ways of interpreting disentanglement \cite{Locatello2019ChallengingRepresentations, carbonneau2022dis_metrics,suter2019irs}, with different metrics measuring different properties that a disentangled representation should have. Formulas for the utilized metrics are provided in Supplementary Information `Supplementary Note: Disentanglement metrics'. 

\paragraph{Mutual information.} We calculate the mutual information (MI) metric \cite{carbonneau2022dis_metrics} between all pairs of latent dimensions in $Z$. Although this metric is unsupervised and cannot be linked to the disentanglement performance across specific generative factors, it can give us information on how well the information is allocated across different dimensions in the latent spaces. The mutual information is calculated on $\tilde{Z}$ from the evaluation sets, based on discretization of the latent space through a histogram with 30 bins. MI is lower bounded by 0 and a representation with a MI close to 0 has dimensions with different information content.

\paragraph{Gaussian correlation.} As a second unsupervised metric we calculate the normalized total correlation based on fitted Gaussian \cite{Locatello2019ChallengingRepresentations} or Gaussian correlation normalized (GCN). The covariance matrix and the mean vector of $\tilde{Z}$ after encoding the whole evaluation set is calculated, and the KL divergence is calculated between a Gaussian with covariance matrix and mean vector equal to $\tilde{Z}$, and the product of this Gaussian's marginals. We scale this quantity by the product of the dimension of $Z$ with the average covariance for a fair comparison between models with different dimensionality in $Z$. GCN is lower bounded by 0 and a representation with a GCN close to 0 has less correlated dimensions.

\paragraph{Disentanglement, completeness, informativeness.} Disentanglement, completeness and informativeness (DCI) is a framework that was proposed as a complete tool that assesses all properties of a disentangled representation through the three metrics \cite{Eastwood2018ARepresentations}. Disentanglement measures the property of modularity, that denotes independence between generative factors, when one of them varies. Completeness measures the property of compactness or completeness, that is maximized as fewer latent dimensions encapsulate information about single factors. Informativeness measures the property of information content or explicitness, that is exemplified when a generalizable relation is learned between factors (labels) and latent codes. DCI is a predictor-based framework, where predictors are trained to infer factors from the latent code to quantify informativeness, and the predictor's inner parameters are used to infer the importance of each dimension in the latent code in predicting each factor. All three metrics evaluate in the $[0,1]$ range.

\paragraph{Modularity and explicitness.} Modularity and explicitness \cite{Ridgeway2018LearningLoss} is another framework proposed to measure the properties of modularity (disentanglement) and explicitness (informativeness). For modularity, the MI is calculated between factors and latent dimensions. For each dimension, the maximum MI is compared to all other MI scores with other generative factors. This is repeated for every dimension and the final modularity is the average over all dimensions.
For explicitness, a Logistic Regression classifier is trained on the latent code per generative factor and the AUROC score is calculated. The final explicitness score is the average AUROC across all classes for all factors that is normalized to obtain a score from 0 to 1. Modularity and explicitness evaluate in the $[0,1]$ range.

\paragraph{Interventional robustness score.} Interventional robustness score (IRS) or robustness \cite{suter2019irs} has been proposed to isolate causal effects of a generative factor that are overlooked through other metrics due to confounding biases. Robustness is assessed by performing interventions (shuffling) on the generative factors associated with observations, while keeping observations of a specific factor fixed. High robustness is measured when changes in other factors do not affect the distance between the fixed-factors reference set and a second intervention set when other factors are changed. IRS evaluates in the $[0,1]$ range.

\subsection{Ranking of methods}
\label{ranking}
Each disentanglement metric is used to rank the models, according to three different schemes. Our main scheme, core set ranking, which is the mean rank over GCN, DCI-D, DCI-I and IRS, a set that contains one metric each from the independence, disentanglement, informativeness, and interventional metric families. In this scheme, the remaining metrics are not averaged, as Modularity has a high chance level (Supplementary Table~23, Supplementary Fig.~29e), MI is very close to zero for most models (Supplementary Fig.~29), Explicitness duplicates DCI-I and is less stable under label noise (Supplementary Fig.~31), and DCI-C measures the compactness of a representation, rather than factor separation. We also provide a ranking scheme containing all eight metrics and all metrics except MI, GCN (Supplementary Fig.~30). It should be noted that we exclude frozen SSL LAMs from the rankings as they were pre-trained on much larger speech corpora.

\subsection*{Metric sensitivity}
\label{metric_sensitivity}
Nine representation learners ($\beta$-DecVAE with EWT and FD, $\beta$-VAE, FHVAE, CoST, SFA, PCA, wav2vec2, and wav2vec2 with PCA) are evaluated on a $10k$ frame stratified (w.r.t. generative factors) subset of the held-out SimVowels sets (validation and test), to examine the effect of different latent dimensionality on the evaluation of disentanglement metrics, by also comparing with random latent Gaussian noise embeddings (Supplementary Fig.~29c,d,e, Supplementary Fig.~31). For wav2vec2 ($d=768$) only, subsets of its embeddings were randomly drawn with increasing dimension as a chance-level test. Seven of these models are also scored ($\beta$-DecVAE with EWT and FD, $\beta$-VAE, FHVAE, CoST, SFA, PCA) on a stratified 10,000-frame subsample of the SimVowels test set (the reference) and under factor imbalance (Zipf $a=1$ and $a=2$), correlated factors (Cramér's $V=0.3, 0.6$), label noise ($10\%$ or $30\%$ of labels flipped), coarsened speaker labels (reduce speakers to 5 groups to test partial observability of the label variable), and appended Gaussian noise dimensions ($d$ from 48 to 768, total latent dimension becomes $d+d$). All rankings are compared with Kendall's $\tau$, whereas chance levels were estimated by cyclically shifting the labels (5 shifts).

\subsection*{Baselines}
\label{baselines}
We compare DecVAE with nine baselines that cover the main alternatives for unsupervised and self-supervised representation learning in time series. (i) Frozen self-supervised pre-trained LAMs: wav2vec2~\cite{Baevski2020Wav2vecRepresentations}, WavLm~\cite{chen2022wavlm}, HuBERT~\cite{Hsu2021HuBERT:Units}, where we extract and evaluate frame level embeddings of dimensionality $768$ from the last transformer layer output without further training; due to their increased dimensionality and the observed effect on disentanglement metrics (see Supplementary Fig.~29 and `Metric sensitivity'), we also employ a PCA to reduce dimensionality and evaluate them at the dimensionality of DecVAE models ($z_{dim}=192$ on SimVowels, SimCoupled, $z_{dim}=320$ on TIMIT, $z_{dim}=240$ on IEMOCAP, VOC-ALS) . All LAMs receive waveform input of the same duration as other methods, and output latent transformer embeddings with the same frame length and receptive field of DecVAE. (ii) Time-series self-supervised and contrastive models: CPC~\cite{vandenOordDeepMind2018RepresentationCoding}, CoST~\cite{Woo2022CoST:Forecasting}, TCL~\cite{Hyvarinen2016UnsupervisedICA}, and TF-C~\cite{Zhang2022Self-SupervisedConsistency}; CPC, CoST and TCL are applied to the same Mel filterbank features of DecVAE and the other benchmark models, whereas TF-C is applied to the raw waveform due to the time-frequency consistency objective that it optimizes. (iii) The Sequential VAE model, FHVAE~\cite{Hsu2017UnsupervisedData}, with Mel filterbank input; we evaluate on its segment-level latent for a fair comparison with how DecVAE extracts frame embeddings, that is without information of other frames in the sequence. (iv) Linear slow feature analysis (SFA)~\cite{wiskott2002slow}, which is fitted in closed form on ordered, contiguous frames. It should be noted that several of the above representation learning baselines contain mechanisms that inform the representation beyond a single frame's information, something that is not the case in DecVAE's frame-level embeddings. SFA requires the whole utterance during fit, whereas TCL, CPC are pre-trained through self-supervised tasks that involve the prediction of frame position or the latent embeddings of other frames; in FHVAE the frame-level latent representation is also jointly modeled with the utterance-level latent. This could inflate frame-level results in these baselines. Moreover, in our evaluations on IEMOCAP, VOC-ALS, and SimCoupled, we carry only the best performing baselines on SimVowels and TIMIT instead of all nine baselines (see Table~\ref{tab:baselines}); since we have examined the effect of dimensionality reduction on frozen LAMs, as well as rankings between LAMs, we use the PCA reduced version of wav2vec2 (as the best performing LAM on SimVowels and TIMIT) on IEMOCAP and VOC-ALS, for efficiency. On VOC-ALS, the PCA projection of wav2vec2 and SFA are fitted on the target data, and therefore serve as non-zero-shot references.

The above models of (ii), (iii), (iv) are pre-trained on the training set of each dataset and evaluated on the development and test sets, whereas for SimCoupled, the correlated test set is used for our main comparisons (independent set results for SimCoupled are reported in Supplementary Table~20). On IEMOCAP and VOC-ALS, PCA, ICA, SFA and the PCA of the frozen wav2vec2 embeddings, are fitted without labels on all frames before the cross-validation split; the classifiers and metric probes are trained on the training folds only. Latent dimensionalities of these methods are set to the equivalent of one DecVAE subspace, wherever this is applicable; for instance, the dimensionality for the autoregressive latent of CPC~\cite{vandenOordDeepMind2018RepresentationCoding}, or the latent dimensionality of the segment variable in FHVAE~\cite{Hsu2017UnsupervisedData}, are set to $48$ for SimVowels, and $64$ for TIMIT. Similarly, for CoST~\cite{Woo2022CoST:Forecasting} that has a trend and a seasonality latent subspace, we use the same rule as in DecVAE and assign a dimensionality of $48$ to each one of them, leading to a total dimension of $96$ (equivalent to $C=2$ in DecVAE). To ensure a fair comparison, regarding the choice of optimizer, learning rates, and stopping criteria, we use the default training setups from each of the papers in (ii), (iii). Moreover, the convolutional encoders involved in CoST and wav2vec2 are identical to the DecVAE convolutional encoder. To evaluate all baselines at the sequence-level, we apply an average pooling operation over a sequence's frames to obtain a single latent embedding per sequence; baselines that already have a sequence-level representation, such as FHVAE, do not have their frame-level embeddings averaged. This is similar to how DecVAE's S-branch (sequence latent variable) aggregates frame information. We did not benchmark methods that require forecasting targets or labels for supervised pre-training, such as LaST~\cite{wang2022latent}, AutoFormer~\cite{Wu2021Autoformer:Forecasting}, and FEDformer~\cite{Zhou2022FEDformer:Forecasting}; we discuss the relation of these methods to VDA in Supplementary Information~11.8.3. Further details on implementation, hyperparameters, and evaluation protocol are given in Supplementary Section~11.8.

\subsection*{Task-specific evaluation}
We also evaluate our models in terms of their ability to learn meaningful representations for downstream classification tasks in the domains of speech recognition, emotion recognition and amyotrophic lateral sclerosis (ALS) dysarthria evaluation from speech. For the task-specific evaluation, we operate directly on the latent spaces $Z$ and $S$ by using simple linear (Logistic Regression) and non-linear (Random Forest, Support Vector Machine) classifiers to assign as little value as possible to the classifier and exemplify the quality of the learned representation.

\paragraph{Weighted and unweighted accuracy.} We use weighted accuracy to measure task-specific performance in our experiments in the tasks of: vowel recognition, speaker identification, phoneme recognition, emotion recognition, disease duration prediction, clinical stage prediction. For emotion recognition, we also calculate unweighted accuracy as a standard metric in the IEMOCAP~\cite{busso2008iemocap} dataset, to account for emotion imbalance~\cite{Antoniou2023DesigningIEMOCAP}.

\paragraph{Weighted and macro-F1 score.} We calculate weighted and macro-F1 score for the tasks mentioned in the previous accuracy paragraph. Real-world datasets have class distribution imbalances that cannot be captured by the accuracy metrics.

\paragraph{Confusion matrices and emotion per-class performance.} For emotion recognition in IEMOCAP~\cite{busso2008iemocap}, we also report the per emotion class precision, recall and weighted F1-score. The four emotion classes in the standard IEMOCAP evaluation protocol are angry, happy, neutral, and sad (see Supplementary Fig.~28, Supplementary Table~19).

\paragraph{Cross-validation and confidence intervals.} All classification tasks are performed in an outer five-fold cross-validation scheme with an inner three-fold cross-validation for classifier parameters selection, for five different classifier random initializations, except emotion recognition in IEMOCAP \cite{busso2008iemocap} that is performed in a leave-one-speaker-out (LOSpO) (ten folds), or leave-one-session-out (LOSessO) (5 folds), outer cross-validation scheme with different random initializations (five for the subsampled protocol and ten for the full protocol), in which one (LOSpO) or both (LOSessO) speakers of the held-out session and all their dialogues are unseen during training. SimVowels is also excluded from confidence interval calculation as the data and class distribution is balanced and uniform by design. The metrics at each fold and random state are used to calculate $95\%$ confidence intervals over all iterations of the classification. For the models' pre-training phase, we use a defined train and development set for all datasets; for SimVowels we use a $N=4000$ utterances training set and $N=500$ utterances development set. For pre-training on TIMIT \cite{garofolo1993timit} we use the standard TIMIT training and development sets. For fine-tuning on IEMOCAP we use a stratified speaker-emotion split and hold $90\%$ of the data as a training set and $5\%$ of the training set as a development set. In the evaluation phase we use the whole datasets. 

For disentanglement metrics DCI, modularity and explicitness, that include training of predictors on the latent representations, we perform a five-fold cross-validation for five different random initializations that dictate the data split into training and test sets. We calculate $95\%$ confidence intervals for these metrics across cross-validation iterations and random initializations. Our results showed an enhanced robustness of the examined metrics in using different subsets of the TIMIT \cite{garofolo1993timit}, VOC-ALS \cite{Dubbioso2024VoiceControls} and IEMOCAP \cite{busso2008iemocap} datasets for quantifying disentanglement in DecVAEs and supporting methods, manifesting with CIs below $\pm0.001$ on TIMIT, VOC-ALS, and full-protocol IEMOCAP (Supplementary Tables~10,11,14); in the subsampled IEMOCAP protocol, the CIs of DCI-disentanglement and completeness reach $\pm0.03$, whereas other metrics CIs stay below $\pm0.02$ (see Supplementary Table~16). 

\subsection*{Hyperparameters}
DecVAE involves selecting parameters both for the decomposition model $\mathcal{D}_w^C$ and the neural network. Values for all parameters for training DecVAE, supporting methods and baselines used in both simulations and real data processing are presented in Supplementary Tables~4-6.

\subsection*{Datasets}
We apply DecVAE and supporting VAE \cite{Kingma2014Auto-EncodingBayes}, $\beta$-VAE \cite{Higgins2017Beta-VAE:Framework}, ICA \cite{hyvarinen2001independent, pedregosa2011scikit}, PCA \cite{Greenacre2022PrincipalAnalysis, pedregosa2011scikit}, kernel-PCA \cite{Scholkopf1997KernelAnalysis} methods, as well as the baseline models in `Baselines' to four audio datasets plus SimCoupled.

\paragraph{SimVowels.} A simulated toy audio dataset inspired from \cite{Boulianne2020ALearning} that models speakers as variations in a vocal tract factor and vowels as unique superpositions of three narrow-band formant carriers; it features $N=4800$, $4$ second sentences from 60 simulated speakers. Each speaker randomly utters one of five vowels $(/a/,/e/,/I/,/aw/,/u/)$ with fast within-utterance one second alterations from vowel to vowel. SimVowels features two generative factors, the frame-level vowel uttered and the sequence-level speaker identity. The simulated vowels have the same structure and frequency-domain properties as real vowels \cite{Ladefoged2006APhonetics} (Supplementary Fig.~4
) and are thus a good fit to examine the ability of VAE models to disentangle factors of frequency-related generative processes. Details about SimVowels generation process are given in Supplementary Alg. 1.

\paragraph{SimCoupled.} An extension of the stochastic system of Ding et al.~\cite{ding200617}, in which an autoregressive resonator $X$ drives $Y$ through a delayed coupling; the observation signal is their sum plus white noise. In our implementation that yields the SimCoupled dataset, two frame-level factors change every $0.1 s$: the coupling lag ($k \in \{2, 4, 8, 16, 32\}$ samples), which shapes the whole spectrum through a comb filter, and the gain ($-4$ to $+4 dB$ in $2 dB$ steps), which shifts all bands equally. The factors are correlated (class correlation $0.77$) in the training, development and test splits (800, 100 and 100 sequences of $4\,s$ at $16 \,kHz$) and independent in a fourth split of 100 sequences used only for evaluation. Details on the generation and the characterisation of SimCoupled are given in Supplementary Section~3.

\paragraph{TIMIT.} A corpus of broadband read American English speech with $N = 4056$ utterances with a maximum duration of $7.15$ seconds, from 630 speakers \cite{garofolo1993timit}. Development of phoneme recognition models uses 48 phoneme classes in TIMIT, but evaluation is made by appropriate mapping on a smaller set of 39 classes \cite{Lee1989Speaker-IndependentModels}. TIMIT has also two known generative factors of phoneme and speaker identity. 

\paragraph{VOC-ALS.} An audio database of ALS patients and healthy controls for evaluating dysarthria severity \cite{Dubbioso2024VoiceControls}. VOC-ALS has $N=1224$ sentences with a maximum duration of $15$ seconds where $153$ participants perform a sustained repetition of the phonemes $(/a/, /e/, /i/, /o/, /u/, /pa/, /ta/, /ka/)$. The clinical variables in VOC-ALS contain rich information about disease duration, progression and other details, as well as scores in various questionnaires. We use VOC-ALS to test DecVAE in generative processes involving three known factors, that are phoneme, speaker and disease staging. We utilize the disease duration and King's clinical stage variables as disease staging indicators. 

\paragraph{IEMOCAP.} An audio database of acted and improvised conversations where actors engage in conversations while exhibiting a range of emotional responses \cite{busso2008iemocap}. IEMOCAP has $N=5531$ utterances of $7$ seconds maximum duration, from five two-person conversations and emotion annotations per utterance. We use IEMOCAP as another test for DecVAE in a three-factor generative process, where beside phoneme (48 phoneme classes) and speaker (10 speakers), the categorical emotion (angry,happy,neutral,sad) is the third generative factor.

\subsection*{Applications}
\paragraph{Disentanglement quality evaluation.} The ability of a model to learn disentangled representations that can separate underlying varying factors in complex, non-stationary and high-dimensional data such as time-series, is an important property that can amortize the data requirements and scale of neural networks in various tasks. Disentanglement of a representation can aid in interpretability and explainability of a neural network bringing a deeper understanding and knowledge of the mechanisms underlying time series phenomena and can also prove beneficial for improving classification performance in various predictive tasks. Given a dataset $X$ we learn representations $Z,S$ by applying DecVAE on the frame and sequence level as in `Variational decomposition autoencoder' and Alg.~\ref{alg: DecVAE}. We evaluate the quality of representations in several aspects by using the eight different metrics as described in `Disentanglement evaluation' and the annotations provided in the datasets. We also examine the visual quality of the learned spaces by learning and visualizing low-dimensional manifolds of the high-dimensional $Z,S$ spaces through TSNE \cite{maaten2008visualizing}. Specifically, we first apply a PCA to reduce dimensions of spaces $Z,S$ to $50$, and then apply TSNE on the reduced space for visualization. For SimVowels and TIMIT, we evaluate disentanglement on the combined development and test sets, with the training set being used only in the pre-training phase.

\paragraph{Speech recognition.} We perform the phoneme and speaker identification tasks in the SimVowels dataset as a more practical complement to the more theoretical disentanglement quality evaluation. As a toy dataset with generative factors that are fully expressed through frequency-domain properties, SimVowels serves as a controlled experiment that validates the need for variational decomposition autoencoding and demonstrates a case where conventional VAE models fail. Generalization of the performance in both vowel and speaker recognition with the additional constraint of interpretability is the goal of this experiment, and where VAE models are less disentangled. 
We demonstrate in Results that DecVAEs are more disentangled than VAE-based models, PCA and ICA in Figures~\ref{fig2},\ref{fig3} and Supplementary Tables~7-10
. We also perform speech recognition in VOC-ALS and IEMOCAP where a third factor that is involved, and examine whether DecVAE separates phoneme and speaker in the presence of this factor (Figs.~\ref{fig4},\ref{fig5}, Supplementary Tables~11-19). For the SimVowels dataset we evaluate speech recognition on the combined development and test sets, with the training set being used only in the pre-training phase. 

\paragraph{Transfer learning.} We also examine the scenario where DecVAE and VAE models are pre-trained on a dataset $X$, then the pre-trained DecVAE is used to embed a dataset $Y$ in the representations $Z,S$. We test both the zero-shot transfer learning scenario when no adaptation of the DecVAE parameters to the new dataset $Y$ is performed, as well as the fine-tuning scenario where a fine-tuning phase using the VDA framework is performed on $Y$ according to Alg.\ref{alg: DecVAE} to adjust the DecVAE parameters, before proceeding to the evaluation phase. For the transfer learning experiments we use the DecVAE models trained on SimVowels and TIMIT.

\paragraph{Zero-shot evaluation on dysarthric speech.} We evaluate dysarthria severity as a classification task to predict disease duration and King's clinical stage \cite{Roche2012ASclerosis} and as a disentanglement task involving three generative factors of phoneme, speaker and the King's stage that describes disease progression according to failure of anatomical body systems. We utilize the zero-shot transfer learning scheme by transferring models from SimVowels and TIMIT. PCA, ICA, SFA and the PCA of frozen wav2vec2 embeddings (see `Baselines') are instead fitted on VOC-ALS without labels and serve as non-zero-shot references. Dysarthria arises as an effect of vocal tract degeneration and thus is a phenomenon that can be described by alterations in the structural and frequency-domain properties of the speech production system. We explore the capacity of DecVAE and VAE models in capturing the different factors involved in dysarthria in Results through disentanglement and task-specific evaluations (Fig.~\ref{fig4}, Supplementary Tables~11,12,13
). We do not split VOC-ALS in sets but rather perform the cross-validation evaluation with participants shared between folds, as speaker identification and disentanglement evaluation requires. 

Each VOC-ALS participant is characterized by a single King's stage and disease duration value; hence a classifier evaluated on a speaker-shared cross-validation protocol i.e., speakers present in both training and test set, can predict both variables, if these are confounded with speaker identity. To that end, we also evaluate zero-shot classification with stratified, five-fold cross-validation grouped by participant. For this analysis, a participant-level RandomForest probe is fitted on the mean latent of each participant (five folds, ten random states) to assess significance of results through a permutation test that shuffles King's stage and disease duration labels across participants, using 1000 permutations. Further details on this analysis are provided in Supplementary Section~11.4. Disentanglement evaluation is not affected by this analysis, as all speakers need to be present in both training and test sets to evaluate disentanglement involving the three generative factors.

\paragraph{Speech emotion recognition fine-tuning evaluation.} We evaluate emotion recognition performance in relation to speaker and phoneme identification by transferring models and fine-tuning them on IEMOCAP before evaluation. We optimize $\mathcal{L_{VDA}}$ over a period of 25 epochs with a constant learning rate of $5*10^{-6}$ for both $Z,S$ branches without early stopping. Emotion is an attribute not directly encoded in the frequency dynamics of the speech production system and thus presents a more challenging task than dysarthria severity in terms of disentanglement. In Results, we show how DecVAE learns can disentangle phoneme and speaker from emotional speech representations compared to other models
(Fig.~\ref{fig5}, Supplementary Tables~14-19). For comparison purposes with the baseline models in `Baselines', we adopt a subsampling procedure for IEMOCAP to amortize the computational cost across models; as emotion is labeled per utterance, consecutive frames within an utterance (i.e. $25ms$ windows with a $20ms$ stride) are highly redundant. Therefore, for the supervised evaluation only, we train classifiers and evaluate disentanglement on $k=20$ evenly spaced frames of each training utterance, thus retaining approximately $10\%$ of the training utterance data, and keeping every utterance, speaker and session. Despite this reduction, classifiers are evaluated on all frames of the held-out speaker or session. This protocol is followed in the baseline comparisons of DecVAE with self-supervised LAMs and time-series self-supervised models, with results presented in Supplementary Tables~16-19 and Supplementary Fig.~28; the core comparison set of VAE-based, ICA, PCA and other models are also evaluated on the full IEMOCAP protocol, i.e. training and evaluating on all frames, as presented in Supplementary Tables~14,15. 

\paragraph{Latent response analysis and interpretability.} For interpretability of the latent space dynamics and model response on different tasks, we perform a latent response analysis, the encoder-only counterpart of a latent traversal \cite{Higgins2017Beta-VAE:Framework}. We collect small subsets from all four datasets where single generative factors are fixed and other factors are left variable. We then apply DecVAE on these subsets without training and collect the set of dimensions $Z_t \in Z, Z_t \subset Z$ that demonstrate the larger and lower variance. We then draw the latent amplitude of the DecVAE responses to each pair of factors to demonstrate how traversing single factor dimensions is reflected on single latent dimensions $z_t$. This analysis provides insight on how VDA differs from variational autoencoding, and how DecVAE allocates information for different combinations of generative factor values in different dimensions, in a way that promotes disentanglement and informativeness (Fig.~\ref{fig6}a,b,c, Supplementary Fig. 18,19,20,21
).

\paragraph{Subspace-wise analysis.} We evaluate the properties of individual subspaces in the DecVAE latent representations. The frame-level latent of DecVAE is the concatenation of $C+1$ subspaces of size $z_{dim}$, $\tilde{Z}=[\tilde{z}_0, ..., \tilde{z}_C]$. On a fixed subsample of $30,000/60,000$ (SimVowels,SimCoupled/TIMIT) frames, we analyze the representations by fitting linear (Logistic regression) probes~\cite{alain2016understanding} and splitting into five folds stratified by speaker; results are averaged over folds and where available, over three training seeds. On SimCoupled, probes are trained on the correlated split and also evaluated on the independent split. We evaluate using chance-normalized balanced accuracy~\cite{brodersen2010balanced} and $s_f$ denotes the score of a probe on the full latent for factor $f$. Further, the prediction score $P_{g,f}$ is the score of a probe trained on subspace $g$ to predict factor $f$, ablation score $A_{g,f}=s_f-s_f^{(-g)}$ is scored by retraining a probe without subspace $g$ to predict $f$ and indicates the information about $f$ that only $g$ provides. For the swap test~\cite{geiger2021causal} where subspaces are exchanged, we form pairs of frames \textit{(a,b)} that differ only in $f$ and agree on all other factors; then subspace $g$ of \textit{a} is replaced by subspace $g$ of \textit{b}, and the probes $s_f$ trained on the full latent are applied. $T_{g,f}$ denotes the fraction of pairs whose predicted $f$ becomes that of \textit{b}. Since the exchange of a subspace perturbs the representation even if no factor changes, the same swap is made with a control frame \textit{c} that has the labels of \textit{a}, and the swap score that we report is $\Delta T_{g,f}= T_{g,f}- T^c_{g,f}$. Since DecVAE has no decoder, the swap acts on the predictions of the probes, and not on generated signals. 

\paragraph{Selectivity and null.} For $M\in\{P,A,\Delta T\}$ as defined above, the selectivity of subspace $g$ is given as the DCI-disentanglement, adapted for subspace $g$~\cite{Eastwood2018ARepresentations}:
\begin{equation}
    \sigma_g = 1 - \frac{H(\mathbf{p}_g)}{\log	|F|},	\qquad	p_{g,f}	=	\frac{M_{g,f}}{\sum_{f'}	M_{g,f'}}
    \label{eq_selectivity}
\end{equation}
\noindent
where $F$ is the set of factors, and the selectivity of a representation is the average of $\sigma_g$ weighted by the row sums of matrix $M$. As a control~\cite{hewitt2019designing}, we compute the same quantity $\sigma_g$ for 20 random partitions of the latent dimensions into $C+1$ groups of size $z_{dim}$ and report how many of these partitions are exceeded by DecVAE subspaces. On SimVowels, the expected allocation of information into subspaces follows the data generator: for each formant we compute the fraction of the variance of its log-frequency explained by the vowel and by the speaker, using Cohen's $\eta^2$~\cite{cohen1973eta} (see Supplementary Fig.~34a). For each factor $f$ we also report the effective number of component subspaces that carry it: $n_{\mathrm{eff}} = \exp(H(q_f))$, where where $H(q_f) = -\sum_{g} q_{g,f} \ln q_{g,f}$, and $q_{g,f} = P_{g,f} / \sum_{g'} P_{g',f}$ over the $C$ component subspaces (the subspace of the original signal is excluded, and negative entries are set to zero). Under this formulation, $n_\mathrm{eff}$ ranges from 1,
when a single component carries the factor, to $C$, when the factor is spread evenly over all components. Supplementary Fig.~35f reports the same quantity computed from the swap scores $\Delta T_{g,f}$.

\paragraph{Stability.} Across random seeds, we report the Pearson correlation between the factor matrices and linear CKA~\cite{kornblith2019similarity} between the subspaces, i.e., $CKA(X,Y) = \|X^{\top}Y\|^2_F / (\|X^\top X\|_F \|Y^\top Y\|_F)$. Across folds, we report the Spearman correlation of the dimension importances ($L2$ norm over classes of the standardized probe coefficients). Implementation details and models analysed are listed in Supplementary Section~11.6.

\paragraph{Sensitivity to the decomposition.} We retrained $\beta$-DecVAE with FD ($\beta=0.1$) with $C \in \{2,3,4,5,6\}$ on SimVowels and SimCoupled. On SimVowels with $C=3$, we test six alternative peak detection schemes w.r.t. the formation of the frequency detection intervals (linear, three and eight power-law intervals, random intervals with three draws). Each configuration is trained once except $C=3$ that used three random seeds. We report the selectivity of the component subspaces minus the mean of 20 random partitions and $n_{eff}$ (see `Subspace-wise analysis'). Results of this analysis are presented in Fig.~\ref{fig8}, Supplementary Figs.~35,36. More details are given in Supplementary Section~11.7.

\subsection*{Reporting summary}
Further information on research design is available in the Nature Portfolio Reporting Summary linked to this article.

\section*{Data Availability}
The pseudo-code for generating SimVowels is provided in Supplementary Alg.~1 
and Python code is also available in our publicly available codebase at \hyperlink{https://github.com/GiannisZgs/DecVAE}{https://github.com/GiannisZgs/DecVAE}. We also make the SimVowels dataset publicly available for downloading at 
\href{https://drive.google.com/drive/folders/1Fl8Ewuwfv-rENiNDAQWm3gd0L3IifqFD?usp=sharing}
{https://drive.google.com/drive/folders/1Fl8E-wuwfvrENiNDAQWm3gd0L3IifqFD?usp=sharing}.
The code for generating SimCoupled is available in the same codebase as SimVowels, and we make SimCoupled publicly available for downloading at \href{https://drive.google.com/drive/folders/10KmxJX6bIQDvss3TLuN8S5nXtptJYcLd?usp=sharing}{https://drive-.google.com/drive/folders/10KmxJX6bIQDvss3TLuN8S5nXtptJYcLd?usp=sharing}.
TIMIT can be downloaded from \url{https://catalog.ldc.upenn.edu/LDC93S1}.
VOC-ALS can be downloaded from \url{https://www.synapse.org/Synapse:syn53009474/wiki/624730}.
IEMOCAP can be downloaded from \url{https://sail.usc.edu/iemocap/}.

\section*{Code Availability}
We release a codebase in a detailed format with usage scenarios to reproduce results of this work, but also to allow researchers and practitioners to apply, build upon and extend variational decomposition autoencoding. The codebase is available at \hyperlink{https://github.com/GiannisZgs/DecVAE}{https://github.com/GiannisZgs/DecVAE} \cite{zenodo2025decvae}. 
All code was implemented in R v.4.5.1 and Python v3.9.11 using PyTorch \cite{paszke2019pytorch} and HuggingFace \cite{wolf2020HFtransformers}. For disentanglement metrics implementation we used $disentanglement\_lib$ from Google Research \cite{Locatello2019ChallengingRepresentations}. The \textit{ggbreak} \cite{xu2021ggbreak} package was used for R visualizations.

\section*{Author Contributions}
I.N.Z. and L.J.H. had the initial concept and formulated the theoretical approach. I.N.Z. implemented the codes, did the experimental work and analysis, wrote the main manuscript text, and prepared all figures (main and supplementary material). L.J.H. reviewed the manuscript drafts and contributed to its final version. A.H.K. and A.A. reviewed the final draft and the final version of the manuscript.

\section*{Acknowledgments}
Not applicable.

\section*{Funding}
This research was funded and supported by the Healthcare Engineering Innovation Group (HEIG), Khalifa University of Science and Technology (KU-HEIG). The funder played no role in study design, data collection, analysis and interpretation of data, or the writing of this manuscript.

\section*{Competing Interests}
All authors declare no financial or non-financial competing interests. 

\bibliographystyle{unsrt}
\bibliography{references}

@inproceedings{Kingma2014Auto-EncodingBayes,
    title = {{Auto-Encoding Variational Bayes}},
    year = {(2014)},
    booktitle = {International Conference on Learning Representations},
    author = {D. P., Kingma  and M., Welling},
    url = {https://dare.uva.nl},
    arxivId = {1312.6114v10}
}

@article{Hsu2017UnsupervisedData,
    title = {{Unsupervised Learning of Disentangled and Interpretable Representations from Sequential Data}},
    year = {(2017)},
    journal = {Advances in Neural Information Processing Systems},
    author = {W.N., Hsu and Y., Zhang and J., Glass},
    pages = {1879-1890},
    volume = {},
    publisher = {Neural information processing systems foundation},
    url = {https://arxiv.org/abs/1709.07902v1},
    issn = {10495258},
    arxivId = {1709.07902}
}

@inproceedings{Higgins2017Beta-VAE:Framework,
    title = {{beta-VAE: Learning Basic Visual Concepts with a Constrained Variational Framework}},
    year = {(2017)},
    booktitle = {International conference on learning representations},
    author = {I. et al., Higgins}
}

@article{Boulianne2020ALearning,
    title = {{A Study of Inductive Biases for Unsupervised Speech Representation Learning}},
    year = {(2020)},
    journal = {IEEE/ACM Transactions on Audio Speech and Language Processing},
    author = {G., Boulianne},
    pages = {2781--2795},
    volume = {\textbf{28}},
    publisher = {Institute of Electrical and Electronics Engineers Inc.},
    doi = {10.1109/TASLP.2020.3030494},
    issn = {23299304}
}

@article{gilles2013ewt,
    title = {{Empirical wavelet transform}},
    year = {(2013)},
    journal = {IEEE Transactions on Signal Processing},
    author = {J., Gilles},
    number = {16},
    pages = {3999--4010},
    volume = {\textbf{61}},
    doi = {10.1109/TSP.2013.2265222},
    issn = {1053587X}
}

@article{dragomiretskiy2014vmd,
    title = {{Variational mode decomposition}},
    year = {(2014)},
    journal = {IEEE Transactions on Signal Processing},
    author = {K., Dragomiretskiy and D., Zosso},
    number = {3},
    pages = {531--544},
    volume = {\textbf{62}},
    doi = {10.1109/TSP.2013.2288675},
    issn = {1053587X}
}

@article{Baevski2020Wav2vecRepresentations,
    title = {{wav2vec 2.0: A Framework for Self-Supervised Learning of Speech Representations}},
    year = {(2020)},
    journal = {Advances in Neural Information Processing Systems},
    author = {A., Baevski and Y., Zhou and A., Mohamed and M., Auli},
    pages = {12449--12460},
    volume = {\textbf{33}},
    url = {https://github.com/pytorch/fairseq}
}

@article{vandenOordDeepMind2018RepresentationCoding,
    title = {{Representation Learning with Contrastive Predictive Coding}},
    year = {(2018)},
    author = {A., van den Oord and Y., Li and O., Vinyals},
    journal = {Preprint at \url{https://arxiv.org/abs/1807.03748}},
    url = {https://arxiv.org/pdf/1807.03748},
    arxivId = {1807.03748}
}

@article{balestriero2023,
    title = {{A Cookbook of Self-Supervised Learning}},
    year = {(2023)},
    journal = {Preprint at \url{https://arxiv.org/abs/2304.12210v1}},
    author = {R. et al.,  Balestriero},
    url = {https://arxiv.org/abs/2304.12210v1},
    arxivId = {2304.12210}
}

@article{Mohamied2024BootstrapApproach,
    title = {{Bootstrap Predictive Coding: Investigating a Non-Contrastive Self-Supervised Learning Approach}},
    year = {(2024)},
    journal = {IEEE International Conference on Acoustics, Speech and Signal Processing (ICASSP)},
    author = {Y., Mohamied and P., Bell},
    pages = {11541--11545},
    publisher = {IEEE},
    url = {https://ieeexplore.ieee.org/document/10447173/},
    isbn = {979-8-3503-4485-1},
    doi = {10.1109/ICASSP48485.2024.10447173}
}

@article{grill2020,
  title={Bootstrap your own latent-a new approach to self-supervised learning},
  author={J.-B. et al., Grill},
  journal={Advances in neural information processing systems},
  volume={\textbf{33}},
  pages={21271--21284},
  year={(2020)}
}

@article{Wu2021Autoformer:Forecasting,
    title = {{Autoformer: Decomposition Transformers with Auto-Correlation for Long-Term Series Forecasting}},
    year = {(2021)},
    journal = {Advances in Neural Information Processing Systems},
    author = {H., Wu and J., Xu and J., Wang and M., Long},
    pages = {22419--22430},
    volume = {\textbf{27}},
    publisher = {Neural information processing systems foundation},
    url = {https://arxiv.org/abs/2106.13008v5},
    isbn = {9781713845393},
    issn = {10495258},
    arxivId = {2106.13008}
}

@article{Woo2022CoST:Forecasting,
    title = {{CoST: Contrastive Learning of Disentangled Seasonal-Trend Representations for Time Series Forecasting}},
    year = {(2022)},
    journal = {International Conference on Learning Representations},
    author = {G., Woo and C., Liu and D., Sahoo and A., Kumar and S., Hoi},
    publisher = {International Conference on Learning Representations, ICLR},
    url = {https://arxiv.org/abs/2202.01575v3},
    arxivId = {2202.01575}
}

@article{Oreshkin2019N-BEATS:Forecasting,
    title = {{N-BEATS: Neural basis expansion analysis for interpretable time series forecasting}},
    year = {(2020)},
    journal = {International Conference on Learning Representations},
    author = {B.N., Oreshkin and D., Carpov and N., Chapados and Y., Bengio},
    publisher = {International Conference on Learning Representations, ICLR},
    url = {https://arxiv.org/abs/1905.10437v4},
    arxivId = {1905.10437}
}

@inproceedings{Zhou2022FEDformer:Forecasting,
  title={Fedformer: Frequency enhanced decomposed transformer for long-term series forecasting},
  author = {T., Zhou and Z., Ma and Q., Wen and X., Wang and L., Sun and R., Jin},
  booktitle={International conference on machine learning},
  pages={27268--27286},
  year={(2022)},
  organization={PMLR}
}

@book{Ladefoged2006APhonetics,
    title = {{A course in phonetics}},
    year = {2006},
    author = {P., Ladefoged and K., Johnson and P., Ladefoged},
    volume = {\textbf{3}},
    publisher = {Thomson Wadsworth Boston}
}

@article{Menendez1997TheDivergence,
    title = {{The Jensen-Shannon divergence}},
    year = {(1997)},
    journal = {Journal of the Franklin Institute},
    author = {M.L., Men{\'{e}}ndez and J.A., Pardo and L., Pardo and M.C., Pardo},
    number = {2},
    pages = {307--318},
    volume = {\textbf{334}},
    publisher = {Pergamon},
    url = {https://www.sciencedirect.com/science/article/abs/pii/S0016003296000634},
    doi = {10.1016/S0016-0032(96)00063-4},
    issn = {0016-0032}
}

@article{Jing2021UnderstandingLearning,
    title = {{Understanding Dimensional Collapse in Contrastive Self-supervised Learning}},
    year = {(2022)},
    journal = {International Conference on Learning Representations},
    author = {L., Jing and P., Vincent and Y., LeCun and Y., Tian},
    publisher = {International Conference on Learning Representations, ICLR},
    url = {https://arxiv.org/abs/2110.09348v3},
    arxivId = {2110.09348}
}

@article{huang1998empirical,
  title={The empirical mode decomposition and the Hilbert spectrum for nonlinear and non-stationary time series analysis},
  author={N.E. et al., Huang},
  journal={Proceedings of the Royal Society of London. Series A: mathematical, physical and engineering sciences},
  volume={\textbf{454}},
  number={1971},
  pages={903--995},
  year={(1998)},
  publisher={The Royal Society}
}

@article{Ba2016LayerNormalization,
    title = {{Layer Normalization}},
    year = {(2016)},
    author = {J.L., Ba and J.R., Kiros and G.E. ,Hinton},
    journal = {Preprint at \url{https://arxiv.org/pdf/1607.06450}},
    url = {https://arxiv.org/pdf/1607.06450},
    arxivId = {1607.06450}
}

@article{Hendrycks2016GaussianGELUs,
    title = {{Gaussian Error Linear Units (GELUs)}},
    year = {(2016)},
    author = {D., Hendrycks and K., Gimpel},
    journal = {Preprint at \url{https://arxiv.org/pdf/1606.08415}},
    url = {https://arxiv.org/pdf/1606.08415},
    arxivId = {1606.08415}
}

@article{kingma2014Adam,
    title = {{Adam: A Method for Stochastic Optimization}},
    year = {(2015)},
    journal = {International Conference on Learning Representations},
    author = {D.P., Kingma and J.L., Ba},
    publisher = {International Conference on Learning Representations, ICLR},
    url = {https://arxiv.org/abs/1412.6980v9},
    arxivId = {1412.6980}
}

@article{Locatello2019ChallengingRepresentations,
    title = {{Challenging Common Assumptions in the Unsupervised Learning of Disentangled Representations}},
    year = {(2019)},
    author = {F., Locatello and S., Bauer and M., Lucic and G., Raetsch and S., Gelly and B., Sch{\"{o}}lkopf and O., Bachem},
    pages = {4114--4124},
    publisher = {PMLR},
    journal = {Proceedings of the 36th International Conference on Machine Learning},
    url = {https://proceedings.mlr.press/v97/locatello19a.html},
    volume = {\textbf{97}},
    issn = {2640-3498}
}

@article{suter2019irs,
    title = {{Robustly Disentangled Causal Mechanisms: Validating Deep Representations for Interventional Robustness}},
    year = {(2019)},
    author = {R., Suter and D., Miladinovic and B., Sch{\"{o}}lkopf and S., Bauer},
    pages = {6056--6065},
    publisher = {PMLR},
    journal = {Proceedings of the 36th International Conference on Machine Learning},
    url = {https://proceedings.mlr.press/v97/suter19a.html},
    volume = {\textbf{97}},
    issn = {2640-3498}
}

@article{Ridgeway2018LearningLoss,
    title = {{Learning Deep Disentangled Embeddings With the F-Statistic Loss}},
    year = {(2018)},
    journal = {Advances in Neural Information Processing Systems},
    author = {K., Ridgeway and M.C., Mozer},
    volume = {\textbf{31}}
}

@inproceedings{Eastwood2018ARepresentations,
    title = {{A Framework for the Quantitative Evaluation of Disentangled Representations}},
    year = {(2018)},
    booktitle = {International Conference on Learning Representations},
    author = {C., Eastwood and C.K.I., Williams},
    url = {https://www.research.ed.ac.uk/en/publications/a-framework-for-the-quantitative-evaluation-of-disentangled-repre}
}

@article{Antoniou2023DesigningIEMOCAP,
    title = {{Designing and Evaluating Speech Emotion Recognition Systems: A Reality Check Case Study with IEMOCAP}},
    year = {(2023)},
    journal = {IEEE International Conference on Acoustics, Speech and Signal Processing (ICASSP)},
    author = {N., Antoniou and A., Katsamanis and T., Giannakopoulos and S., Narayanan},
    pages = {1--5},
    publisher = {Institute of Electrical and Electronics Engineers (IEEE)},
    doi = {10.1109/ICASSP49357.2023.10096808},
    arxivId = {2304.00860}
}

@article{carbonneau2022dis_metrics,
    title = {{Measuring Disentanglement: A Review of Metrics}},
    year = {(2022)},
    journal = {IEEE Transactions on Neural Networks and Learning Systems},
    author = {M.A., Carbonneau and J., Zaidi and J., Boilard and G., Gagnon},
    pages = {8747--8761},
    volume = {\textbf{35}},
    publisher = {Institute of Electrical and Electronics Engineers Inc.},
    doi = {10.1109/TNNLS.2022.3218982},
    issn = {21622388},
    pmid = {36374888},
    arxivId = {2012.09276}
}

@article{Lee1989Speaker-IndependentModels,
    title = {{Speaker-Independent Phone Recognition Using Hidden Markov Models}},
    year = {(1989)},
    journal = {IEEE Transactions on Acoustics, Speech, and Signal Processing},
    author = {K.F., Lee and H.W., Hon},
    pages = {1641--1648},
    volume = {\textbf{37}},
    doi = {10.1109/29.46546},
    issn = {00963518}
}

@article{Dubbioso2024VoiceControls,
    title = {{Voice signals database of ALS patients with different dysarthria severity and healthy controls}},
    year = {(2024)},
    journal = {Scientific Data},
    author = {R. et al., Dubbioso},
    pages = {1--14},
    volume = {\textbf{11}},
    publisher = {Nature Research},
    url = {https://www.nature.com/articles/s41597-024-03597-2},
    doi = {10.1038/S41597-024-03597-2;SUBJMETA},
    issn = {20524463},
    pmid = {39030186}
}

@article{busso2008iemocap,
    title = {{IEMOCAP: Interactive emotional dyadic motion capture database}},
    year = {(2008)},
    journal = {Language Resources and Evaluation},
    author = {C. et al., Busso},
    pages = {335--359},
    volume = {\textbf{42}},
    publisher = {Springer},
    url = {https://link.springer.com/article/10.1007/s10579-008-9076-6},
    doi = {10.1007/S10579-008-9076-6/METRICS},
    issn = {1574020X}
}

@article{maaten2008visualizing,
  title={Visualizing data using t-SNE},
  author={L., van der Maaten and G., Hinton},
  journal={Journal of machine learning research},
  volume={\textbf{9}},
  pages={2579--2605},
  year={(2008)}
}

@incollection{hyvarinen2001independent,
  title={Independent component analysis},
  author={A. Hyv{\"a}rinen and J. , Hurri and P.O., Hoyer},
  booktitle={Natural Image Statistics: A Probabilistic Approach to Early Computational Vision},
  pages={151--175},
  year={(2001)},
  publisher={Springer}
}

@article{pedregosa2011scikit,
  title={Scikit-learn: Machine learning in Python},
  author={F. et al., Pedregosa},
  journal={The Journal of Machine Learning Research},
  volume={\textbf{12}},
  pages={2825--2830},
  year={(2011)},
  publisher={JMLR. org}
}

@article{Greenacre2022PrincipalAnalysis,
    title = {{Principal component analysis}},
    year = {(2022)},
    journal = {Nature Reviews Methods Primers},
    author = {M., Greenacre and P.J.F., Groenen and T., Hastie and A.I., D’Enza and A., Markos and E., Tuzhilina},
    pages = {1--21},
    volume = {\textbf{2}},
    publisher = {Springer Nature},
    url = {https://www.nature.com/articles/s43586-022-00184-w},
    doi = {10.1038/S43586-022-00184-W;SUBJMETA},
    issn = {26628449}
}

@article{Scholkopf1997KernelAnalysis,
    title = {{Kernel principal component analysis}},
    year = {(1997)},
    journal = {Lecture Notes in Computer Science (including subseries Lecture Notes in Artificial Intelligence and Lecture Notes in Bioinformatics)},
    author = {B., Sch{\"{o}}lkopf and A., Smola and K.R., M{\"{u}}ller},
    pages = {583--588},
    volume = {\textbf{1327}},
    publisher = {Springer, Berlin, Heidelberg},
    url = {https://link.springer.com/chapter/10.1007/BFb0020217},
    isbn = {978-3-540-69620-9},
    doi = {10.1007/BFB0020217},
    issn = {1611-3349}
}

@article{kim2018factorVAE,
  title = 	 {Disentangling by Factorising},
  author =       {H., Kim and A., Mnih},
  journal = 	 {International Conference on Machine Learning},
  pages = 	 {2649--2658},
  year = 	 {(2018)},
  volume = 	 {\textbf{80}},
  url = 	 {https://proceedings.mlr.press/v80/kim18b.html}
}

@article{Mathieu2018DisentanglingAutoencoders,
    title = {{Disentangling Disentanglement in Variational Autoencoders}},
    year = {(2019)},
    journal = {International Conference on Machine Learning},
    author = {E., Mathieu and T., Rainforth and N., Siddharth and Y.W., Teh},
    pages = {7744--7754},
    publisher = {International Machine Learning Society (IMLS)},
    url = {https://arxiv.org/abs/1812.02833v3},
    isbn = {9781510886988},
    arxivId = {1812.02833}
}

@article{Rolinek2018VariationalAccident,
    title = {{Variational Autoencoders Pursue PCA Directions (by Accident)}},
    year = {(2018)},
    journal = {Proceedings of the IEEE Computer Society Conference on Computer Vision and Pattern Recognition},
    author = {M., Rolinek and D., Zietlow and G., Martius},
    pages = {12398--12407},
    publisher = {IEEE Computer Society},
    url = {https://arxiv.org/abs/1812.06775v2},
    isbn = {9781728132938},
    doi = {10.1109/CVPR.2019.01269},
    issn = {10636919},
    arxivId = {1812.06775}
}

@article{Ridgeway2016ARepresentation-Learning,
    title = {{A Survey of Inductive Biases for Factorial Representation-Learning}},
    year = {(2016)},
    author = {K., Ridgeway},
    journal = {Preprint at \url{https://arxiv.org/pdf/1612.05299}},
    url = {https://arxiv.org/pdf/1612.05299},
    isbn = {1612.05299v1},
    arxivId = {1612.05299}
}

@article{fiser2010,
    title = {{Statistically optimal perception and learning: from behavior to neural representations}},
    year = {(2010)},
    journal = {Trends in Cognitive Sciences},
    author = {J., Fiser and P., Berkes and G., Orb{\'{a}}n  and M., Lengyel},
    pages = {119--130},
    volume = {\textbf{14}},
    publisher = {Elsevier},
    url = {http://www.cell.com/article/S1364661310000045/fulltext http://www.cell.com/article/S1364661310000045/abstract https://www.cell.com/trends/cognitive-sciences/abstract/S1364-6613(10)00004-5},
    doi = {10.1016/J.TICS.2010.01.003},
    issn = {1364-6613},
    pmid = {20153683}
}

@article{CohenKalafut2023JointEmbedding,
    title = {{Joint variational autoencoders for multimodal imputation and embedding}},
    year = {(2023)},
    journal = {Nature Machine Intelligence},
    author = {N.K., Cohen and X., Huang and D., Wang},
    pages = {631--642},
    volume = {\textbf{5}},
    publisher = {Nature Research},
    url = {https://www.nature.com/articles/s42256-023-00663-z},
    doi = {10.1038/S42256-023-00663-Z;SUBJMETA},
    issn = {25225839}
}

@article{Kay2020AFMRI,
    title = {{A temporal decomposition method for identifying venous effects in task-based fMRI}},
    year = {(2020)},
    journal = {Nature Methods},
    author = {K., Kay and K.W., Jamison and R.Y., Zhang and K., U{\u{g}}urbil},
    pages = {1033--1039},
    volume = {\textbf{17}},
    publisher = {Nature Research},
    url = {https://www.nature.com/articles/s41592-020-0941-6},
    doi = {10.1038/S41592-020-0941-6;SUBJMETA},
    issn = {15487105},
    pmid = {32895538}
}

@article{Isomura2021DimensionalityCapability,
    title = {{Dimensionality reduction to maximize prediction generalization capability}},
    year = {(2021)},
    journal = {Nature Machine Intelligence},
    author = {T., Isomura and T., Toyoizumi},
    pages = {434--446},
    volume = {\textbf{3}},
    publisher = {Nature Publishing Group},
    url = {https://www.nature.com/articles/s42256-021-00306-1},
    doi = {10.1038/s42256-021-00306-1},
    issn = {2522-5839},
    arxivId = {2003.00470}
}

@article{chen2022disentangled,
    title = {{Speaker-Independent Emotional Voice Conversion via Disentangled Representations}},
    year = {(2022)},
    journal = {IEEE Transactions on Multimedia},
    author = {X., Chen and X., Xu  and T., Kamihigashi and J., Chen and Z., Zhang and T., Takiguchi and E.R., Hancock},
    publisher = {Institute of Electrical and Electronics Engineers Inc.},
    doi = {10.1109/TMM.2022.3222646},
    issn = {19410077}
}

@article{apostolidis2017swd,
    title = {{Swarm decomposition: A novel signal analysis using swarm intelligence}},
    year = {(2017)},
    journal = {Signal Processing},
    author = {G.K., Apostolidis and L.J., Hadjileontiadis},
    pages = {40--50},
    volume = {\textbf{132}},
    publisher = {Elsevier},
    doi = {10.1016/J.SIGPRO.2016.09.004},
    issn = {0165-1684}
}

@article{Hyvarinen2016UnsupervisedICA,
    title = {{Unsupervised Feature Extraction by Time-Contrastive Learning and Nonlinear ICA}},
    year = {(2016)},
    journal = {Advances in Neural Information Processing Systems},
    author = {A., Hyvarinen and H., Morioka},
    volume = {\textbf{29}}
}

@article{Chen2018IsolatingAutoencoders,
    title = {{Isolating Sources of Disentanglement in Variational Autoencoders}},
    year = {(2018)},
    journal = {Advances in Neural Information Processing Systems},
    author = {R.T.Q., Chen and X., Li and R.B., Grosse and D.K., Duvenaud},
    volume = {\textbf{31}}
}

@article{Roche2012ASclerosis,
    title = {{A proposed staging system for amyotrophic lateral sclerosis}},
    year = {(2012)},
    journal = {Brain},
    author = {J.C. et al., Roche},
    pages = {847--852},
    volume = {\textbf{135}},
    publisher = {Oxford Academic},
    url = {https://dx.doi.org/10.1093/brain/awr351},
    doi = {10.1093/BRAIN/AWR351},
    issn = {0006-8950},
    pmid = {22271664}
}

@article{garofolo1993timit,
  title={TIMIT acoustic-phonetic continuous speech corpus},
  year = {(1993)},
  author = {J.S. et al., Garofolo},
  journal = {LDC93S1 Philadelphia: Linguistic Data Consortium},
  publisher={Linguistic data consortium}
}

@inproceedings{wolf2020HFtransformers,
    title = {Transformers: State-of-the-Art Natural Language Processing},
    author = {T. et al., Wolf},
    booktitle = {Proceedings of the 2020 Conference on Empirical Methods in Natural Language Processing: System Demonstrations},
    year = {(2020)},
    pages = {38--45}
}

@article{paszke2019pytorch,
  title={Pytorch: An imperative style, high-performance deep learning library},
  author={A. et al., Paszke},
  journal={Advances in neural information processing systems},
  volume={\textbf{32}},
  year={(2019)}
}

@article{xu2021ggbreak,
  title={Use ggbreak to effectively utilize plotting space to deal with large datasets and outliers},
  author={X., Shuangbin and M., Chen and T., Feng and L., Zhan and L., Zhou and G., Yu},
  journal={Frontiers in Genetics},
  volume={\textbf{12}},
  year={(2021)},
  publisher={Frontiers}
}

@article{zenodo2025decvae,
    author = {I.N., Ziogas and A., Al Shehhi and A.H., Khandoker and L.J., Hadjileontiadis},
    title = {Variational decomposition autoencoding improves disentanglement of latent representations},
    journal = {Zenodo, \url{https://doi.org/10.5281/zenodo.17773435}},
    year = {(2025)}
}

@article{chen2022wavlm,
  title = {{WavLM: Large-Scale Self-Supervised Pre-Training for Full Stack Speech Processing}},
  year = {(2022)},
  journal = {IEEE Journal on Selected Topics in Signal Processing},
  author = {S. et al., Chen},
  number = {6},
  month = {10},
  pages = {1505--1518},
  volume = {\textbf{16}},
  publisher = {Institute of Electrical and Electronics Engineers Inc.},
  doi = {10.1109/JSTSP.2022.3188113},
  issn = {19410484},
  arxivid = {2110.13900},
}

@article{Hsu2021HuBERT:Units,
  title = {{HuBERT: Self-Supervised Speech Representation Learning by Masked Prediction of Hidden Units}},
  year = {(2021)},
  journal = {IEEE/ACM Transactions on Audio Speech and Language Processing},
  author = {W.N., Hsu and B., Bolte and Y.H.H., Tsai and K., Lakhotia and R., Salakhutdinov and A., Mohamed},
  month = {6},
  pages = {3451--3460},
  volume = {\textbf{29}},
  publisher = {Institute of Electrical and Electronics Engineers Inc.},
  url = {https://arxiv.org/abs/2106.07447v1},
  doi = {10.1109/TASLP.2021.3122291},
  issn = {23299304},
  arxivid = {2106.07447},
}

@article{Zhang2022Self-SupervisedConsistency,
  title = {{Self-Supervised Contrastive Pre-Training For Time Series via Time-Frequency Consistency}},
  year = {(2022)},
  journal = {Advances in Neural Information Processing Systems},
  author = {X., Zhang and Z., Zhao and T., Tsiligkaridis and M., Zitnik},
  month = {6},
  volume = {\textbf{35}},
  publisher = {Neural information processing systems foundation},
  url = {https://arxiv.org/abs/2206.08496v3},
  isbn = {9781713871088},
  issn = {10495258},
  arxivid = {2206.08496},
}

@article{wiskott2002slow,
  title={Slow feature analysis: Unsupervised learning of invariances},
  author={L., Wiskott and T.J., Sejnowski},
  journal={Neural computation},
  volume={\textbf{14}},
  number={4},
  pages={715--770},
  year={(2002)},
  publisher={MIT Press}
}

@article{alain2016understanding,
  title={Understanding intermediate layers using linear classifier probes},
  author={G., Alain and Y., Bengio},
  journal={arXiv preprint arXiv:1610.01644},
  year={(2016)}
}

@inproceedings{kornblith2019similarity,
  title={Similarity of neural network representations revisited},
  author={S., Kornblith and M., Norouzi and H., Lee and G., Hinton},
  booktitle={International conference on machine learning},
  pages={3519--3529},
  year={(2019)},
  organization={PMLR}
}

@article{geiger2021causal,
  title={Causal abstractions of neural networks},
  author={A., Geiger and H., Lu and T., Icard and C., Potts},
  journal={Advances in neural information processing systems},
  volume={\textbf{34}},
  pages={9574--9586},
  year={(2021)}
}

@article{cohen1973eta,
  title={Eta-squared and partial eta-squared in fixed factor ANOVA designs},
  author={J., Cohen},
  journal={Educational and psychological measurement},
  volume={\textbf{33}},
  number={1},
  pages={107--112},
  year={(1973)},
  publisher={Sage Publications, Sage CA: Thousand Oaks, CA}
}

@inproceedings{hewitt2019designing,
  title={Designing and interpreting probes with control tasks},
  author={J., Hewitt and P., Liang},
  booktitle={Proceedings of the 2019 conference on empirical methods in natural language processing and the 9th international joint conference on natural language processing (EMNLP-IJCNLP)},
  pages={2733--2743},
  year={(2019)}
}

@inproceedings{brodersen2010balanced,
  title={The balanced accuracy and its posterior distribution},
  author={K.H., Brodersen and C.S., Ong and K.E., Stephan and J.M., Buhmann},
  booktitle={2010 20th international conference on pattern recognition},
  pages={3121--3124},
  year={(2010)},
  organization={IEEE}
}

@article{ding200617,
  title={Granger causality: basic theory and application to neuroscience},
  author={M., Ding and Y., Chen and S.L., Bressler},
  journal={Handbook of time series analysis: recent theoretical developments and applications},
  volume={\textbf{437}},
  year={(2006)}
}

@article{yang2018causal,
  title={Causal decomposition in the mutual causation system},
  author={A.C., Yang and C.K., Peng and N.E., Huang},
  journal={Nature communications},
  volume={\textbf{9}},
  number={1},
  pages={3378},
  year={(2018)},
  publisher={Nature Publishing Group UK, London}
}

@article{wang2022latent,
  title={Learning latent seasonal-trend representations for time series forecasting},
  author={Z., Wang and  X., Xu and  W., Zhang and G., Trajcevski and  T., Zhong and F.,Zhou},
  journal={Advances in Neural information processing systems},
  volume={\textbf{35}},
  pages={38775--38787},
  year={(2022)}
}

\end{document}